\documentclass[]{llncs}
\usepackage{graphicx}

\newcommand{\RNum}[1]{\uppercase\expandafter{\romannumeral #1\relax}}
\usepackage{hyperref}

\usepackage{tikz}
\usepackage{comment}
\usepackage{amsmath,amssymb} 
\usepackage{color}

\usepackage[accsupp]{axessibility}  

\usepackage[width=122mm,left=12mm,paperwidth=146mm,height=193mm,top=12mm,paperheight=217mm]{geometry}

\usepackage{bbding} 
\usepackage{subfigure}

\usepackage{multirow}

\begin{document}
\pagestyle{headings}
\mainmatter
\def\ECCVSubNumber{196}  

\title{P-STMO: Pre-Trained Spatial Temporal Many-to-One Model for 3D Human Pose Estimation } 

\titlerunning{ECCV-22 submission ID \ECCVSubNumber} 
\authorrunning{ECCV-22 submission ID \ECCVSubNumber} 
\author{Anonymous ECCV submission}
\institute{Paper ID \ECCVSubNumber}

\titlerunning{P-STMO}
%
\author{Wenkang Shan\inst{1} \and
Zhenhua Liu\inst{1,3} \and
Xinfeng Zhang\inst{2} \and Shanshe Wang\inst{1,4} \and Siwei Ma\inst{1,4} \and Wen Gao\inst{1,4}}
\authorrunning{W. Shan et al.}
%
\institute{National Engineering Research Center of Visual Technology, Peking University \and
University of Chinese Academy of Sciences \and
Huawei Noah’s Ark Lab \and
Peng Cheng Laboratory
\\
\email{\{wkshan,liu-zh,sswang,swma,wgao\}@pku.edu.cn,  xfzhang@ucas.ac.cn}
}
\maketitle

\begin{abstract}
This paper introduces a novel Pre-trained Spatial Temporal Many-to-One (P-STMO) model for 2D-to-3D human pose estimation task. To reduce the difficulty of capturing spatial and temporal information, we divide this task into two stages: pre-training (Stage \RNum{1}) and fine-tuning (Stage \RNum{2}). In Stage \RNum{1}, a self-supervised pre-training sub-task, termed \emph{masked pose modeling}, is proposed. The human joints in the input sequence are randomly masked in both spatial and temporal domains. A general form of denoising auto-encoder is exploited to recover the original 2D poses and the encoder is capable of capturing spatial and temporal dependencies in this way. In Stage \RNum{2}, the pre-trained encoder is loaded to STMO model and fine-tuned. The encoder is followed by a many-to-one frame aggregator to predict the 3D pose in the current frame. Especially, an MLP block is utilized as the spatial feature extractor in STMO, which yields better performance than other methods. In addition, a temporal downsampling strategy is proposed to diminish data redundancy. Extensive experiments on two benchmarks show that our method outperforms state-of-the-art methods with fewer parameters and less computational overhead. For example, our P-STMO model achieves 42.1mm MPJPE on Human3.6M dataset when using 2D poses from CPN as inputs. Meanwhile, it brings a 1.5--7.1$\times$ speedup to state-of-the-art methods. Code is available at https://github.com/paTRICK-swk/P-STMO.

\end{abstract}

\section{Introduction}

Monocular 3D human pose estimation in videos is a long-standing computer vision task with extensive applications, such as virtual reality, medical assistance, and self-driving. Two-step estimation methods~\cite{lin2019trajectory,fang2018learning,liu2020learning,li2019generating,hu2021conditional} first detect 2D human keypoints, and then regress the 3D position of each joint. They use an off-the-shelf 2D keypoint detector (e.g.,~\cite{newell2016stacked,chen2018cascaded}) to obtain 2D poses and mainly focus on lifting 3D poses from these 2D keypoints. Despite considerable success achieved, this task remains an ill-posed problem due to depth ambiguity.

Existing works~\cite{martinez2017simple,pavllo20193d,liu2020attention,chen2021anatomy,ci2019optimizing,zhao2019semantic} rely on fully-connected (\emph{fc}) layers, graph convolutions, or 1D convolutions to integrate information over spatial and temporal domains. Recently, some works~\cite{zheng20213d,lin2021end,li2022exploiting} introduce Transformer~\cite{vaswani2017attention} to 3D human pose estimation. The self-attention mechanism of Transformer is exploited to depict spatial dependencies between joints in each frame or temporal dependencies between frames in a sequence. However, there are two drawbacks to these methods: (1) They directly learn 2D-to-3D spatial and temporal correlations, which is a challenging task. This might make it difficult to optimize the model. (2) Previous works~\cite{dosovitskiy2021an} show that Transformer requires more training data than convolutional neural networks. 

For these two issues, self-supervised pre-training of Transformer, which has proven to be effective in natural language processing (NLP) and computer vision (CV), is a promising solution. Previous approaches~\cite{devlin2018bert,chen2020generative,he2022masked,bao2022beit} randomly mask a portion of input data and then recover the masked content. In this way, the model is enabled to represent the inherent characteristics within the data. Thus, we are motivated to exploit the self-supervised pre-training method for 3D human pose estimation.

In this paper, we propose a Pre-trained Spatial Temporal Many-to-One (P-STMO) model for 2D-to-3D human pose estimation. The whole process is split into two stages: pre-training (Stage \RNum{1}) and fine-tuning (Stage \RNum{2}). In Stage \RNum{1}, a self-supervised spatial temporal pre-training task, called \emph{masked pose modeling} (MPM), is constructed. We randomly mask some frames\footnote{If not specified, ``frame" in the following text refers to the 2D pose in this frame.} (temporally) as well as some 2D joints (spatially) in the remaining frames. The model in this stage, a general form of denoising auto-encoder~\cite{vincent2008extracting}, is intended to reconstruct the corrupted 2D poses. This gives the network a favorable initialization. In Stage \RNum{2}, the pre-trained encoder, combined with a many-to-one frame aggregator, is retrained to predict the 3D pose of the current (middle) frame by using a sequence of 2D poses as input. With this two-stage strategy, the encoder is supposed to capture 2D spatial temporal dependencies in Stage \RNum{1} and extract 3D spatial and temporal features in Stage \RNum{2}. Experimental results show that this strategy can reduce the optimization difficulty of STMO and improve the prediction performance.

The proposed STMO model consists of three modules: $(i)$ spatial encoding module (SEM): capturing spatial information within a single frame, $(ii)$ temporal encoding module (TEM): capturing temporal dependencies between different frames, $(iii)$ many-to-one frame aggregator (MOFA): aggregating information from multiple frames to assist in the prediction of the current pose. These three modules play different roles and are organically linked to each other, which improves the overall performance. Herein, we propose to use an MLP block as the backbone network of SEM. Compared with \emph{fc}~\cite{li2022exploiting} and Transformer~\cite{zheng20213d}, it achieves better performance while having moderate computational complexity. In addition, a temporal downsampling strategy (TDS) is introduced on the input side to reduce data redundancy while enlarging the temporal receptive field.


Our contributions can be summarized as follows:
\begin{itemize}

\item To the best of our knowledge, P-STMO is the first approach that introduces the pre-training technique to 3D human pose estimation. A pre-training task, namely MPM, is proposed in a self-supervised manner to better capture both spatial and temporal dependencies.
\item The proposed STMO model simplifies the responsibility of each module and therefore significantly reduces the optimization difficulty. An MLP block is utilized as an effective spatial feature extractor for SEM. In addition, a temporal downsampling strategy is employed to mitigate the data redundancy problem for TEM.
\item Compared with other approaches, our method achieves state-of-the-art performance on two benchmarks with fewer parameters and smaller computational budgets.

\end{itemize}


\section{Related Work}

\subsection{3D Human Pose Estimation}
Recently, there is a research trend that uses 2D keypoints to regress corresponding 3D joint positions. The advantage is that it is compatible with any existing 2D pose estimation method. Our approach falls under this category. Extensive works~\cite{xu2020deep,wang2014robust,sun2017compositional,martinez2017simple,pavlakos2018ordinal,fang2018learning,mehta2017vnect} have been carried out around an important issue in videos, which is how to exploit the information in spatial and temporal domains. Some works~\cite{sun2017compositional,martinez2017simple,zeng2021learning,fang2018learning} only focus on 3D single-frame pose estimation and ignore temporal dependencies. Other recent approaches~\cite{xu2020deep,zheng20213d,wang2020motion,cai2019exploiting,hossain2018exploiting} explore the way of integrating spatio-temporal information. The following are the shortcomings of these methods. Approaches relying on recurrent neural network (RNN)~\cite{hossain2018exploiting,lee2018propagating} suffer from high computational complexity. Graph convolutional network (GCN)-based methods~\cite{wang2020motion,hu2021conditional} perform graph convolutions on the spatial temporal graph, and predict all poses in the sequence. This diminishes the capability of the network to model the 3D pose in a particular frame. The Transformer-based method~\cite{zheng20213d} predicts the 3D pose of the current (middle) frame by performing a weighted average at the last layer over the features of all frames in the sequence. It ignores the importance of the current pose and its near neighbors.

On the other hand, some works~\cite{pavllo20193d,liu2020attention,chen2021anatomy} focus on the process of many-to-one frame aggregation. Pavllo \textit{et al.}~\cite{pavllo20193d} first propose a temporal convolutional network (TCN) that uses multiple frames to aid the modeling of one frame by progressively reducing the temporal dimension. Since then many methods~\cite{zeng2020srnet,chen2021anatomy,shan2021improving,liu2020attention} have been proposed based on TCN. However, these methods do not explicitly extract spatial and temporal features. Li \textit{et al.}~\cite{li2022exploiting} alleviate this problem by using a vanilla Transformer to capture long-range temporal dependencies in the sequence. But they acquire spatial information by a single \emph{fc} layer, which has insufficient representation capability.

In contrast, we propose a Spatial Temporal Many-to-One (STMO) model. The most important steps in 3D human pose estimation are represented as three modules in STMO: SEM, TEM, and MOFA. Our method clearly delineates the responsibility of each module and promotes the modeling of intrinsic properties.

\subsection{Pre-Training of Transformer}
Transformer~\cite{vaswani2017attention} has become the \emph{de facto} backbone in NLP~\cite{devlin2018bert,brown2020language,radford2019language,conneau2019cross,qiu2020pre} and CV~\cite{carion2020end,dosovitskiy2021an,touvron2021training,han2020survey,chen2021empirical}. Transformer owes its widespread success to the pre-training technique~\cite{yosinski2014transferable,Peters2018DeepCW,devlin2018bert,bao2022beit,yan2022crossloc}. In NLP, Devlin \textit{et al.}~\cite{devlin2018bert} propose a \emph{masked language modeling} (MLM) task, which triggers a wave of research on pre-training. They randomly mask some words and aim to predict these masked words based only on their context. Subsequent developments in CV have followed a similar trajectory to NLP. Some works~\cite{dosovitskiy2021an,he2022masked,bao2022beit} transfer self-supervised pre-trained models to image-based CV tasks, such as classification, object detection, semantic segmentation, etc. They replace the words to be reconstructed in MLM with pixels or discrete visual tokens. The pre-training task in CV is termed as \emph{masked image modeling} (MIM). 

Inspired by these works, we apply the pre-training technique to 3D human pose estimation and propose a \emph{masked pose modeling} (MPM) task similarly. We randomly mask some joints in spatial and temporal domains and try to recover the original 2D poses. Note that METRO~\cite{lin2021end} also masks the input 2D poses, but the goal is to directly regress 3D joint positions rather than recovering the input. This approach is essentially a data augmentation method and therefore different from the proposed pre-training task.

\section{Method}


\subsection{Overview}

\begin{figure}[t]
\centering
\includegraphics[width=0.8\linewidth]{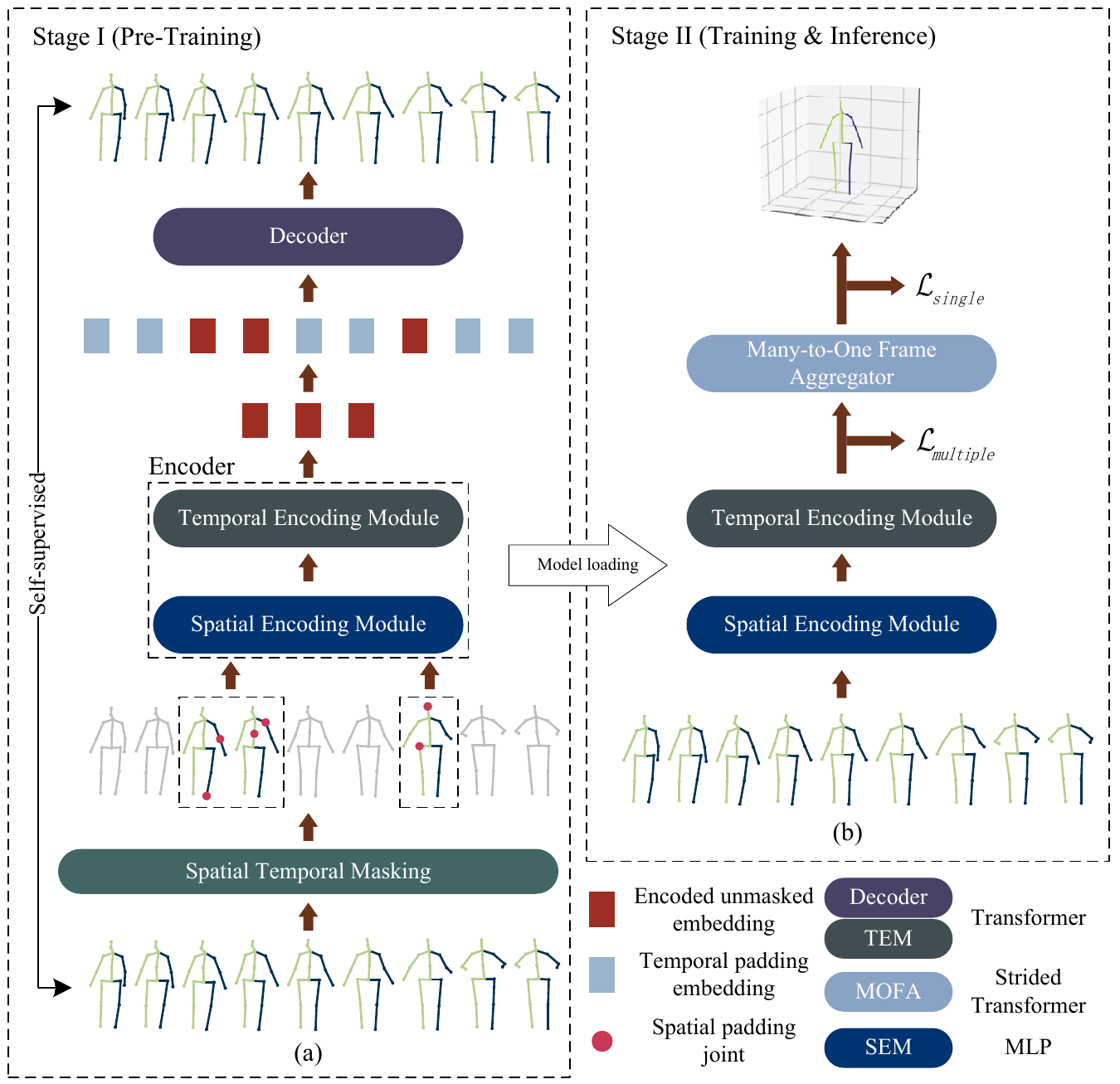}
\vspace{-0.2cm}
\caption{(a) The pre-training procedure for STMO. The 2D pose sequence is randomly masked and fed to the encoder. The encoded unmasked embeddings as well as the temporal padding embeddings are sent to the decoder to reconstruct the original 2D poses in the input sequence. (b) Overview of our STMO model, which consists of SEM, TEM, and MOFA in series.}
\vspace{-0.8cm}
\label{fig:network}
\end{figure}

Fig.~\ref{fig:network} depicts an overview of the proposed P-STMO method, which divides the optimization process into two stages: pre-training and fine-tuning. Firstly, a part of STMO model is pre-trained by solving the \emph{masked pose modeling} (MPM) task in a self-supervised manner. The goal of this stage is to recover the input sequence from the corrupted 2D poses. Secondly, STMO model is fine-tuned to predict the 3D pose in the current (middle) frame given a sequence of 2D poses obtained by an off-the-shelf 2D pose detector. For both stages, we take a sequence of 2D poses as input, denoted as
\begin{equation}
X=\{x_n\}_{n=-(N-1)/2}^{(N-1)/2},\quad x_n=\{p_i\}_{i=1}^J,\label{eq1}
\end{equation}
where $p_i \in \mathbb{R}^2$ is the 2D position of $i^{th}$ joint. $N, J$ are the number of frames and human joints in each frame, respectively. Usually, $N$ is an odd number, which means that the current frame and $(N-1)/2$ frames to the left and right of it are used as inputs.


Stage \RNum{1}: As shown in~Fig.~\ref{fig:network}a, a proportion of frames as well as some joints in the remaining frames are randomly masked. This spatially and temporally masked input is denoted as $X^{\text{ST}}$. The whole network architecture consists of a spatial temporal encoder (SEM+TEM) that maps the masked input $X^{\text{ST}}$ to the latent space, and a decoder that recovers the original 2D poses $X$ from latent representations. To predict the complete 2D poses, the model has to seek relevant unmasked joints for help. In this way, SEM and TEM are enabled to learn 2D spatial and temporal relationships. Since monocular motion contributes to depth estimation~\cite{rogers1979motion}, acquiring 2D spatio-temporal relationships is beneficial for 3D human pose estimation.

Stage \RNum{2}: Fig.~\ref{fig:network}b shows the proposed STMO model. The pre-trained encoder is loaded and fine-tuned in this stage to obtain knowledge about the 3D space. This encoder is followed by MOFA that aggregates multiple frames to estimate the 3D pose in the middle frame of the sequence, denoted as $Y=\{y_0\}, y_0 \in \mathbb{R}^{J \times 3}$.

\subsection{Pre-Training of STMO}
\begin{figure}[t]
\centering
\includegraphics[width=\linewidth]{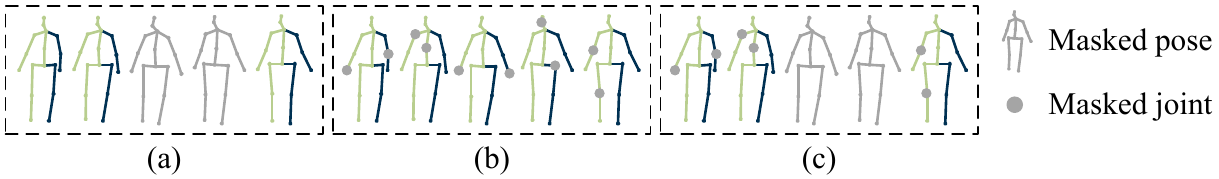}
\vspace{-0.8cm}
\caption{Illustration of three masking strategies. (a) Temporal masking. (b) Spatial masking. (c) Spatial temporal masking. }
\vspace{-0.4cm}
\label{fig:masking}
\end{figure}

The spatial and temporal dependencies of the pose sequence are mainly captured by SEM and TEM in our STMO model. To improve the overall estimation accuracy, we propose a self-supervised spatial temporal pre-training task, namely MPM. As shown in Fig.~\ref{fig:masking}, we use three masking strategies to mask the input 2D poses, which are described in detail below.


\subsubsection{Temporal Masking.} 
A portion of input frames are randomly masked, which is called temporal masking. The masked frame is replaced with a \textit{temporal padding embedding} $e^\text{T}$ which is a shared learnable vector. Similar to~\cite{he2022masked}, to improve the efficiency of the model, we only use the unmasked frames as inputs to the encoder, excluding the temporal padding embeddings. Instead, the decoder takes the temporal padding embeddings as well as the encoded unmasked embeddings as inputs and reconstructs the original 2D poses. In this way, the encoder models temporal dependencies between two unmasked frames that are not adjacent to each other in the original sequence, and then the decoder fills in the missing 2D poses between these two frames. We denote the indices of the masked frames as a set: $\mathcal{M}^{
\text{T}} \subseteq \{-\frac{N-1}{2},-\frac{N-1}{2}+1,\dots, \frac{N-1}{2}\}, \mid \mathcal{M}^{\text{
T}} \mid ={q^\text{T}\cdot N}$, where $q^\text{T} \in \mathbb{R}$ is the temporal masking ratio. We use a large temporal masking ratio (e.g., 90\%), so this task cannot be solved easily by interpolation. The input $X$ to the encoder (eq.~\ref{eq1}) is modified to
\begin{equation}
X^\text{T}=\{x_n^{\text{T}}:n\notin \mathcal{M}^{\text{T}}\}_{n=-(N-1)/2}^{(N-1)/2},\quad x_n^{\text{T}}=\{p_i\}_{i=1}^J.
\end{equation}
After that, the input to the decoder is denoted as
\begin{equation}
\{h_n:n\notin \mathcal{M}^\text{T}\}_{n=-(N-1)/2}^{(N-1)/2} \bigcup \{e^\text{T}:n\in \mathcal{M}^\text{T}\}_{n=-(N-1)/2}^{(N-1)/2},
\end{equation}
where $h_n,e^\text{T} \in \mathbb{R}^d$ are the encoded unmasked embedding and temporal padding embedding respectively. $d$ is the dimension of the latent features. The output is $Y=\{y_n\}_{n=-(N-1)/2}^{(N-1)/2}$, where $y_n \in \mathbb{R}^{J\times2}$ is the recovered 2D pose in frame $n$. 

\subsubsection{Spatial Masking.}
A fixed number of joints in each frame are randomly masked, which is called spatial masking. The masked joint is replaced with a \textit{spatial padding joint} $e^S$ that is a shared learnable vector. The spatial padding joints, together with other unmasked joints, are sent to the encoder. Since the pose sequence is not masked temporally, no temporal padding embedding is used at the decoder side. In such manner, the encoder models spatial dependencies between joints, and then the decoder is capable of recovering the contaminated joints. The indices of the masked joints in frame $n$ are denoted as a set: $\mathcal{M}^{
\text{S}}_n \subseteq \{-\frac{N-1}{2},-\frac{N-1}{2}+1,\dots, \frac{N-1}{2}\}, \mid \mathcal{M}^{
\text{S}}_n \mid ={m^\text{S}}$, where $m^\text{S} \in \mathbb{N}$ is the number of masked joints. The spatial masking ratio is $q^{\text{S}}={m^\text{S}}/{J}$. Although the number of masked joints is the same for each frame, the indices of the masked joints are various in different frames to increase the diversity. The input $X$ to the encoder (eq.~\ref{eq1}) is modified to
\begin{equation}
X^\text{S}=\{x_n^\text{S}\}_{n=-(N-1)/2}^{(N-1)/2},\quad x_n^\text{S}=\{p_i :i\notin \mathcal{M}^\text{S}_n\}_{i=1}^J \bigcup \{e^\text{S} :i\in \mathcal{M}^\text{S}_n\}_{i=1}^J,
\end{equation}
where $e^\text{S} \in \mathbb{R}^2$ is the spatial padding joint. 

\subsubsection{Spatial Temporal Masking.}
To integrate the information on the spatio-temporal domain, we propose a spatial temporal masking strategy, which is a combination of the above two masking methods. Specifically, the temporal masking is implemented on the input pose sequence, followed by the spatial masking on the unmasked frames. This strategy is utilized in the proposed P-STMO model. The total spatial temporal masking ratio is calculated by $q^{\text{ST}}=q^\text{T}+(1-q^\text{T})\cdot q^\text{S}$. The input $X$ to the encoder (eq.~\ref{eq1}) is modified to 
\begin{align}
X^{\text{ST}}&=\{x_n^{\text{ST}}:n\notin \mathcal{M}^\text{T}\}_{n=-(N-1)/2}^{(N-1)/2},\\ x_n^{\text{ST}}&=\{p_i :i\notin \mathcal{M}^\text{S}_n\}_{i=1}^J \bigcup \{e^\text{S} :i\in \mathcal{M}^\text{S}_n\}_{i=1}^J.
\end{align}

\subsection{Spatial Temporal Many-to-One (STMO) Model}
\begin{figure}[t]
\centering
\includegraphics[width=\linewidth]{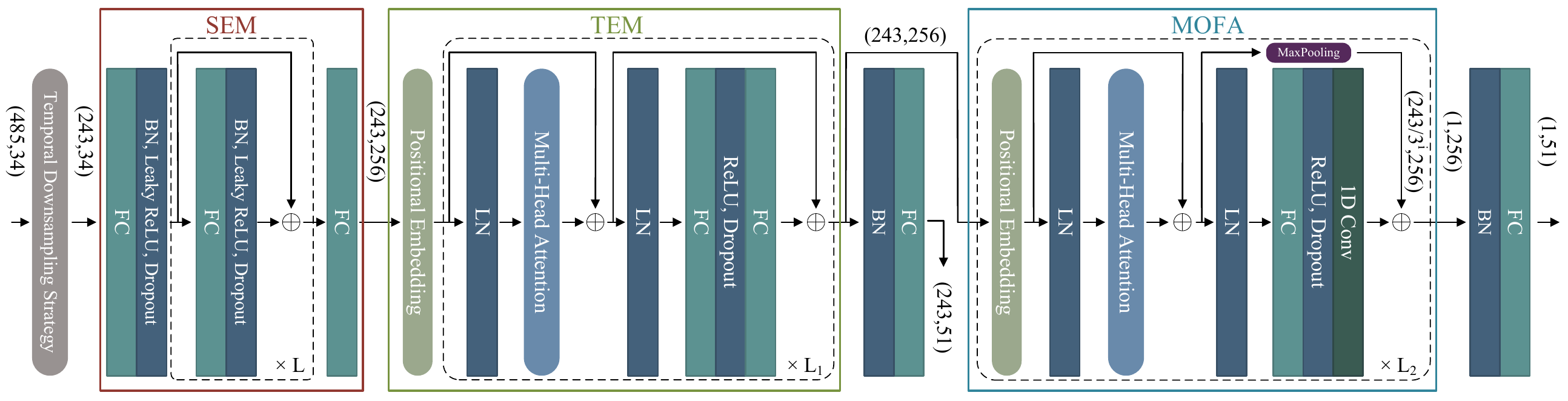}
\vspace{-0.6cm}
\caption{Detailed architecture of the proposed STMO model in Stage \RNum{2}. For $N=243$, we choose 485 frames and downsample them using TDS with the temporal downsampling rate $s$ set to 2. The tensor sizes are shown in parentheses. For example, (243,34) denotes 243 frames and 34 channels, meaning that there are $J=17$ joints in each frame, and each joint has 2 coordinates $(x,y)$. (243/$3^i$,256) means the temporal dimension is reduced by $3^i$ as 1D temporal convolution is used and the stride is the same as the kernel size $M=3$. $i=1,2,...,L_2$, where $L_2=5$ is the number of layers of MOFA. }
\vspace{-0.4cm}
\label{fig:detailed_arc}
\end{figure}

In Stage \RNum{2}, the pre-trained encoder is loaded to the proposed STMO model and fine-tuned on 3D poses. The detailed architecture of STMO is illustrated in Fig.~\ref{fig:detailed_arc}. The input sequence will go through SEM, TEM, and MOFA to obtain the final output. The role of each module is described in detail below.

\subsubsection{Spatial Encoding Module (SEM).}
SEM aims to capture the characteristics of each frame in the spatial domain. Zheng \textit{et al.}~\cite{zheng20213d} propose to use a Transformer~\cite{vaswani2017attention} as the backbone network of SEM to integrate information across all joints in a single frame. However, the self-attention operation brings a great computational overhead, which limits the scalability of the network in the case of using multiple frames as inputs. Therefore, we propose to use a simple MLP block as the backbone network to establish spatial relationships between joints. Compared to~\cite{zheng20213d}, this lightweight design allows the network to accommodate more frames with the same computational budget. Each 2D pose in the input sequence is independently sent to the MLP block whose weights are shared across all frames.

\subsubsection{Temporal Encoding Module (TEM).}
As 2D keypoints are used as inputs to the network, the amount of data per frame is small. Thus, we can take advantage of the extra-long 2D input pose sequence (e.g., 243 frames). Since the objective of this module is to exploit temporal information from the changes of human posture within this sequence, it is inefficient to focus on the relationships between highly redundant frames in a local region. Therefore convolution operations, which introduce the inductive bias of \emph{locality}, can be discarded. Instead, the self-attention mechanism is exploited, which allows the network to easily focus on correlations between frames that are far apart. We use a standard Transformer architecture as the backbone network of TEM. It is capable of capturing non-local self-attending associations. TEM takes a sequence of latent features from SEM as input and treats the features in each frame as an individual token. As the inputs are already embedded, the \emph{token embedding} process in Transformer is skipped.

Previous works~\cite{pavllo20193d,chen2021anatomy} show that the performance of 3D human pose estimation can be improved with more frames as inputs. However, there are two drawbacks when increasing the number of frames. (1) The computational complexity of Transformer is $O(N^2)$, which constrains the network from utilizing a larger scale of temporal information. (2) The input data is highly redundant. For example, the video frame rate of Human3.6M~\cite{ionescu2013human3} is 50 fps, which means that there is only a slight change between adjacent frames. To alleviate these two problems, we propose to use a temporal downsampling strategy (TDS), which uniformly downsamples the input pose sequence. In this way, the network can accommodate a longer time span of information with the same number of input frames. In other words, the receptive field in the temporal domain can be expanded without increasing the number of parameters and computational complexity. Meanwhile, TDS removes highly similar neighboring frames, thus promoting data diversity. After TDS is applied, the input $X$ is modified to $X^{\text{TDS}}=\{x_{n\cdot s}\}_{n=-(N-1)/2}^{(N-1)/2}$, where $s \in \mathbb{N}^+$ is the downsampling rate. The output $Y$ remains unchanged. Some works~\cite{mehta2017vnect,ionescu2014iterated,zhou2016sparseness} also downsample the data, but they reduce the size of the dataset. Since the scale of the dataset is preserved by TDS, the proposed strategy is more effective than these works.

\subsubsection{Many-to-One Frame Aggregator (MOFA).}
TEM mainly concerns the understanding of the temporal dependencies in the overall sequence, while MOFA aims to aggregate information from multiple frames and extract the 3D pose of the current frame. Therefore, it is effective to leverage local information around the current pose, which allows convolution to shine in this module. Li \textit{et al.}~\cite{li2022exploiting} propose a Strided Transformer Encoder (STE), which is a combination of Transformer and convolution. Specifically, STE replaces the MLP block in Transformer with strided 1D temporal convolutions proposed by TCN~\cite{pavllo20193d}. We use STE as the backbone network of MOFA. For the MLP block in vanilla Transformer, the \emph{fc} layer is formulated as $l_{i,n}^{(k)}=\sum_{j=0}^{d_{k-1}}w_{i,j} * l_{j,n}^{(k-1)}$, where $l_{i,n}^{(k)}$ is the $i^{th}$ channel of $n^{th}$ token (frame) in $k^{th}$ layer, $w_{i,j}$ is the weight shared across frames, and $d_{k-1}$ is the number of channels in $(k-1)^{th}$ layer. In the case that 1D temporal convolution is used instead of the \emph{fc} layer, the formulation is modified to $l_{i,n}^{(k)}=\sum_{j=0}^{d_{k-1}}\sum_{m=-(M-1)/2}^{(M-1)/2}w_{i,j,m} * l_{j,n+m}^{(k-1)}$, where $M$ is the kernel size. It aggregates adjacent $M$ frames in the sequence into a single frame. The stride is set to $M$ so that there will be no overlap between any two convolution operations. The stacking of multiple layers of 1D temporal convolution eventually enables the mapping from $N$ frames to one frame (current frame).

\subsection{Loss Function}

\subsubsection{Stage \RNum{1}.}
The objective of the pre-training stage is to minimize the negative log-likelihood of the correct 2D pose $x_n=\{p_i\}_{i=1}^J$ in each frame given the corrupted input sequence: 
\begin{equation}
\mathcal{L}_{pretrain}= \sum_{x \in \mathcal{D}} \mathbb{E}_{n}\left[\sum_{n=-(N-1)/2}^{(N-1)/2} -\log p\left(x_{n} \mid X^{\text{ST}}\right)\right],
\end{equation}
where $\mathcal{D}$ is the training corpus, and $p(\cdot \mid \cdot)$ is the conditional probability.

\subsubsection{Stage \RNum{2}.}

The final estimated 3D pose of the middle frame in the sequence is obtained via a linear projection after MOFA. To optimize the network, we use L2 loss to minimize the errors between predictions and ground truths : 
\begin{equation}
\mathcal{L}_{single}=\frac{1}{J} \sum_{i=1}^{J}\left\|y_{i}-\widetilde{y}_{i}\right\|_{2},
\end{equation}
where $y_{i}$ and $\widetilde{y}_{i}$ are the ground truth and estimated 3D positions of $i^{th}$ joint in the current pose respectively.

In addition, to better supervise the optimization, we append a multi-frame loss~\cite{li2022exploiting} after TEM. To be specific, a linear projection is added to obtain the 3D poses of all frames in the sequence. In this way, we force TEM to exploit the temporal relationships between frames to predict 3D poses in all frames. Since the output features of TEM are quite close to 3D poses after applying the multi-frame loss, the role of MOFA is to aggregate these 3D pose-related features of all frames into a single representation. The multi-frame loss is formulated as: 
\begin{equation}
\mathcal{L}_{multiple}=\frac{1}{J\cdot N} \sum_{i=1}^{J}\sum_{n=-(N-1)/2}^{(N-1)/2}\left\|y_{i,n}-\widetilde{y}_{i,n}\right\|_{2}.
\end{equation}
The final loss function of the whole network is $\mathcal{L} = \mathcal{L}_{single}+\lambda \mathcal{L}_{multiple}$, where $\lambda$ is the balance factor.

\section{Experiments}
\subsection{Datasets and Evaluation Metrics}

\subsubsection{Human3.6M~\cite{ionescu2013human3}} is the most commonly used indoor dataset, which consists of 3.6 million frames captured by four 50 Hz cameras. The dataset is divided into 15 daily activities (e.g., walking and sitting) performed by 11 subjects. Following~\cite{pavllo20193d,shan2021improving,cai2019exploiting}, we use 5 subjects for training (S1, S5, S6, S7, S8), and 2 subjects for testing (S9, S11). 

We report the mean per joint position error (MPJPE) and Procrustes MPJPE (P-MPJPE) for Human3.6M dataset. The former computes the Euclidean distance between the predicted joint positions and the ground truth positions. The latter is the MPJPE after the predicted results align to the ground truth via a rigid transformation.

\subsubsection{MPI-INF-3DHP~\cite{mehta2017monocular}} is a more challenging dataset with both indoor and outdoor scenes. The training set comprises of 8 subjects, covering 8 activities. The test set covers 7 activities, containing three scenes: green screen, non-green screen, and outdoor.  Following~\cite{zheng20213d,chen2021anatomy}, we train the proposed method using all activities from 8 camera views in the training set and evaluate on valid frames in the test set.

We report MPJPE, percentage of correct keypoints (PCK) within 150mm range, and area under curve (AUC) as the evaluation metrics for MPI-INF-3DHP dataset.

\subsection{Comparison with State-of-the-Art Methods}

\subsubsection{Results on Human3.6M.}
We compare our P-STMO with existing state-of-the-art methods on Human3.6M dataset. Following~\cite{zheng20213d,li2022exploiting,pavllo20193d,liu2020attention}, we use CPN~\cite{chen2018cascaded} as the 2D keypoint detector, and then train our networks on the detected 2D pose sequence. As shown in Table~\ref{tab:results} (top and middle), our method achieves promising results under both MPJPE (42.8mm) and P-MPJPE (34.4mm) when using 243 frames as inputs. We also train our model using the same refining module as~\cite{li2022exploiting,cai2019exploiting,wang2020motion}. It achieves 42.1mm under MPJPE, which surpasses all other methods. Our method yields better performance on hard poses (such as \emph{Photo} and \emph{SittingDown}) than the previous works. This demonstrates the robustness of our model in the case of depth ambiguity and severe self-occlusion. Additionally, we propose a smaller model P-STMO-S\emph{mall}. The only difference between P-STMO and P-STMO-S lies in the number of Transformer layers. For more details, please refer to the supplementary materials. Compared with P-STMO, P-STMO-S achieves 43.0mm MPJPE with a smaller number of FLOPs and parameters. To explore the lower bound of the proposed method, we utilize the ground truth of 2D poses as inputs. In this way, the input noise is removed. As shown in Table~\ref{tab:results} (bottom), P-STMO outperforms other methods with 29.3mm under MPJPE.

\setlength{\tabcolsep}{2pt}
\begin{table}[t]\small
\begin{center}
\caption{Results on Human3.6M in millimeter under MPJPE and P-MPJPE. Top\&Middle table: 2D poses detected by CPN are used as inputs. Bottom table: the ground truth of 2D poses are used as inputs. $N$ is the number of input frames. (*) - uses the refining module proposed in~\cite{cai2019exploiting}. The best result is shown in bold, and the second-best result is underlined.}
\vspace{-0.5cm}
\label{tab:results}
\resizebox{\columnwidth}{!}{\begin{tabular}{c|ccccccccccccccc|c}
\hline
\noalign{\smallskip}
MPJPE (CPN) & Dir. & Disc. & Eat & Greet & Phone & Photo & Pose & Pur. & Sit & SitD. & Smoke & Wait & WalkD. & Walk & WalkT. & Avg \\
\noalign{\smallskip}
\hline
\noalign{\smallskip}
Pavllo \textit{et al.}~\cite{pavllo20193d} CVPR'19 ($N$=243)& 45.2&46.7&43.3&45.6&48.1&55.1&44.6&44.3&57.3&65.8&47.1&44.0&49.0&32.8&33.9&46.8 \\
Lin \textit{et al.}~\cite{lin2019trajectory} BMVC'19 ($N$=50)& 42.5&44.8&42.6&44.2&48.5&57.1&42.6&41.4&56.5&64.5&47.4&43.0&48.1&33.0&35.1&46.6 \\
Xu \textit{et al.}~\cite{xu2020deep} CVPR'20 ($N$=9)
&\textbf{37.4}&43.5&42.7&42.7&46.6&59.7&41.3&45.1&\textbf{52.7}&60.2&45.8&43.1&47.7&33.7&37.1&45.6\\
Wang \textit{et al.}~\cite{wang2020motion} ECCV'20 ($N$=96)
&41.3&43.9&44.0&42.2&48.0&57.1&42.2&43.2&57.3&61.3&47.0&43.5&47.0&32.6&31.8&45.6\\
Liu \textit{et al.}~\cite{liu2020attention} CVPR'20 ($N$=243)& 41.8&44.8&41.1&44.9&47.4&54.1&43.4&42.2&56.2&63.6&45.3&43.5&45.3&31.3&32.2&45.1 \\
Zeng \textit{et al.}~\cite{zeng2020srnet} ECCV'20 ($N$=243)& 46.6&47.1&43.9&41.6&45.8&\underline{49.6}&46.5&40.0&53.4&61.1&46.1&42.6&43.1&31.5&32.6&44.8 \\
Zeng \textit{et al.}~\cite{zeng2021learning} ICCV'21 ($N$=9)
&-&-&-&-&-&-&-&-&-&-&-&-&-&-&-&45.7\\
Zheng \textit{et al.}~\cite{zheng20213d} ICCV'21 ($N$=81)&
41.5&44.8&39.8&42.5&46.5&51.6&42.1&42.0&\underline{53.3}&60.7&45.5&43.3&46.1&31.8&32.2&44.3\\
Shan \textit{et al.}~\cite{shan2021improving} MM'21 ($N$=243)&
40.8&44.5&41.4&42.7&46.3&55.6&41.8&41.9&53.7&60.8&45.0&\textbf{41.5}&44.8&30.8&31.9&44.3\\
Chen \textit{et al.}~\cite{chen2021anatomy} TCSVT'21 ($N$=243)&
41.4&43.5&40.1&42.9&46.6&51.9&41.7&42.3&53.9&60.2&45.4&41.7&46.0&31.5&32.7&44.1\\
Hu \textit{et al.}~\cite{hu2021conditional} MM'21 ($N$=96)&
\underline{38.0}&43.3&\underline{39.1}&\textbf{39.4}&45.8&53.6&41.4&41.4&55.5&61.9&44.6&41.9&44.5&31.6&\underline{29.4}&43.4\\
Li \textit{et al.}~\cite{li2022exploiting} TMM'22 ($N$=351)(*)&
39.9&43.4&40.0&40.9&46.4&50.6&42.1&\underline{39.8}&55.8&61.6&44.9&43.3&44.9&29.9&30.3&43.6\\
\noalign{\smallskip}
\hline
\noalign{\smallskip}
P-STMO-S ($N$=81)  &
41.7&44.5&41.0&42.9&46.0&51.3&42.8&41.3&54.9&61.8&45.1&42.8&43.8&30.8&30.7&44.1\\
P-STMO-S ($N$=243)&
40.0&\underline{42.5}&\textbf{38.3}&41.5&45.8&50.8&41.6&40.9&54.2&\underline{59.3}&\underline{44.4}&41.9&43.6&30.3&30.1&43.0\\
P-STMO ($N$=243)&
38.9&42.7&40.4&41.1&\underline{45.6}&49.7&\underline{40.9}&39.9&55.5&59.4&44.9&42.2&\underline{42.7}&\underline{29.4}&\underline{29.4}&\underline{42.8}\\
P-STMO ($N$=243)(*)  &
38.4&\textbf{42.1}&39.8&\underline{40.2}&\textbf{45.2}&\textbf{48.9}&\textbf{40.4}&\textbf{38.3}&53.8&\textbf{57.3}&\textbf{43.9}&\underline{41.6}&\textbf{42.2}&\textbf{29.3}&\textbf{29.3}&\textbf{42.1}\\
\noalign{\smallskip}
\hline
\hline
\noalign{\smallskip}
P-MPJPE (CPN) & Dir. & Disc. & Eat & Greet & Phone & Photo & Pose & Pur. & Sit & SitD. & Smoke & Wait & WalkD. & Walk & WalkT. & Avg \\
\noalign{\smallskip}
\hline
\noalign{\smallskip}
Lin \textit{et al.}~\cite{lin2019trajectory} BMVC'19 ($N$=50)&
32.5&35.3&34.3&36.2&37.8&43.0&33.0&32.2&45.7&51.8&38.4&32.8&37.5&25.8&28.9&36.8 \\
Pavllo \textit{et al.}~\cite{pavllo20193d} CVPR'19 ($N$=243)& 34.1&36.1&34.4&37.2&36.4&42.2&34.4&33.6&45.0&52.5&37.4&33.8&37.8&25.6&27.3&36.5 \\
Xu \textit{et al.}~\cite{xu2020deep} CVPR'20 ($N$=9)
&\textbf{31.0}&\textbf{34.8}&34.7&\underline{34.4}&36.2&43.9&\underline{31.6}&33.5&\textbf{42.3}&49.0&37.1&33.0&39.1&26.9&31.9&36.2\\
Liu \textit{et al.}~\cite{liu2020attention} CVPR'20 ($N$=243)& 32.3&\underline{35.2}&33.3&35.8&35.9&41.5&33.2&32.7&44.6&50.9&37.0&\underline{32.4}&37.0&25.2&27.2&35.6 \\
Wang \textit{et al.}~\cite{wang2020motion} ECCV'20 ($N$=96)
&32.9&\underline{35.2}&35.6&\underline{34.4}&36.4&42.7&\textbf{31.2}&32.5&45.6&50.2&37.3&32.8&36.3&26.0&\textbf{23.9}&35.5\\
Chen \textit{et al.}~\cite{chen2021anatomy} TCSVT'21 ($N$=243)&
33.1&35.3&33.4&35.9&36.1&41.7&32.8&33.3&\underline{42.6}&49.4&37.0&32.7&36.5&25.5&27.9&35.6\\
Shan \textit{et al.}~\cite{shan2021improving} MM'21 ($N$=243)&
32.5&36.2&33.2&35.3&35.6&42.1&32.6&\underline{31.9}&\underline{42.6}&\textbf{47.9}&36.6&\textbf{32.1}&\underline{34.8}&\underline{24.2}&\underline{25.8}&35.0\\
Zheng \textit{et al.}~\cite{zheng20213d} ICCV'21 ($N$=81)&
32.5&\textbf{34.8}&\textbf{32.6}&34.6&\textbf{35.3}&\underline{39.5}&32.1&32.0&42.8&48.5&\textbf{34.8}&\underline{32.4}&35.3&24.5&26.0&\underline{34.6}\\

\noalign{\smallskip}
\hline
\noalign{\smallskip}
P-STMO ($N$=243)  &
\underline{31.3}&\underline{35.2}&\underline{32.9}&\textbf{33.9}&\underline{35.4}&\textbf{39.3}&32.5&\textbf{31.5}&44.6&\underline{48.2}&\underline{36.3}&32.9&\textbf{34.4}&\textbf{23.8}&\textbf{23.9}&\textbf{34.4}\\
\noalign{\smallskip}
\hline
\hline
\noalign{\smallskip}
MPJPE (GT) & Dir. & Disc. & Eat & Greet & Phone & Photo & Pose & Pur. & Sit & SitD. & Smoke & Wait & WalkD. & Walk & WalkT. & Avg \\
\noalign{\smallskip}
\hline
\noalign{\smallskip}
Pavllo \textit{et al.}~\cite{pavllo20193d} CVPR'19 ($N$=243) &35.2&40.2&32.7&35.7&38.2&45.5&40.6&36.1&48.8&47.3&37.8&39.7&38.7&27.8&29.5&37.8  \\
Lin \textit{et al.}~\cite{lin2019trajectory} BMVC'19 ($N$=50)& -&-&-&-&-&-&-&-&-&-&-&-&-&-&-&32.8 \\
Liu \textit{et al.}~\cite{liu2020attention} CVPR'20 ($N$=243) &34.5&37.1&33.6&34.2&32.9&37.1&39.6&35.8&40.7&41.4&33.0&33.8&33.0&26.6&26.9&34.7   \\
Zeng \textit{et al.}~\cite{zeng2020srnet} ECCV'20 ($N$=243) &34.8&32.1&\textbf{28.5}&30.7&31.4&36.9&35.6&\underline{30.5}&38.9&40.5&32.5&31.0&29.9&\underline{22.5}&24.5&32.0   \\
Chen \textit{et al.}~\cite{chen2021anatomy} TCSVT'21 ($N$=243)&-&-&-&-&-&-&-&-&-&-&-&-&-&-&-&32.3\\
Zheng \textit{et al.}~\cite{zheng20213d} ICCV'21 ($N$=81)&
30.0&33.6&29.9&31.0&\underline{30.2}&\underline{33.3}&34.8&31.4&37.8&38.6&31.7&31.5&29.0&23.3&\underline{23.1}&31.3\\
Shan \textit{et al.}~\cite{shan2021improving} MM'21 ($N$=243)&
\underline{29.5}&\underline{30.8}&28.8&\underline{29.1}&30.7&35.2&\underline{31.7}&\textbf{27.8}&\textbf{34.5}&\textbf{36.0}&\underline{30.3}&\textbf{29.4}&\underline{28.9}&24.1&24.7&\underline{30.1}\\
\noalign{\smallskip}
\hline
\noalign{\smallskip}
P-STMO ($N$=243)  &
\textbf{28.5}&\textbf{30.1}&\underline{28.6}&\textbf{27.9}&\textbf{29.8}&\textbf{33.2}&\textbf{31.3}&\textbf{27.8}&\underline{36.0}&\underline{37.4}&\textbf{29.7}&\underline{29.5}&\textbf{28.1}&\textbf{21.0}&\textbf{21.0}&\textbf{29.3}\\
\noalign{\smallskip}
\hline
\end{tabular}}
\end{center}
\vspace{-0.8cm}
\end{table}

\subsubsection{Results on MPI-INF-3DHP.}
Table~\ref{tab:3dhp} reports the performance of state-of-the-art methods and the proposed method on MPI-INF-3DHP dataset. Following~\cite{hu2021conditional,wang2020motion,zheng20213d}, we adopt the ground truth of 2D poses as inputs. Since the sequence length in this dataset is shorter than that in Human3.6M dataset, we set the number of input frames to 81. Our method achieves significant improvements in terms of AUC (9.1\%) and MPJPE (24.2\%). The PCK metric is at a competitive level compared to~\cite{hu2021conditional}. The results suggest that our method has a strong generalization ability.

\subsubsection{Qualitative Results.}
\begin{figure}[t]
\centering
\includegraphics[width=\linewidth]{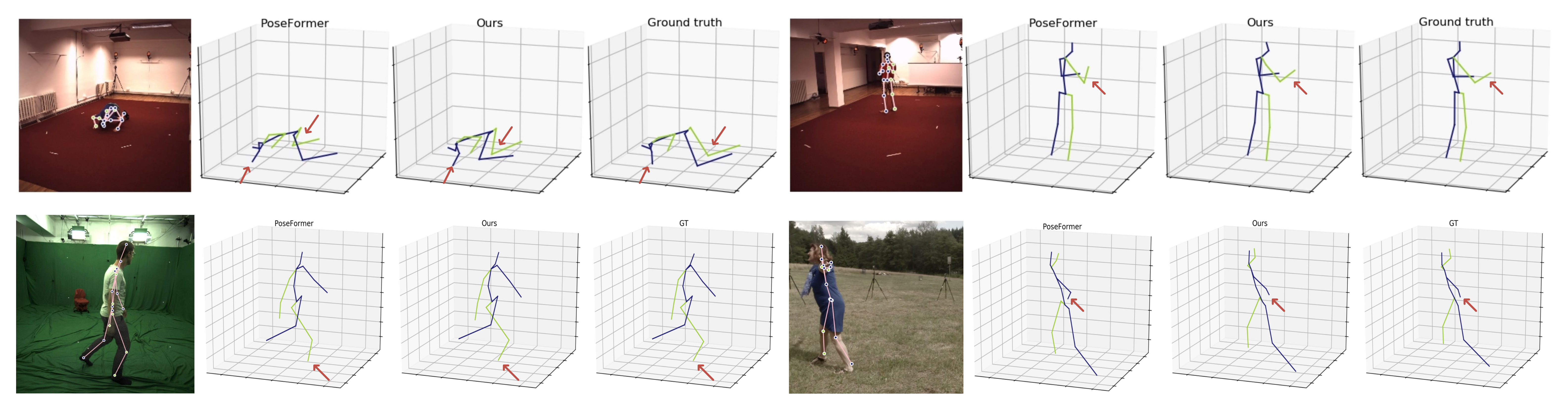}
\vspace{-0.7cm}
\caption{Qualitative comparison between our P-STMO method and Poseformer~\cite{zheng20213d} on Human3.6M and MPI-INF-3DHP datasets.}
\vspace{-0.5cm}
\label{fig:qualitative}
\end{figure}

We provide some qualitative results in Fig.~\ref{fig:qualitative}. We compare the proposed method with Poseformer~\cite{zheng20213d} on Human3.6M and MPI-INF-3DHP datasets. Our method achieves good qualitative performance in both indoor scenes and complex outdoor scenes.



\begin{table}[t]\scriptsize
\begin{center}
\caption{Results on MPI-INF-3DHP under three evaluation metrics.}
\vspace{-0.1cm}
\label{tab:3dhp}
\begin{tabular}{c|ccc}
\hline\noalign{\smallskip}
Method&PCK$\uparrow$&AUC$\uparrow$&MPJPE$\downarrow$\\
\noalign{\smallskip}
\hline
\noalign{\smallskip}
Mehta \textit{et al.}~\cite{mehta2017monocular} 3DV'17 ($N$=1)&75.7&39.3&117.6\\
Pavllo \textit{et al.}~\cite{pavllo20193d} CVPR'19 ($N$=81)&86.0&51.9&84.0\\
Lin \textit{et al.}~\cite{lin2019trajectory} BMVC'19 ($N$=25)&83.6&51.4&79.8\\
Zeng \textit{et al.}~\cite{zeng2020srnet} ECCV'20 ($N$=1)&77.6&43.8&-\\
Wang \textit{et al.}~\cite{wang2020motion} ECCV'20 ($N$=96)&86.9&62.1&68.1\\
Zheng \textit{et al.}~\cite{zheng20213d} ICCV'21 ($N$=9)&\underline{88.6}&56.4&77.1\\
Chen \textit{et al.}~\cite{chen2021anatomy} TCSVT'21 ($N$=81)&87.9&54.0&78.8\\
Hu \textit{et al.}~\cite{hu2021conditional} MM'21 ($N$=96)&\textbf{97.9}&\underline{69.5}&\underline{42.5}\\
\noalign{\smallskip}
\hline
\noalign{\smallskip}
P-STMO ($N$=81)&\textbf{97.9}&\textbf{75.8}&\textbf{32.2}\\
\noalign{\smallskip}
\hline
\end{tabular}
\end{center}
\vspace{-0.6cm}
\end{table}

\subsection{Ablation Study}
To verify the effectiveness of each component, we conduct ablation experiments on Human3.6M dataset using our P-STMO-S model. We utilize 2D keypoints from CPN as inputs. MPJPE is reported for analysis.

\subsubsection{Impact of Each Component.} 
As shown in Table~\ref{table:component}, we validate the contributions of different modules in STMO model and the overall performance gain brought by the proposed MPM and TDS methods. A network consisting of only SEM achieves 51.0 mm under MPJPE. To evaluate the effect of TEM, we use a vanilla Transformer as the backbone and yield a result of 49.6mm. After combining SEM and TEM, MPJPE drops to 46.0mm. Finally, we end up with the proposed STMO model by assembling SEM, TEM as well as MOFA, and achieve 44.2mm under MPJPE. Furthermore, MPM and TDS improve upon STMO by 1.0mm and 0.2mm respectively. Previous works~\cite{pavllo20193d,liu2020attention} observe decreasing returns when increasing the temporal receptive field, so a small gain (0.2mm) from TDS is to be expected in the case of 243 input frames.

\begin{table}[t]
\centering
\begin{minipage}[t]{.5\linewidth}
\centering
\caption{The effectiveness of different components.}
\vspace{-0.2cm}
\label{table:component}
\resizebox{\linewidth}{!}{\begin{tabular}{ccccc@{\hskip 0.1in}|@{\hskip 0.1in}ccc}
\hline\noalign{\smallskip}
SEM&TEM&MOFA&MPM&TDS&Params(M)&FLOPs(M)&MPJPE$\downarrow$\\
\noalign{\smallskip}
\hline
\noalign{\smallskip}
\Checkmark&\XSolidBrush&\XSolidBrush&\XSolidBrush&\XSolidBrush&1.1&536&51.0\\
\XSolidBrush&\Checkmark&\XSolidBrush&\XSolidBrush&\XSolidBrush&1.6&769&49.6\\
\Checkmark&\Checkmark&\XSolidBrush&\XSolidBrush&\XSolidBrush&2.2&1094&46.0\\
\Checkmark&\Checkmark&\Checkmark&\XSolidBrush&\XSolidBrush&6.2&1482&44.2\\
\Checkmark&\Checkmark&\Checkmark&\Checkmark&\XSolidBrush&6.2&1482&43.2\\
\Checkmark&\Checkmark&\Checkmark&\Checkmark&\Checkmark&6.2&1482&\textbf{43.0}\\
\noalign{\smallskip}
\hline
\end{tabular}}
\end{minipage}
\hfill
\begin{minipage}[t]{.47\linewidth}
\centering
\caption{Analysis on designs of SEM. $L$ is the number of sub-blocks in MLP.}
\vspace{-0.2cm}
\label{table:SEM}
\centering
\resizebox{\linewidth}{!}{\begin{tabular}{c|c|ccc}
\hline\noalign{\smallskip}
Backbone&$N$&Params(M)&FLOPs(M)&MPJPE$\downarrow$\\
\noalign{\smallskip}
\hline
\noalign{\smallskip}
\emph{fc}&243&1.6&769&49.6\\
Transformer&81&7.2&1218&48.8\\
MLP ($L=1$)&243&2.3&1094&\underline{46.0}\\
MLP ($L=2$)&243&2.8&1350&\textbf{45.9}\\
MLP ($L=3$)&243&3.3&1608&46.1\\
\noalign{\smallskip}
\hline
\end{tabular}}
\end{minipage}%
\vspace{-0.6cm}
\end{table}

\begin{figure}[t]
\centering
\scriptsize
\subfigure[]{
\begin{minipage}[t]{0.31\linewidth}
\includegraphics[width=\linewidth]{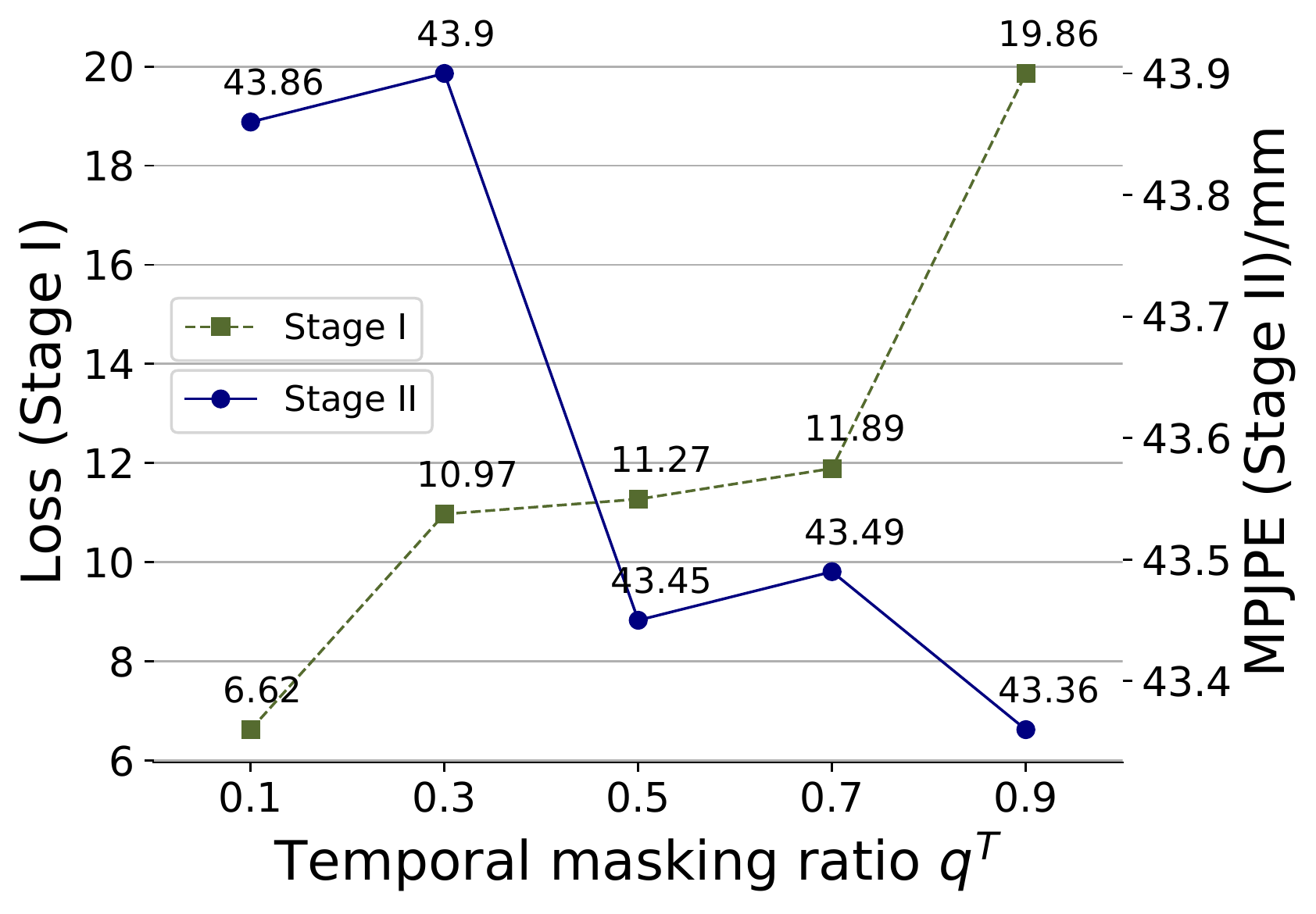}
\end{minipage}}
\subfigure[]{
\begin{minipage}[t]{0.31\linewidth}
\includegraphics[width=\linewidth]{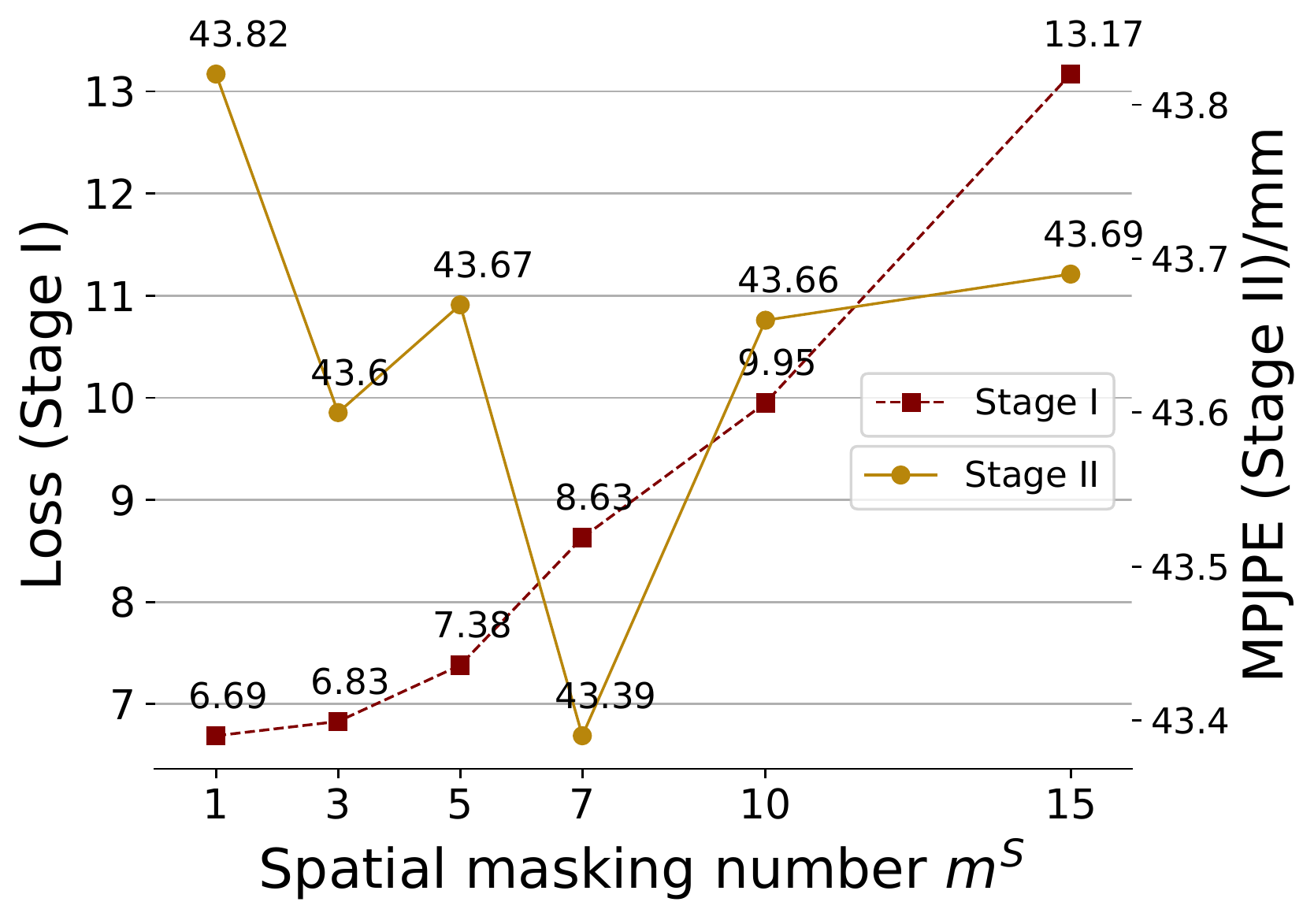}
\end{minipage}}
\subfigure[]{
\begin{minipage}[t]{0.31\linewidth}
\includegraphics[width=\linewidth]{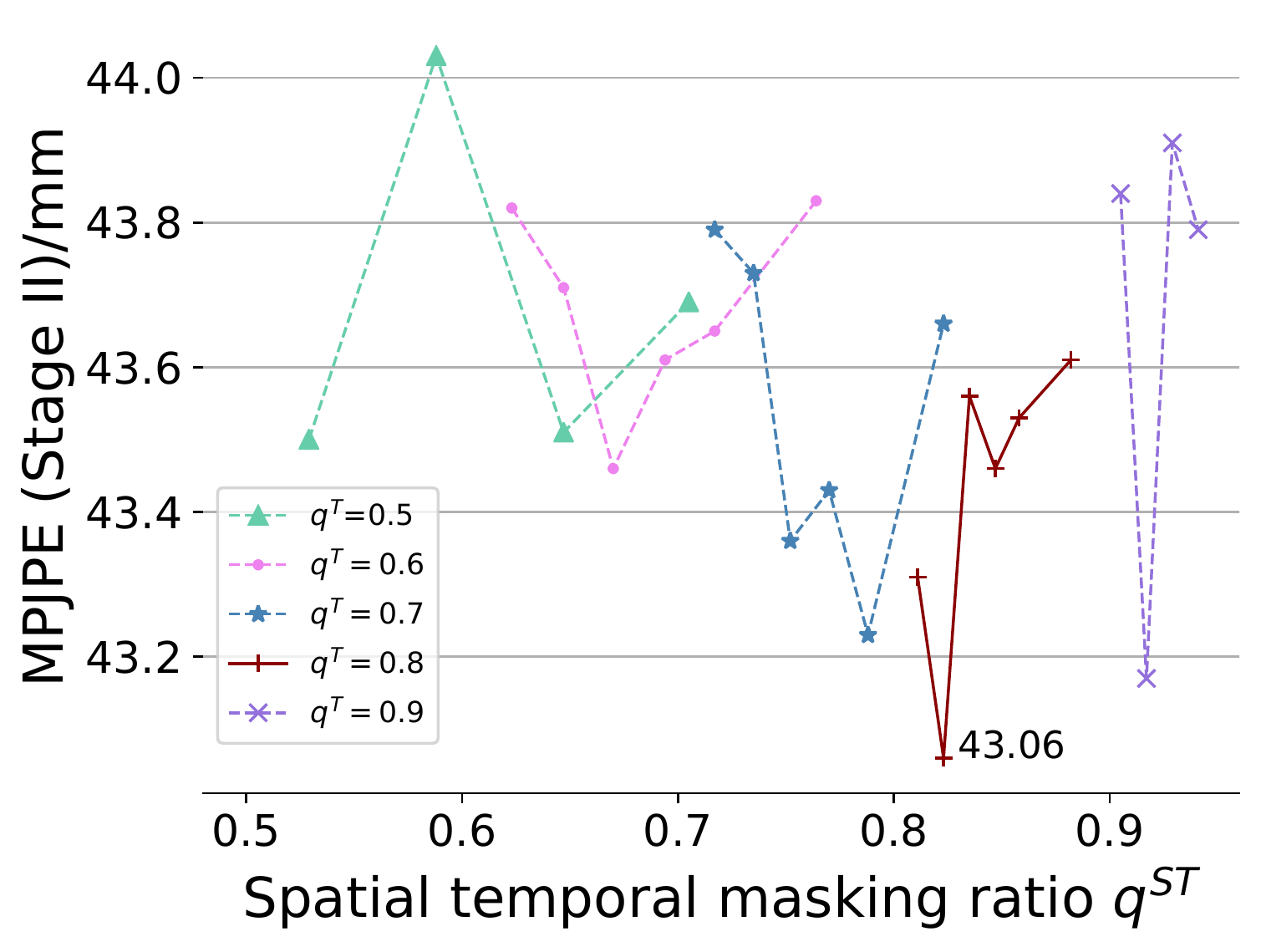}
\end{minipage}}
\vspace{-0.5cm}
\caption{Performance of three different masking strategies. (a) Temporal masking. (b) Spatial masking. (c) Spatial temporal masking.}
\vspace{0cm}
\label{fig:st_mask}
\end{figure}

\subsubsection{Analysis on Masking Ratio of Pre-Training.} Fig.~\ref{fig:st_mask} shows the influence of the masking ratio of three masking strategies in the pre-training stage. As shown in Fig.~\ref{fig:st_mask}a, the optimal ratio is $q^\text{T}=90\%$ when only temporal masking is used. This ratio is higher than MAE~\cite{he2022masked} in CV and BERT~\cite{devlin2018bert} in NLP, whose masking ratios are 75\% and 15\%, respectively. This is because 2D pose sequences are more redundant than images and sentences. Since two adjacent poses are very similar, we need a high masking ratio to increase the difficulty of the task. We observe a negative correlation between the loss of Stage \RNum{1} and the error of Stage \RNum{2}. This implies that increasing the difficulty of the MPM task in Stage \RNum{1} does promote the fitting ability of the encoder, thus aiding the training of Stage \RNum{2}.

The way of spatial masking is different from that of temporal masking. In the case of temporal masking, the decoder perceives the indices of the masked frames through positional embeddings. In the case of spatial masking, only the latent features in each frame are visible to the decoder, so the indices of the masked joints cannot be inferred directly. For the above reasons, it is clear that the spatial masking task is more troublesome than the temporal masking task. Therefore, the optimal masking ratio of the former should be smaller than 
that of the latter. Fig.~\ref{fig:st_mask}b shows that the optimal number is $m^\text{S}=7$ ($q^\text{S}=41.1\%$) when only spatial masking is used. When the masking number is greater than 7, continuing to increase the difficulty of the task will result in too much noise in the input data. Subsequently, the encoder is unable to obtain useful information.

We illustrate the effect of combining temporal masking and spatial masking in Fig.~\ref{fig:st_mask}c. Since this hybrid masking strategy is more challenging, the spatial masking ratio and temporal masking ratio should be slightly reduced. The best performance is achieved when $q^\text{T}=80\%$ and $m^\text{S}=2$ ($q^\text{S}=11.7\%$). The optimal spatial temporal masking ratio is $q^{\text{ST}}=82.3\%$. The performance improvement of spatial temporal masking over spatial masking and temporal masking is 0.33mm and 0.3mm, respectively.

\subsubsection{Analysis on Different Designs of SEM.} 
To verify the superiority of using an MLP block as the backbone network of SEM, we evaluate different architecture designs in Table~\ref{table:SEM}. The results show that MLP outperforms \emph{fc} and Transformer by 3.7mm and 2.9mm respectively. The structure of a single \emph{fc} layer is so simple that it does not have sufficient capability to represent spatial characteristics. Meanwhile, Transformer is too complex and thus difficult to optimize. Therefore, they are not as effective as an MLP block. MLP achieves better performance than the other two methods, while its parameters and FLOPs are in between. Additionally, we explore the influence of the number of sub-blocks $L$ in MLP. The best performance is achieved when $L=2$. We choose $L=1$ in other experiments because it is less computationally intensive and achieves similar performance compared to the best one. 

\begin{table}[t]
\centering
\begin{minipage}[t]{.4\linewidth}
\centering
\caption{Analysis on temporal downsampling rate $s$. RF is the temporal receptive field.}
\vspace{-0.2cm}
\label{table:tds}
\resizebox{\linewidth}{!}{\begin{tabular}{ccc@{\hskip 0.1in}|@{\hskip 0.1in}ccc}
\hline\noalign{\smallskip}
$N$&$s$&RF&Params(M)&FLOPs(M)&MPJPE$\downarrow$\\
\noalign{\smallskip}
\hline
\noalign{\smallskip}
27&1&27&4.6&163&48.2\\
27&3&81&4.6&163&46.8\\
27&9&243&4.6&163&46.1\\
81&1&81&5.4&493&45.6\\
81&3&243&5.4&493&44.1\\
243&1&243&6.2&1482&43.2\\
243&2&486&6.2&1482&\textbf{43.0}\\

\noalign{\smallskip}
\hline
\end{tabular}}
\end{minipage}%
\hfill
\begin{minipage}[t]{.56\linewidth}
\centering
\caption{Analysis on computational complexity. Top table: taken from~\cite{zheng20213d}. Bottom table: our implementation. }
\vspace{-0.2cm}
\label{tab:flops}
\centering
\resizebox{\linewidth}{!}{\begin{tabular}{c|cccc}
\hline\noalign{\smallskip}
Method&Params(M)&FLOPs(M)&FPS&MPJPE$\downarrow$\\
\noalign{\smallskip}
\hline
\noalign{\smallskip}
Pavllo \textit{et al.}~\cite{pavllo20193d} CVPR'19 ($N$=243)&16.9&33&863&46.8\\
Chen \textit{et al.}~\cite{chen2021anatomy} TCSVT'21 ($N$=243)&58.1&116&264&44.1\\
Zheng \textit{et al.}~\cite{zheng20213d} ICCV'21 ($N$=81)&9.5&1358&269&44.3\\
\noalign{\smallskip}
\hline
\noalign{\smallskip}
Chen \textit{et al.}~\cite{chen2021anatomy} TCSVT'21 ($N$=243)&58.1&656&429&44.1\\
Zheng \textit{et al.}~\cite{zheng20213d} ICCV'21 ($N$=81)&9.5&1624&1952&44.3\\
P-STMO-S ($N$=81)&5.4&493&\textbf{7824}&44.1\\
P-STMO-S ($N$=243)&6.2&1482&\underline{3504}&\underline{43.0}\\
P-STMO ($N$=243)&6.7&1737&3040&\textbf{42.8}\\
\noalign{\smallskip}
\hline
\end{tabular}}
\end{minipage}
\vspace{-0.4cm}
\end{table}

\subsubsection{Analysis on Temporal Downsampling Rate of TDS.} As shown in Table~\ref{table:tds}, we investigate the impact of temporal downsampling rate $s$ for different numbers of input frames $N$. TDS brings up to 2.1mm, 1.5mm, 0.2mm performance improvement in the case of $N=27,81,243$ respectively. The results show that when $N$ is fixed, increasing $s$ can enlarge the temporal receptive field (RF) and reduce data redundancy, thus improving the prediction performance. In other words, when RF is fixed, TDS is an effective approach to reduce computational overhead without much performance sacrifice. Besides, the gain from TDS gradually decreases as $N$ becomes larger. When $N=27$ and $s=3$, TDS improves the performance by 1.4mm. However, when $N=243$ and $s=2$, TDS only brings 0.2mm improvement. This can be explained by the fact that the farther the two frames are separated in the time domain, the less they are correlated. Therefore, the farther the added frames are from the middle frame, the less help it brings to the 3D pose estimation of the middle frame. 

\subsubsection{Computational Complexity.}
We report the number of parameters of our framework, the number of output frames per second (FPS), floating-point operations (FLOPs) per frame at inference time, and the performance. We conduct experiments on a single GeForce GTX 3080Ti GPU. For our method, we do not include the number of parameters and FLOPs of Stage \RNum{1} in the calculation results, because we only care about the speed of inference. Once the training is completed, the decoder in Stage \RNum{1} is discarded, which does not impose any burden on the inference process. As shown in Table~\ref{tab:flops}, our 243-frame P-STMO-S has fewer parameters and FLOPs than 81-frame Poseformer. P-STMO outperforms PoseFormer and anatomy-aware model~\cite{chen2021anatomy} by 1.5/1.3mm while bringing a $\sim$1.5/7.1$\times$ speedup. This reveals the effectiveness and efficiency of the proposed method. Note that we only calculate the time for the input data to pass through the model, which does not include data pre-processing time. Therefore, the FPS we report for Poseformer is larger than the result in the original paper.

\section{Conclusion}

In this paper, we present P-STMO, a Pre-trained Spatial Temporal Many-to-One model for 3D human pose estimation. A pre-training task, called MPM, is proposed to enhance the representation capability of SEM and TEM. For SEM, an MLP block is used as the backbone, which is more effective than Transformer and \emph{fc}. For TEM, a temporal downsampling strategy is introduced to mitigate input data redundancy and increase the temporal receptive field. Comprehensive experiments on two datasets demonstrate that our method achieves superior performance over state-of-the-art methods with smaller computational budgets.

%
%

\appendix
\section{Supplementary Material}

\subsection{Other Related Work}
Hu \textit{et al.}~\cite{hu2021signbert} also explores the spatial temporal masking strategy in the field of sign language recognition. In addition to being applied in different fields, our work differs from theirs in the following ways: 1) They set the coordinates of the masked joints to (0,0), which is a very common position (center of the image). Thus, this approach will lead to confusion between the masked and unmasked joints. In contrast, we replace the masked joints with learnable vectors. These special identifiers enable the model to distinguish whether a joint is masked or not. 2) They take all 2D poses as inputs, while we only take 2D unmasked poses as inputs, which greatly reduces the computational complexity of the encoder. 

Liu \textit{et al.}~\cite{liu2022view} learn 2D human pose embeddings for downstream tasks, which is similar to our work. Our method differs from theirs in these ways: 1) They aim to learn view-invariant embeddings by metric learning, while we aim to learn spatial temporal embeddings by solving the MPM task. 2) The masking strategy is utilized to increase the difficulty of pre-training in our method, while they use it to simulate the real-world occlusion situation. 3) We use Transformer, whose token design is more suitable for masking. 4) We explore the spatial temporal masking strategy, while they do not.

\subsection{Implementation Details}
All experiments are carried out on a single GeForce GTX 3080Ti GPU. The proposed method is implemented using PyTorch~\cite{paszke2019pytorch}. For both stages, we train our model using the Adam~\cite{kingma2014adam} optimizer for 80 epochs. The initial learning rate is $1e^{-4}$ for Stage \RNum{1} and $7e^{-4}$ for Stage \RNum{2}. The learning rate decays by 3\% after each epoch. The batch size is set to 160. We use horizontal flipping as the data augmentation approach during training and testing. The balance factor $\lambda$ in the loss function is set to 1 empirically.

For SEM, we utilize an MLP block as the backbone. The number of sub-blocks is $L=1$. The dimension of the latent features is set to 256. For TEM as well as the decoder in Stage \RNum{1}, we utilize a vanilla Transformer~\cite{vaswani2017attention} as the backbone. The depth of Transformer for these two modules is set to 3,2 (for P-STMO-S) or 4,3 (for P-STMO), respectively. Besides, the dimension of the QKV matrices is 256. The number of heads in self-attention is 8. For MOFA, STE~\cite{li2022exploiting} is used as the backbone. The basic settings are the same as TEM. For $N=27,81,243$, the depth of Transformer is set to 3, 4, 5, respectively. The stride of 1D convolution is set to 3 for each Transformer layer. 

\subsection{Analysis on Model Hyperparameters}
As shown in Table~\ref{tab:hyperparam}, we mainly investigate three hyperparameters in our method: the depth of TEM ($L_1$), the depth of the decoder in Stage \RNum{1} ($L_D$), and the dimension of latent features of all Transformers ($d$). Following~\cite{he2022masked}, we adopt an asymmetric encoder-decoder design.  Unlike classical auto-encoders, the depth of the decoder can be different from that of the encoder. Since we only care about the speed of inference, the number of parameters and FLOPs of the pre-training stage are not included in the calculation results. Thus, these results are not affected by different settings of $L_D$. The best performance is achieved when $L_1=4, L_D=3, d=256$.

\begin{figure}[t]
\centering
\subfigure[Directions]{
\begin{minipage}[t]{.45\linewidth}
\includegraphics[width=\linewidth]{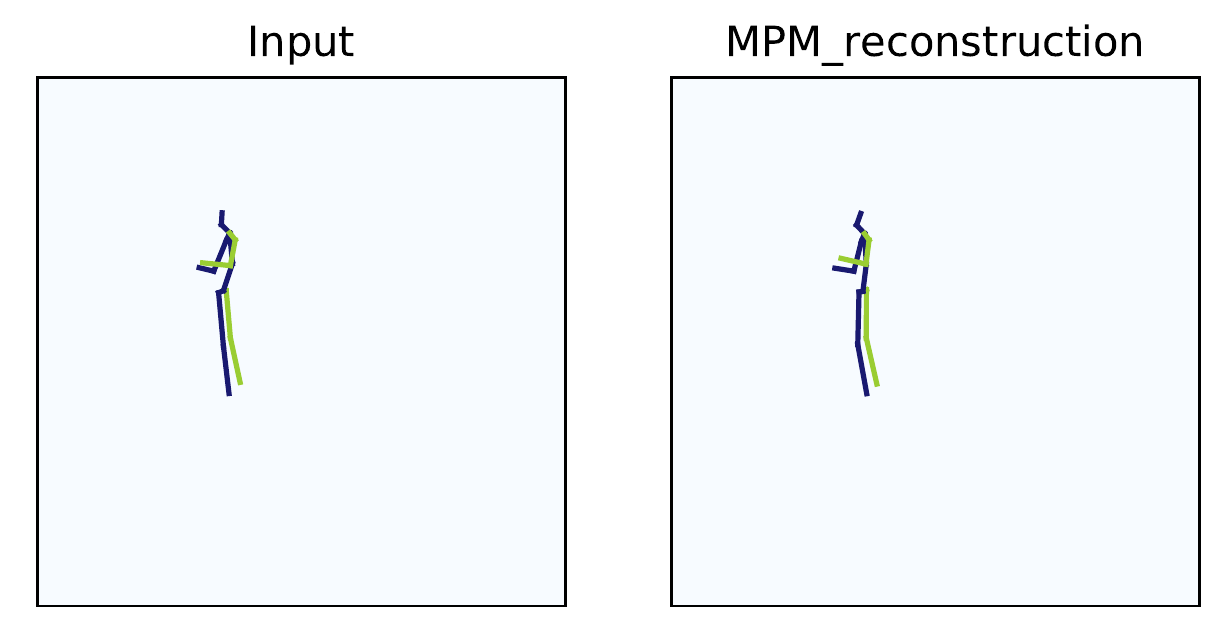}
\end{minipage}}
\subfigure[Walking]{
\begin{minipage}[t]{.45\linewidth}
\includegraphics[width=\linewidth]{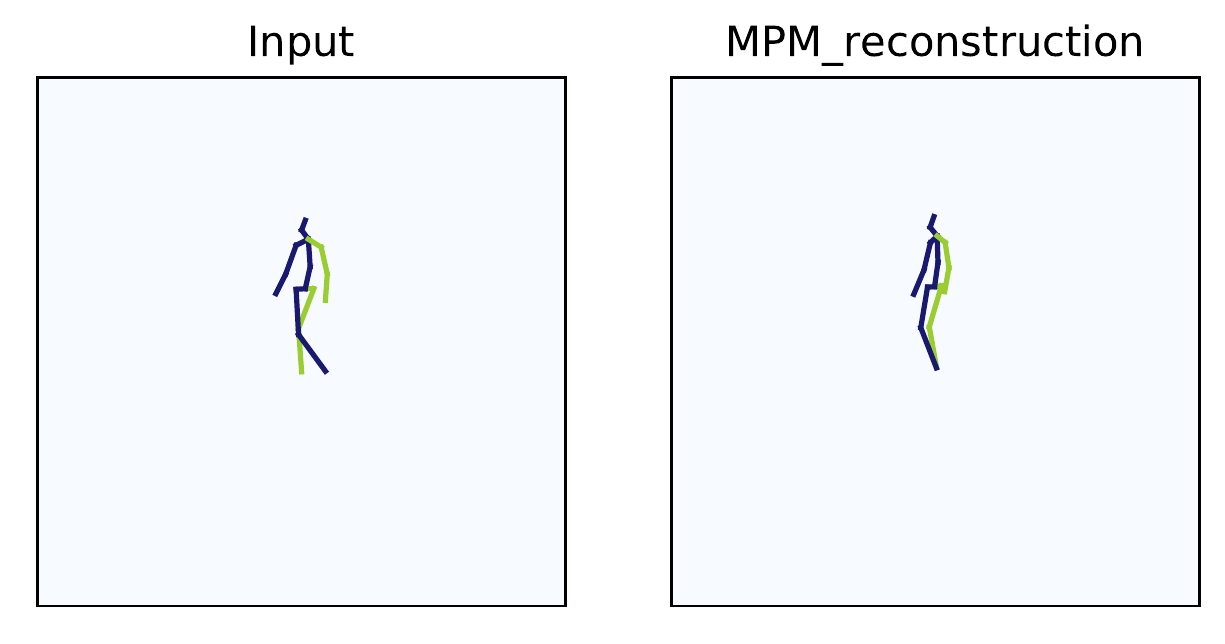}
\end{minipage}}
\subfigure[SittingDown]{
\begin{minipage}[t]{.45\linewidth}
\includegraphics[width=\linewidth]{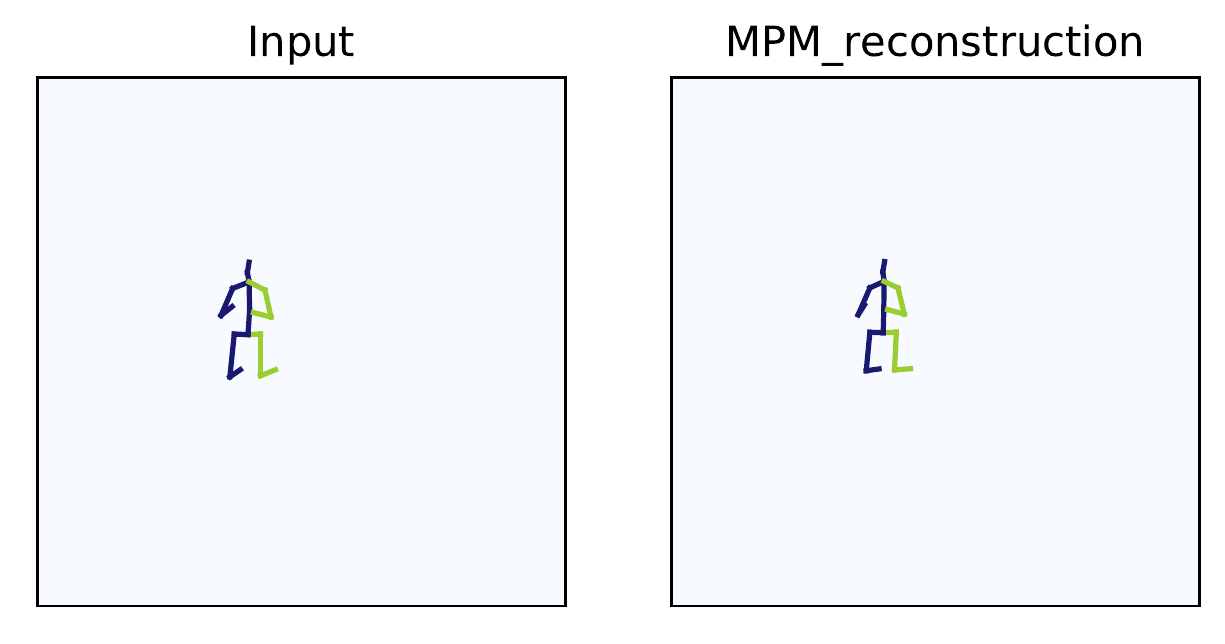}
\end{minipage}}
\subfigure[Photo]{
\begin{minipage}[t]{.45\linewidth}
\includegraphics[width=\linewidth]{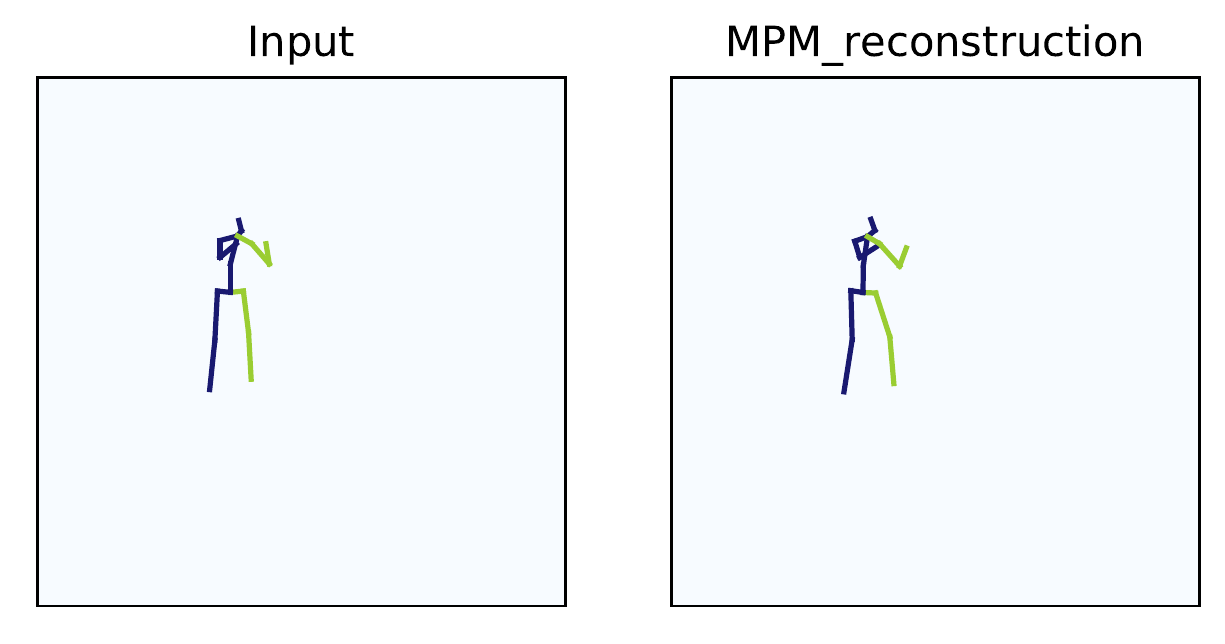}
\end{minipage}}
\vspace{-0.2cm}
\caption{Qualitative results of the 2D pose reconstruction in Stage \RNum{1}.}
\vspace{-0.2cm}
\label{fig:stage1_qualitative}
\end{figure}

\subsection{Reconstruction Results in the Pre-Training Stage}
In Fig.~\ref{fig:stage1_qualitative}, we give some qualitative results of the 2D pose reconstruction in the pre-training stage. The 2D input sequence is masked with $q^\text{T}=0.8, m^\text{S}=2$, which means only 18.7\% of the joints are visible to the network. It can be seen that the network is able to perform a rough recovery of the 2D poses in the original sequence by virtue of only a small number of visible joints.

\subsection{Visualization of Multi-Head Self-Attention} We perform a subjective analysis of the self-attention mechanism in Transformer in both stages. The results are shown in Fig.~\ref{fig:stage1_enc}-\ref{fig:stage2_STE}. The x-axis (horizontal) and y-axis (vertical) correspond to the index of key (K) and query (Q) in the multi-head self-attention respectively. The output is the normalized attention weight. We set the input frame number to $N=243$.
\subsubsection{Stage \RNum{1}.} 
\begin{figure}[t]
\centering
\subfigure{
\begin{minipage}[t]{\linewidth}
\centering
\includegraphics[width=0.24\linewidth]{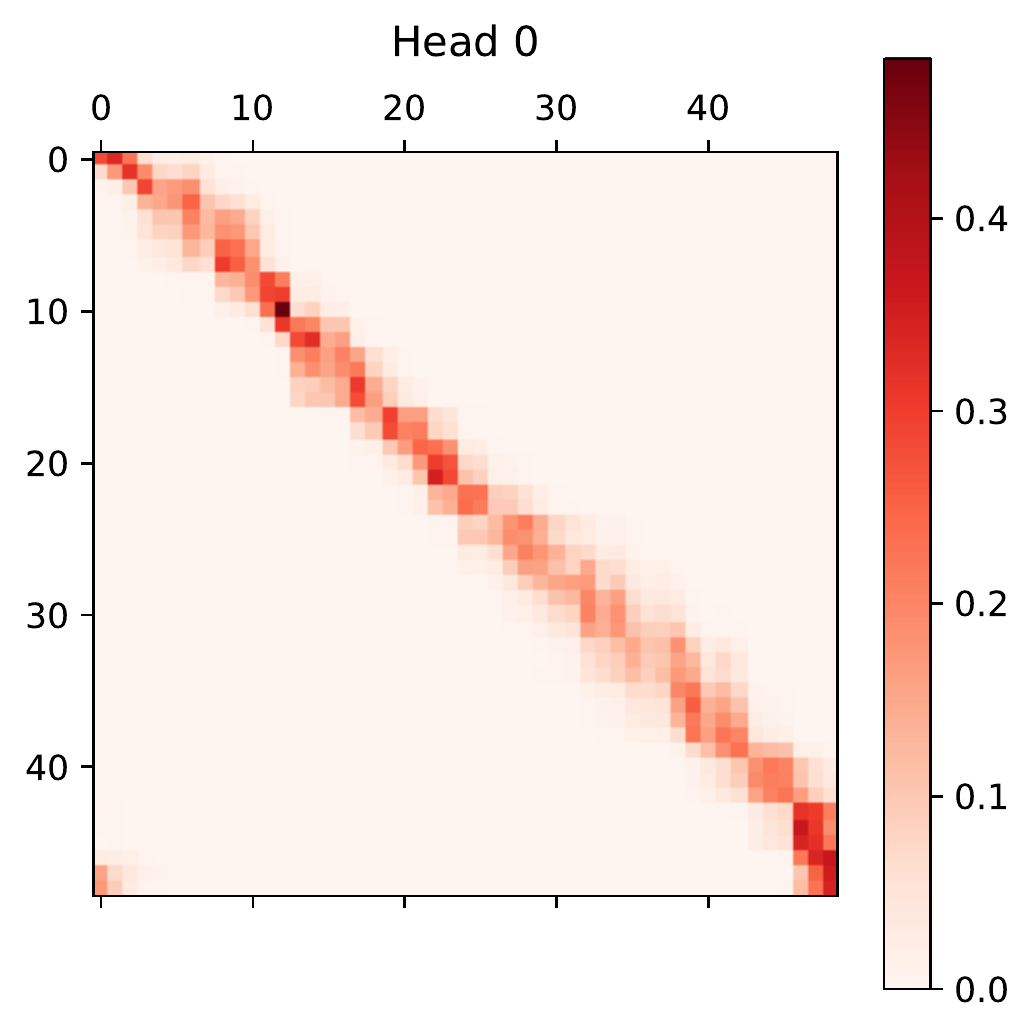}
\includegraphics[width=0.24\linewidth]{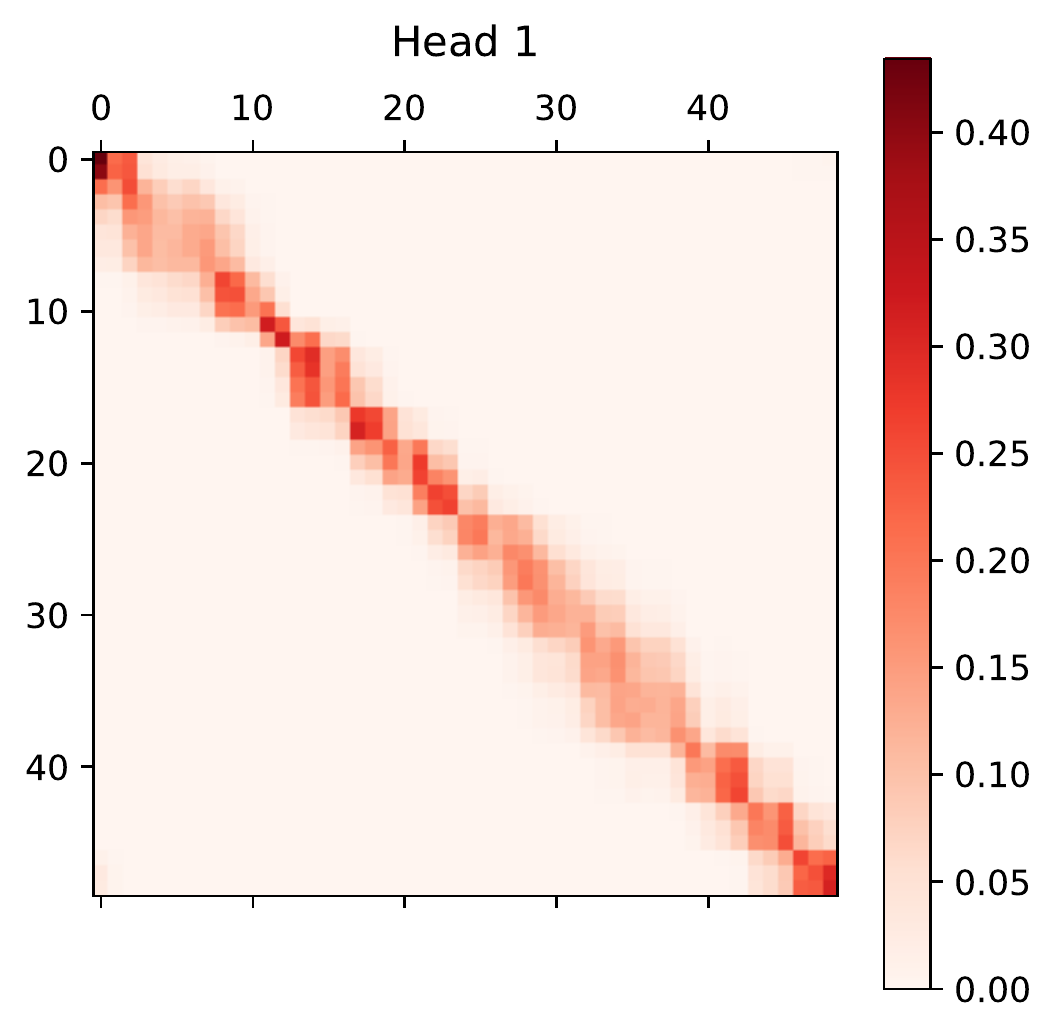}
\includegraphics[width=0.24\linewidth]{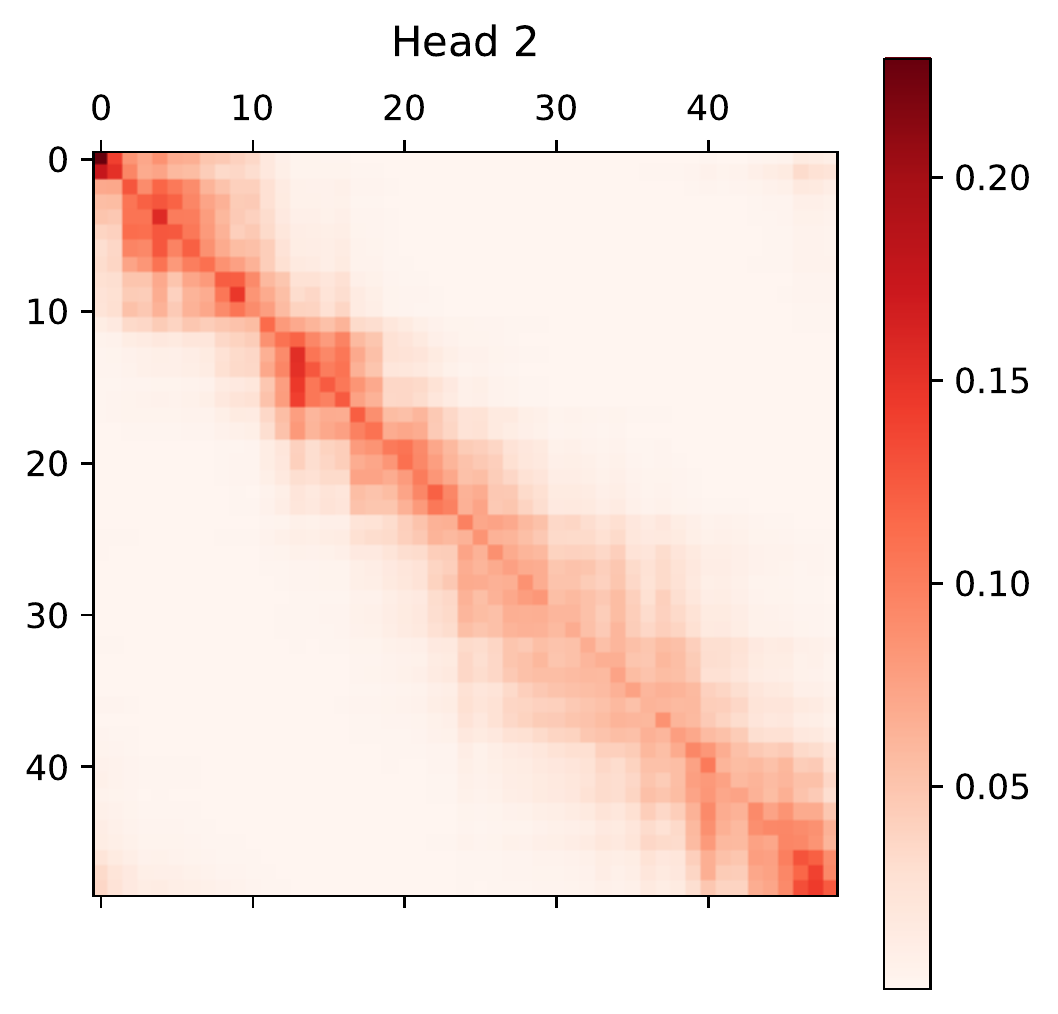}
\includegraphics[width=0.24\linewidth]{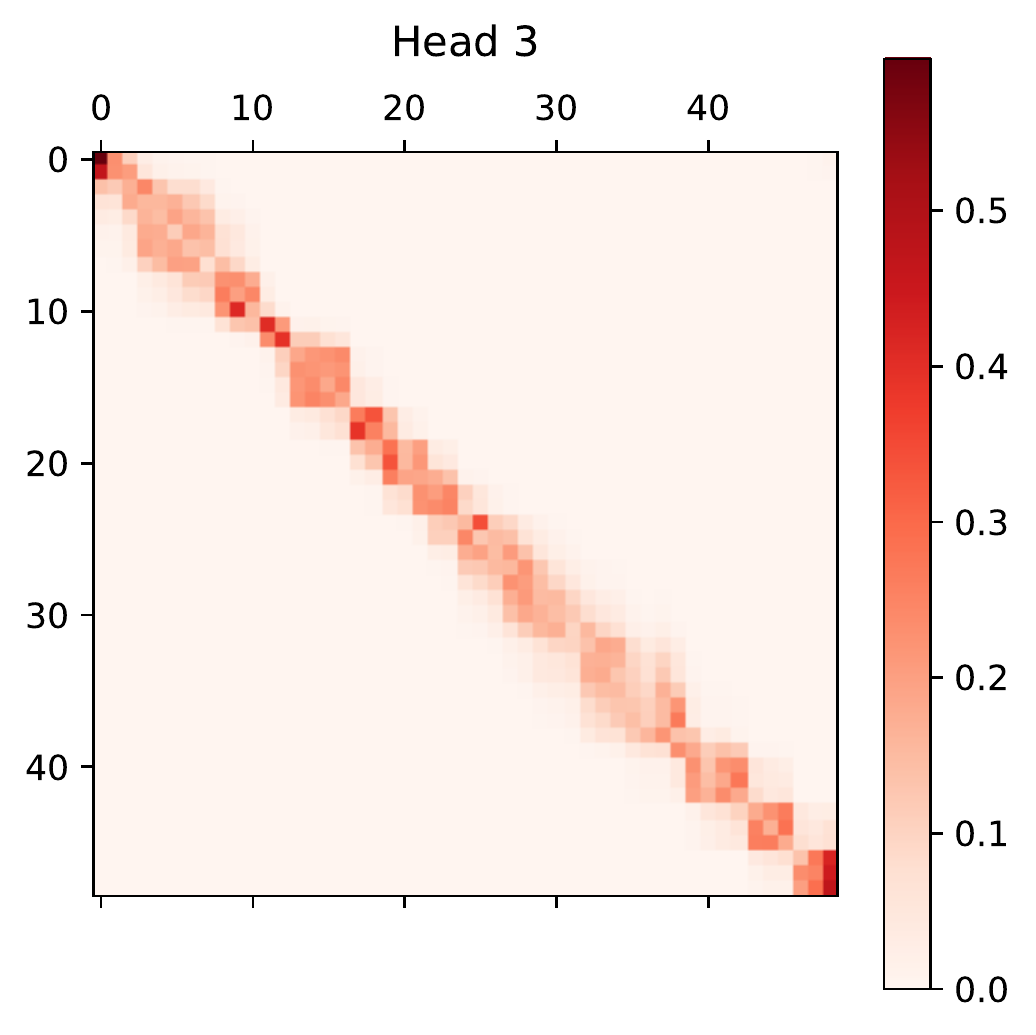}
\end{minipage}}
\subfigure{
\begin{minipage}[t]{\linewidth}
\centering
\includegraphics[width=0.24\linewidth]{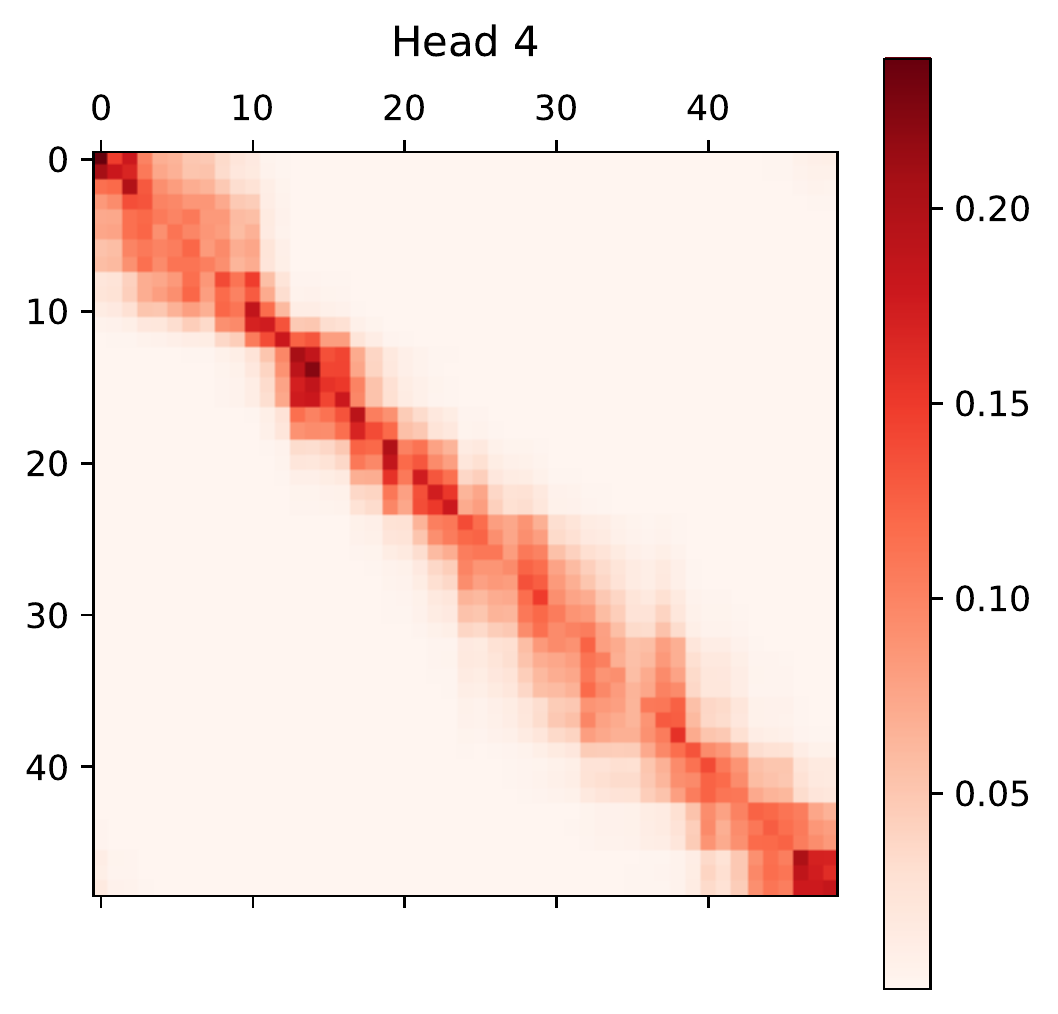}
\includegraphics[width=0.24\linewidth]{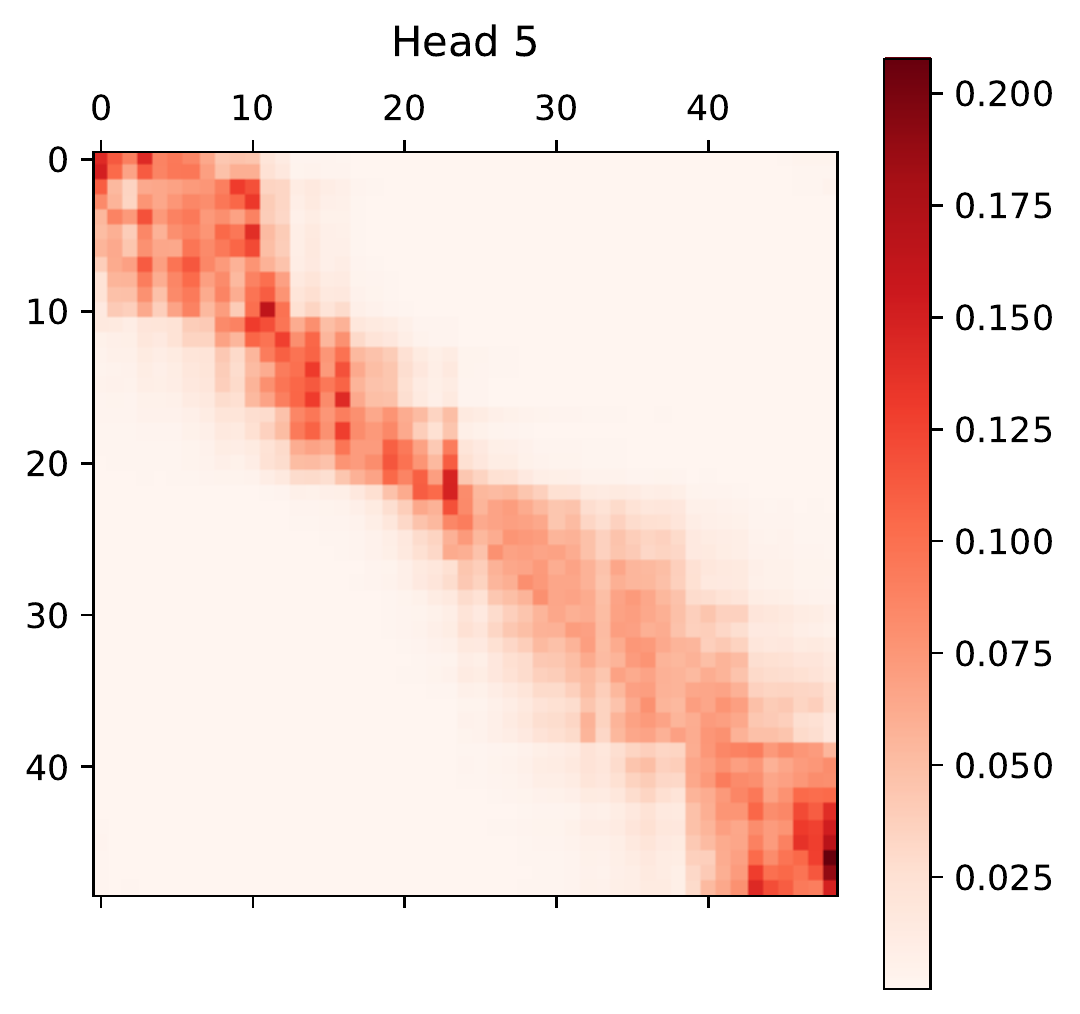}
\includegraphics[width=0.24\linewidth]{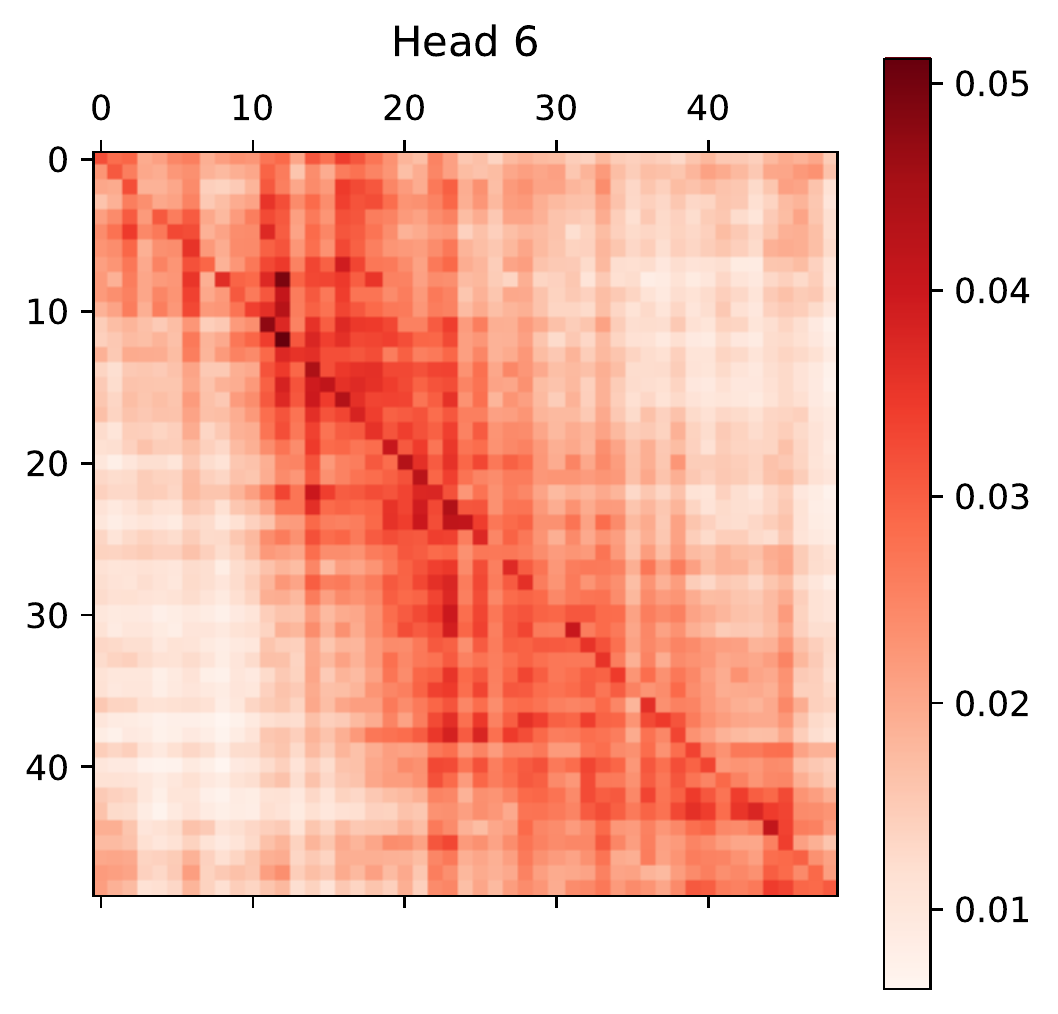}
\includegraphics[width=0.24\linewidth]{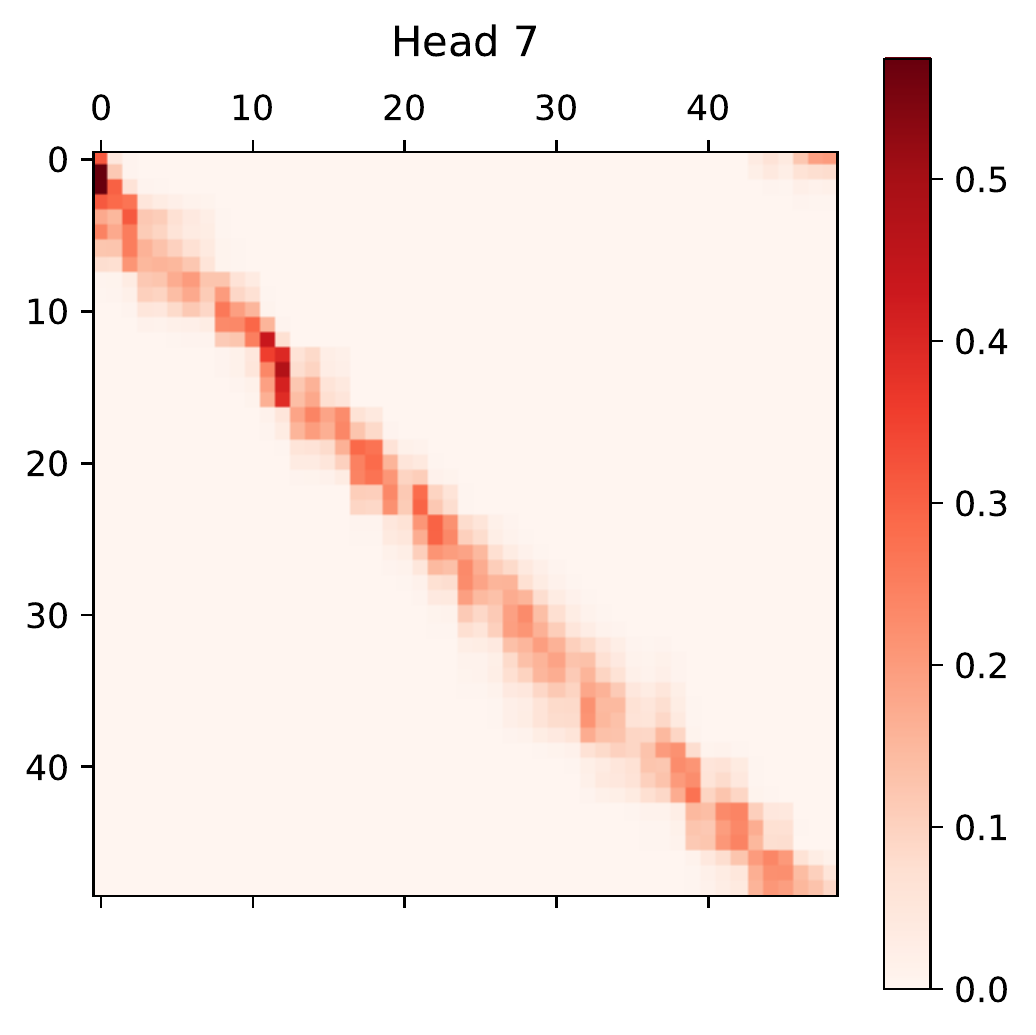}
\end{minipage}}
\vspace{-0.2cm}
\caption{Visualization of multi-head attention maps of TEM in Stage \RNum{1}.}
\vspace{-0.2cm}
\label{fig:stage1_enc}
\end{figure}

We set the temporal masking ratio to $q^\text{T}=0.8$ and spatial masking number to $m^\text{S}=2$. Therefore, the number of unmasked frames is $a=(1-q^\text{T}) \cdot N=\lfloor48.6\rfloor=48$. The number of masked frames is $b=N-a=195$. The unmasked frames are fed to the encoder (SEM+TEM). The attention maps of TEM are shown in Fig.~\ref{fig:stage1_enc}. Since the input is randomly masked, two adjacent frames in the unmasked frame sequence are most likely to be non-adjacent in the original sequence. Our goal is to recover all the masked frames between two adjacent frames in the unmasked frame sequence. Therefore, most heads of TEM focus on aggregating neighbouring information around the query frame. Others (e.g., head 6) capture long-term dependencies.

\begin{table}[t]\small
\begin{center}
\caption{Analysis on model hyperparameters. Only the number of parameters and FLOPS of Stage \RNum{2} are included in the calculation results. $L_1, L_D$ are the depth of TEM and the decoder in Stage \RNum{1} respectively. $d$ is the dimension of latent features of all Transformers.}
\vspace{-0.1cm}
\label{tab:hyperparam}
\begin{tabular}{ccc@{\hskip 0.1in}|@{\hskip 0.1in}ccc}
\hline\noalign{\smallskip}
$L_1$&$L_D$&$d$&Params(M)&FLOPs(M)&MPJPE$\downarrow$\\
\noalign{\smallskip}
\hline
\noalign{\smallskip}
4&2&256&6.7&868.5&43.0\\
4&3&256&6.7&868.5&\textbf{42.8}\\
4&4&256&6.7&868.5&42.9\\
4&5&256&6.7&868.5&43.3\\
\noalign{\smallskip}
\hline
\noalign{\smallskip}
2&3&256&5.7&613.7&44.4\\
3&3&256&6.2&741.4&43.1\\
4&3&256&6.7&868.5&\textbf{42.8}\\
5&3&256&7.3&995.9&43.4\\
\noalign{\smallskip}
\hline
\noalign{\smallskip}
4&3&64&1.1&121.8&44.9\\
4&3&128&2.6&307.0&43.7\\
4&3&256&6.7&868.5&\textbf{42.8}\\
4&3&512&19.5&2755.1&45.5\\
\noalign{\smallskip}
\hline
\end{tabular}
\end{center}
\vspace{-0.6cm}
\end{table}

\begin{figure}[t]
\centering
\subfigure{
\begin{minipage}[t]{\linewidth}
\includegraphics[width=0.24\linewidth]{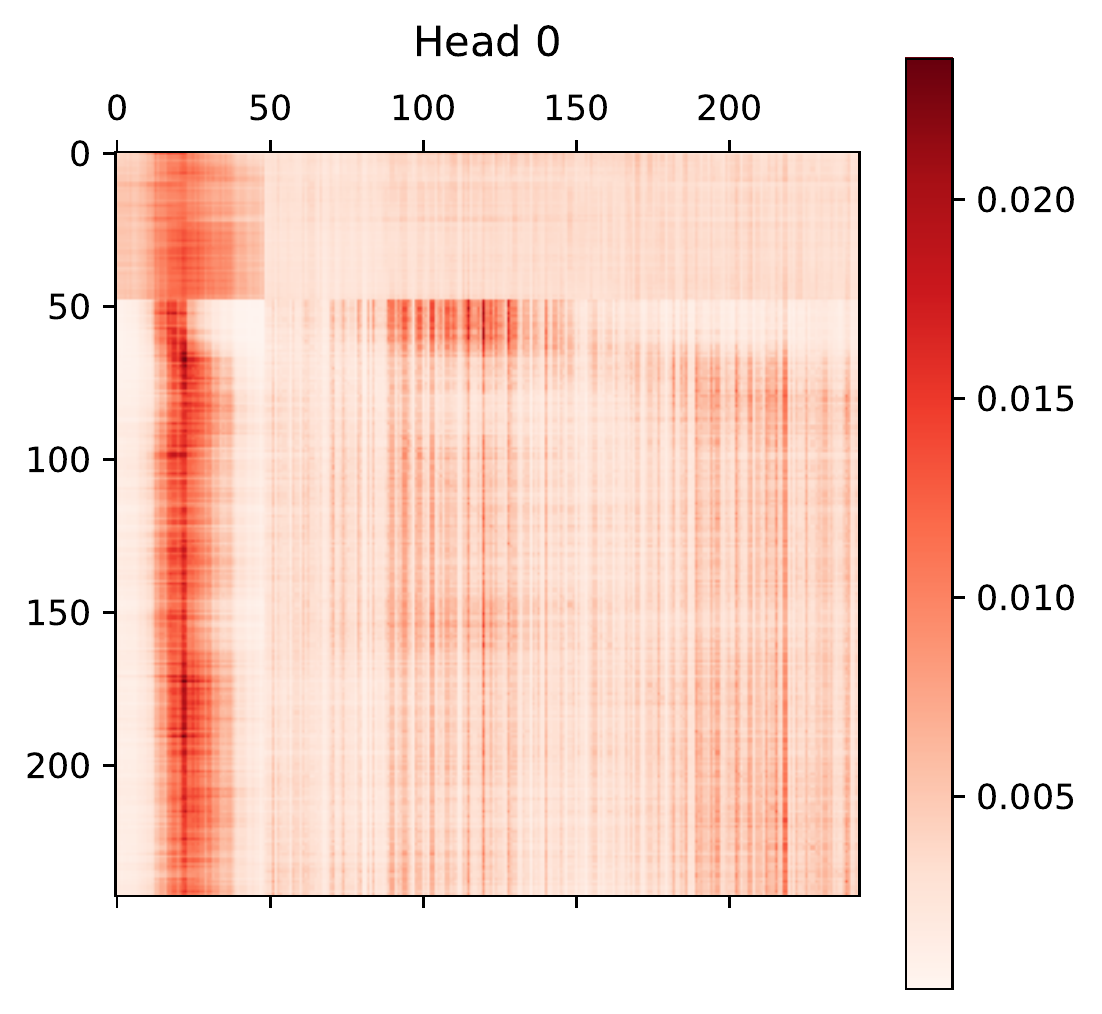}
\includegraphics[width=0.24\linewidth]{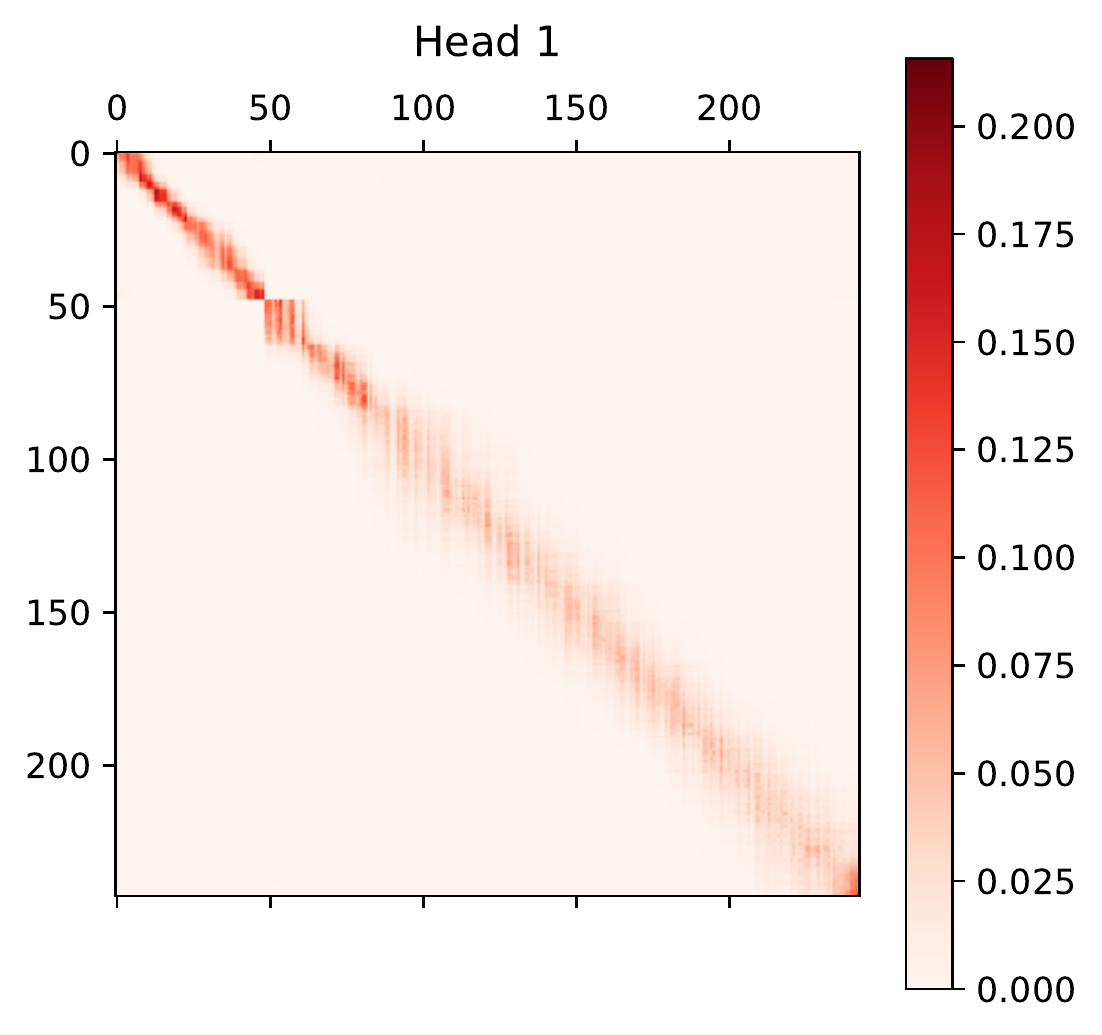}
\includegraphics[width=0.24\linewidth]{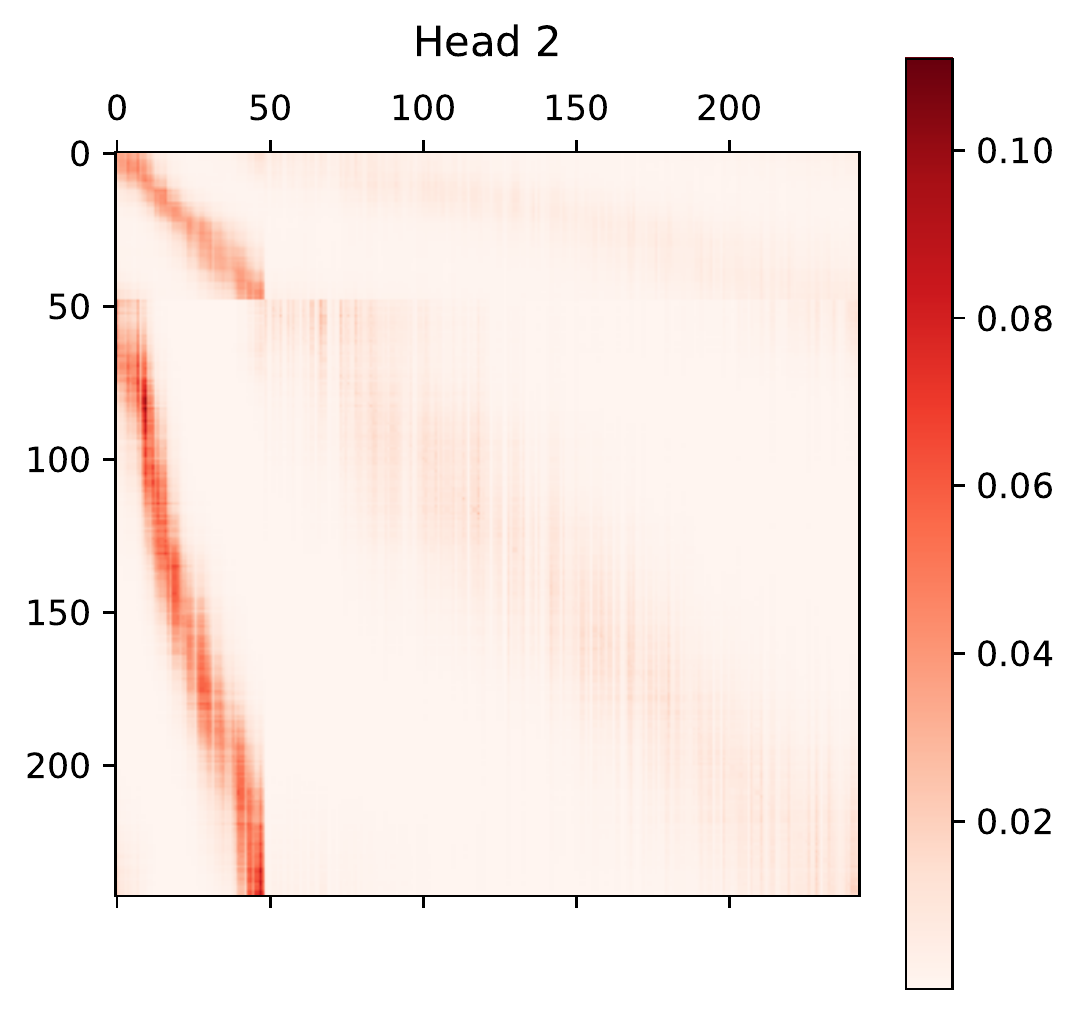}
\includegraphics[width=0.24\linewidth]{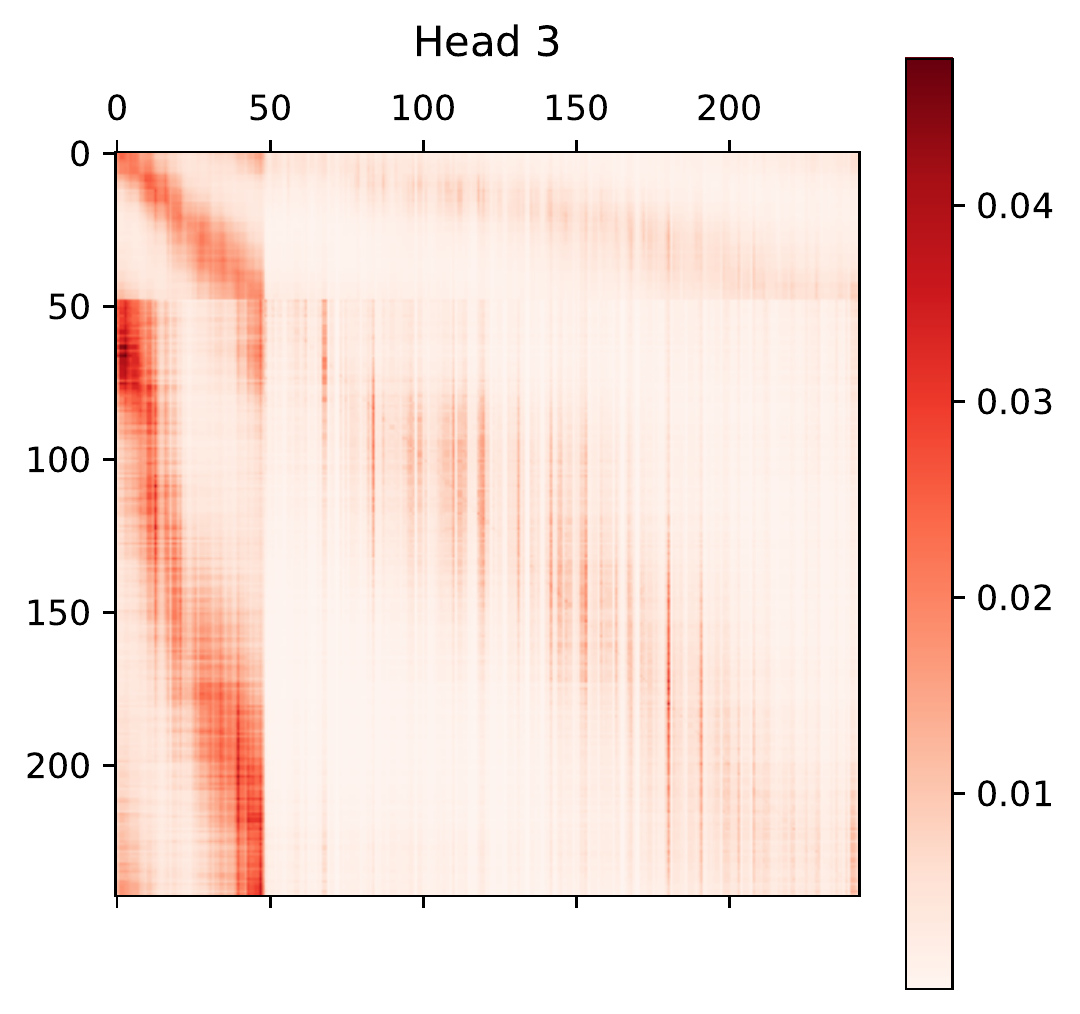}
\end{minipage}}
\subfigure{
\begin{minipage}[t]{\linewidth}
\includegraphics[width=0.24\linewidth]{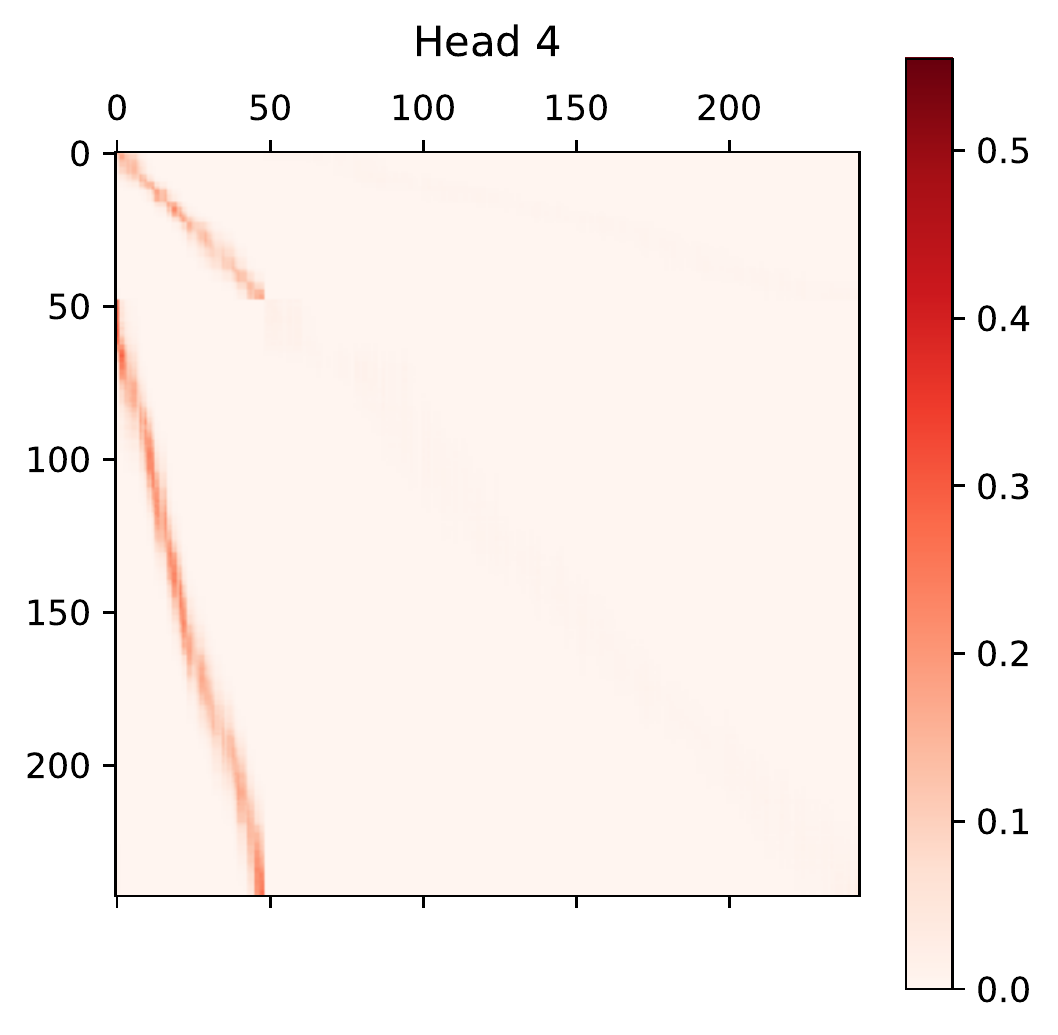}
\includegraphics[width=0.24\linewidth]{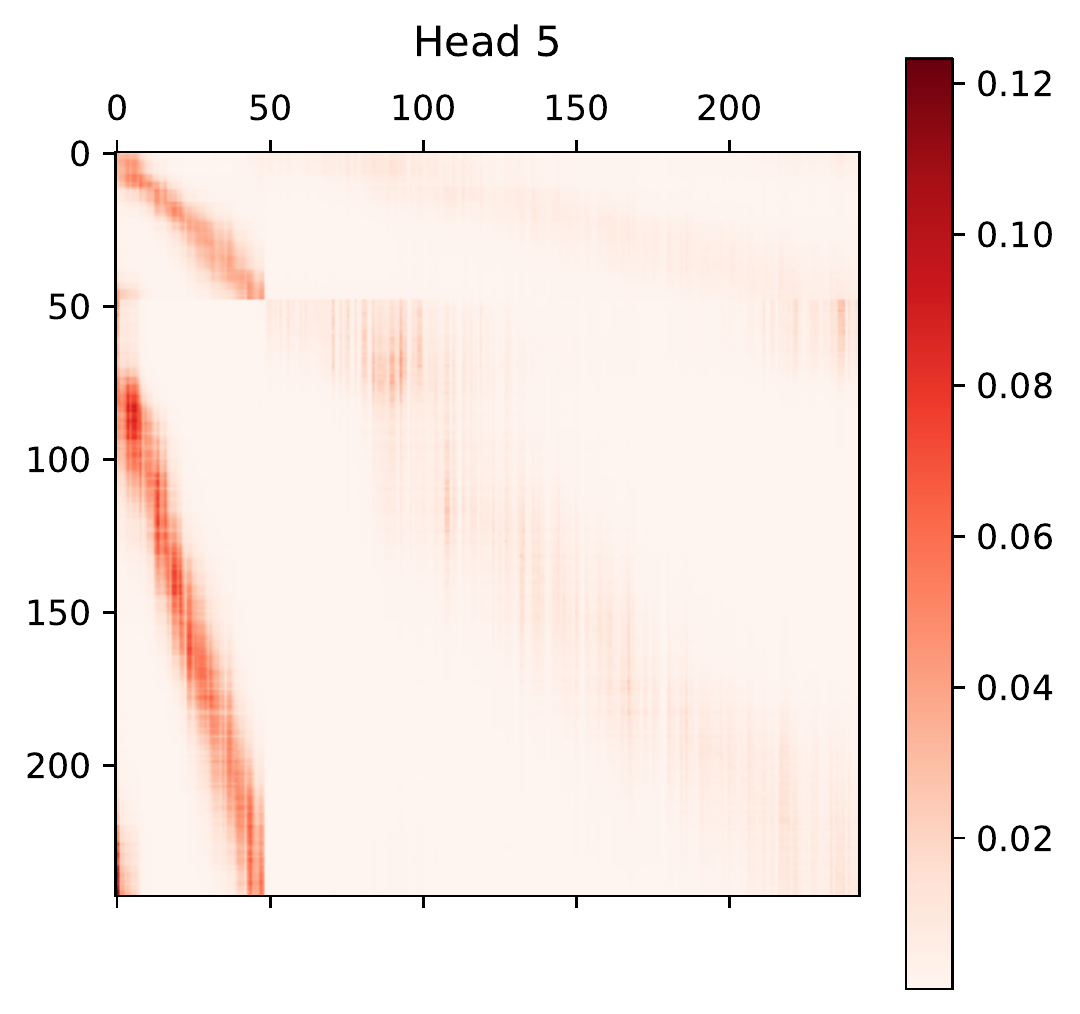}
\includegraphics[width=0.24\linewidth]{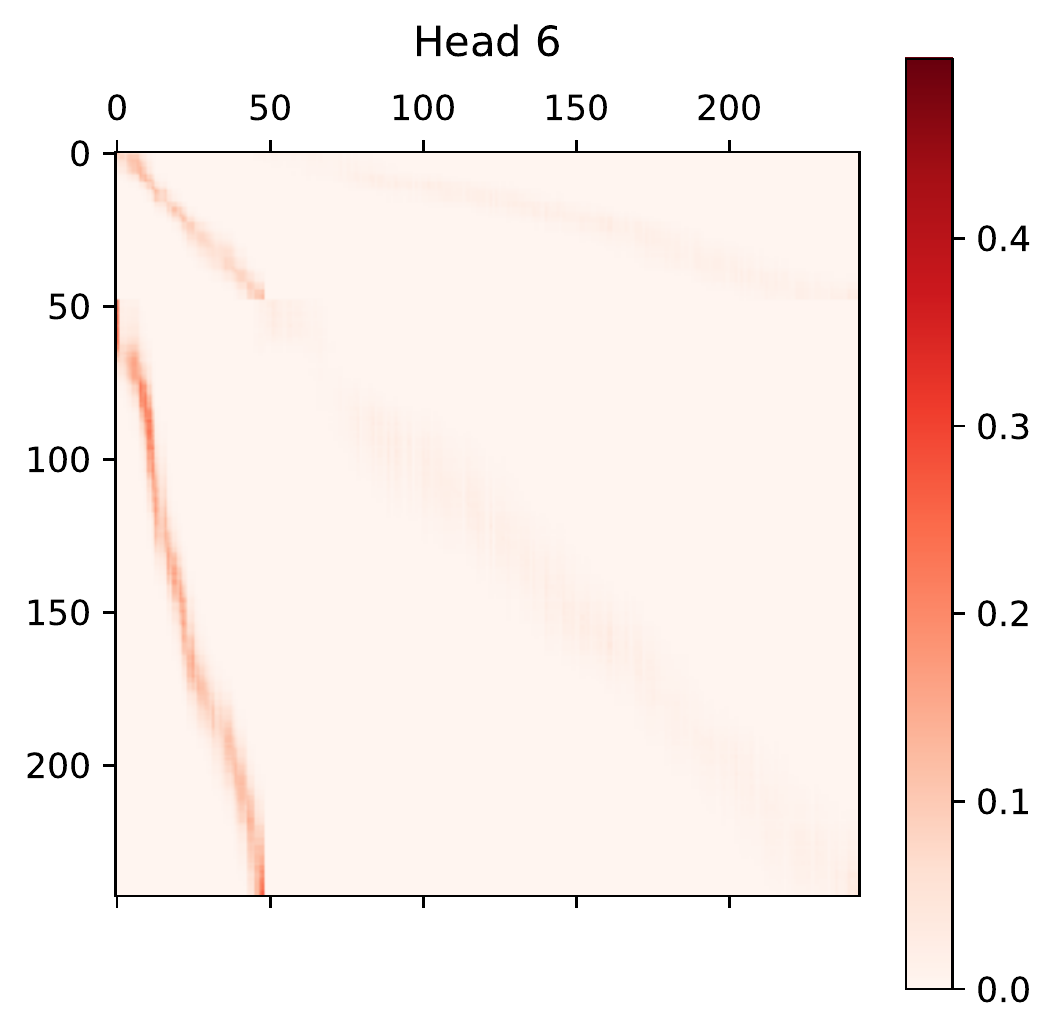}
\includegraphics[width=0.24\linewidth]{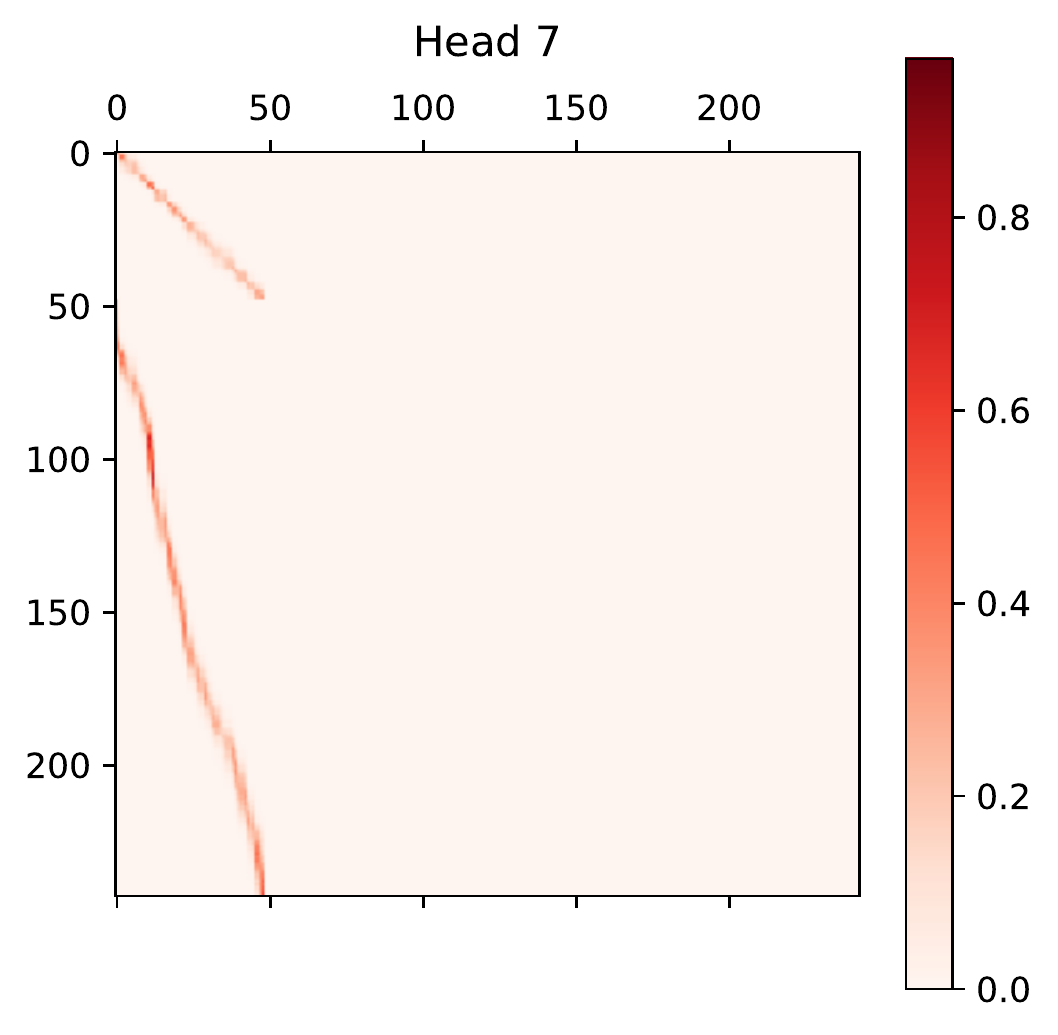}
\end{minipage}}
\vspace{-0.2cm}
\caption{Visualization of multi-head attention maps of the decoder in Stage \RNum{1}. }
\vspace{-0.2cm}
\label{fig:stage1_dec}
\end{figure}

The attention maps of the decoder are shown in Fig.~\ref{fig:stage1_dec}. An obvious cross can be seen in each attention map because we utilize the same implementation as~\cite{he2022masked}. Specifically, to improve the efficiency, we append a list of temporal padding embeddings after the encoded unmasked embeddings at the decoder side. Then, we add positional embeddings to these padding embeddings to restore their original positions. Therefore, the first half of the sequence (before frame 48) behaves differently from the second half (after frame 48). Each attention map can be divided into four matrices as follows. 1) The matrix $D_1 \in \mathbb{R}^{a\times a}$ in the upper left corner measures the self-attention of unmasked frames. The patterns are very similar to those in Fig.~\ref{fig:stage1_enc}. 2) The matrix $D_2 \in \mathbb{R}^{a\times b}$ in the upper right corner measures the degree of attention paid to the masked frames by the unmasked frames. The values of most heads are 0, which means that the masked frames provide little help to the unmasked ones. A small number of heads behave differently (e.g., head 0), which can be explained by the need to interpret the continuity of the entire pose sequence. 3) The matrix $D_3 \in \mathbb{R}^{b\times a}$ in the bottom left corner measures the degree of attention paid to the unmasked frames by the masked frames. For a particular masked frame, it resorts to the unmasked frames that are close to it. As a result, some heads (e.g., head 4) exhibit a locally relevant pattern. In addition, non-local patterns can also be observed in other heads (e.g., head 3). 4) The matrix $D_4 \in \mathbb{R}^{b\times b}$ in the bottom right corner measures the self-attention of masked frames. Since the masked frames mainly obtain information from the unmasked frames, the self-attention pattern of masked frames is not pronounced. Nevertheless, we can still find some patterns learned by a small number of heads, such as capturing long-term dependencies (e.g., head 0) and short-term dependencies (e.g., head 1).



\subsubsection{Stage \RNum{2}.}
\begin{figure}[t]
\centering
\subfigure{
\begin{minipage}[t]{\linewidth}
\centering
\includegraphics[width=0.24\linewidth]{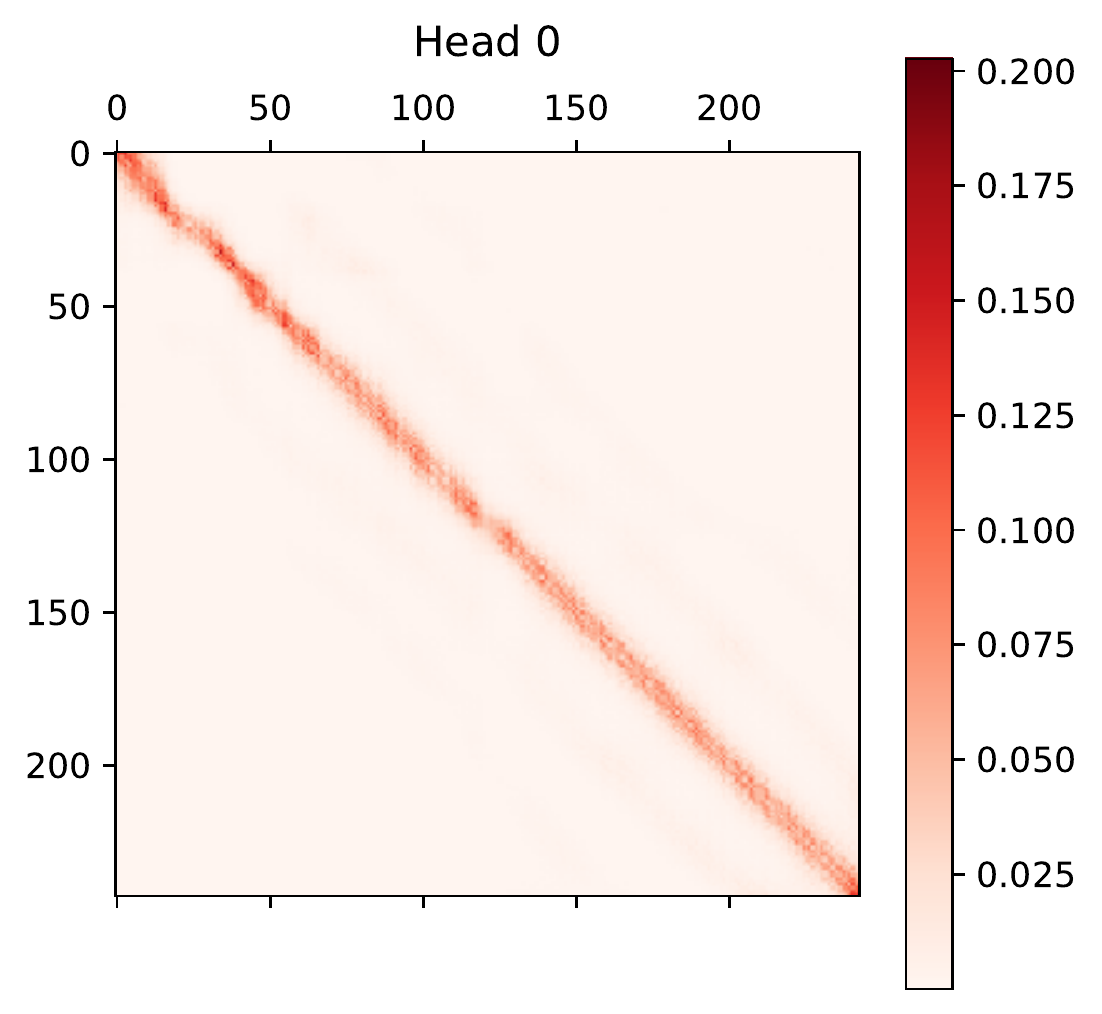}
\includegraphics[width=0.24\linewidth]{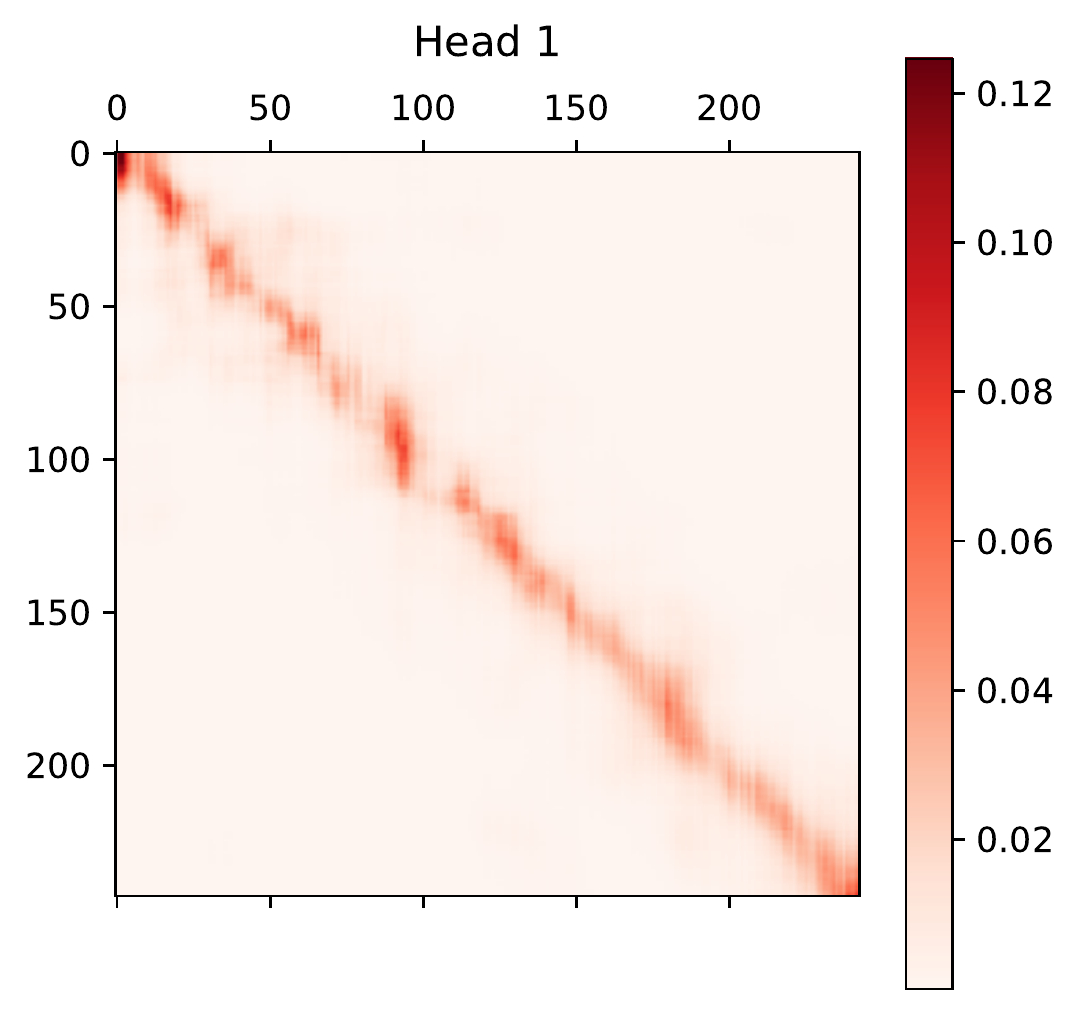}
\includegraphics[width=0.24\linewidth]{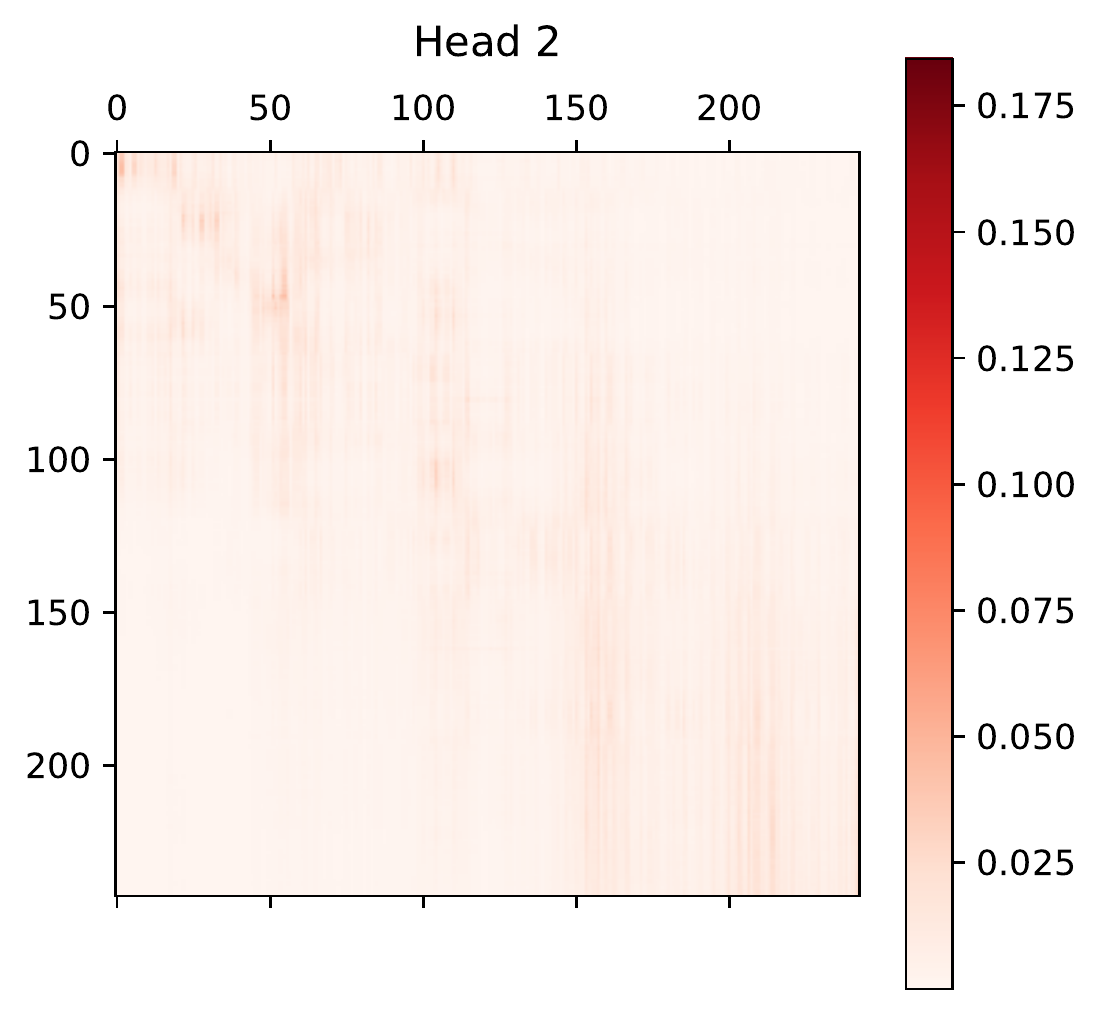}
\includegraphics[width=0.24\linewidth]{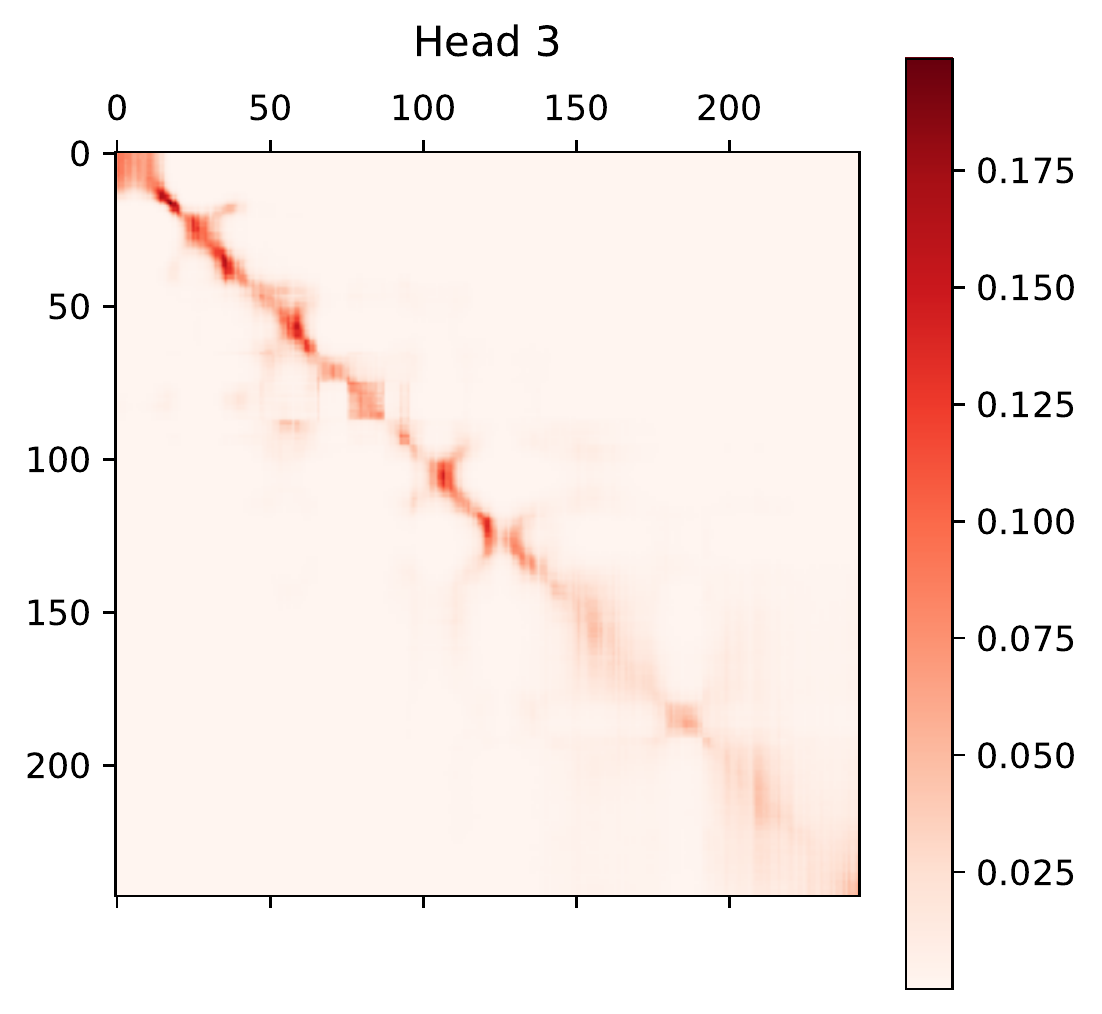}
\end{minipage}}
\subfigure{
\begin{minipage}[t]{\linewidth}
\centering
\includegraphics[width=0.24\linewidth]{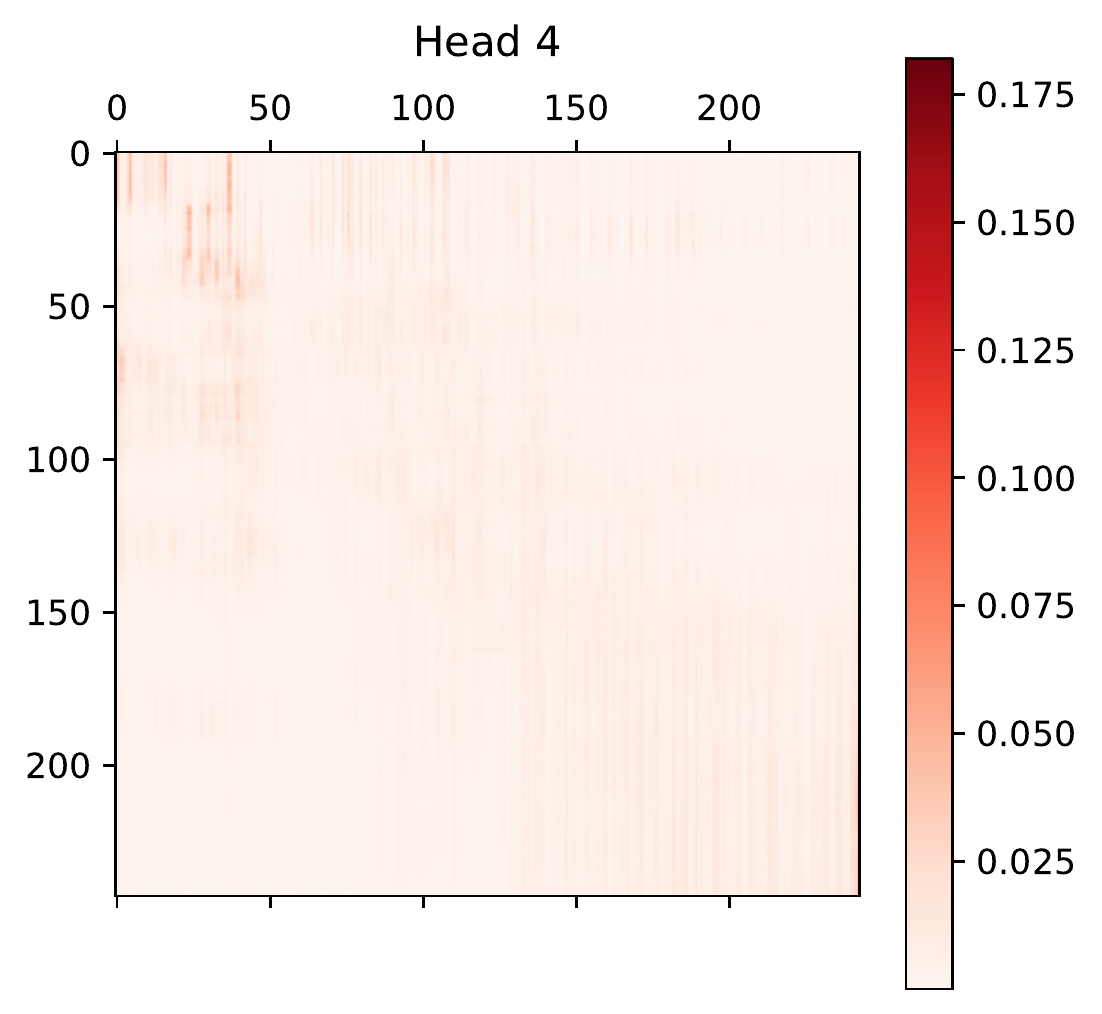}
\includegraphics[width=0.24\linewidth]{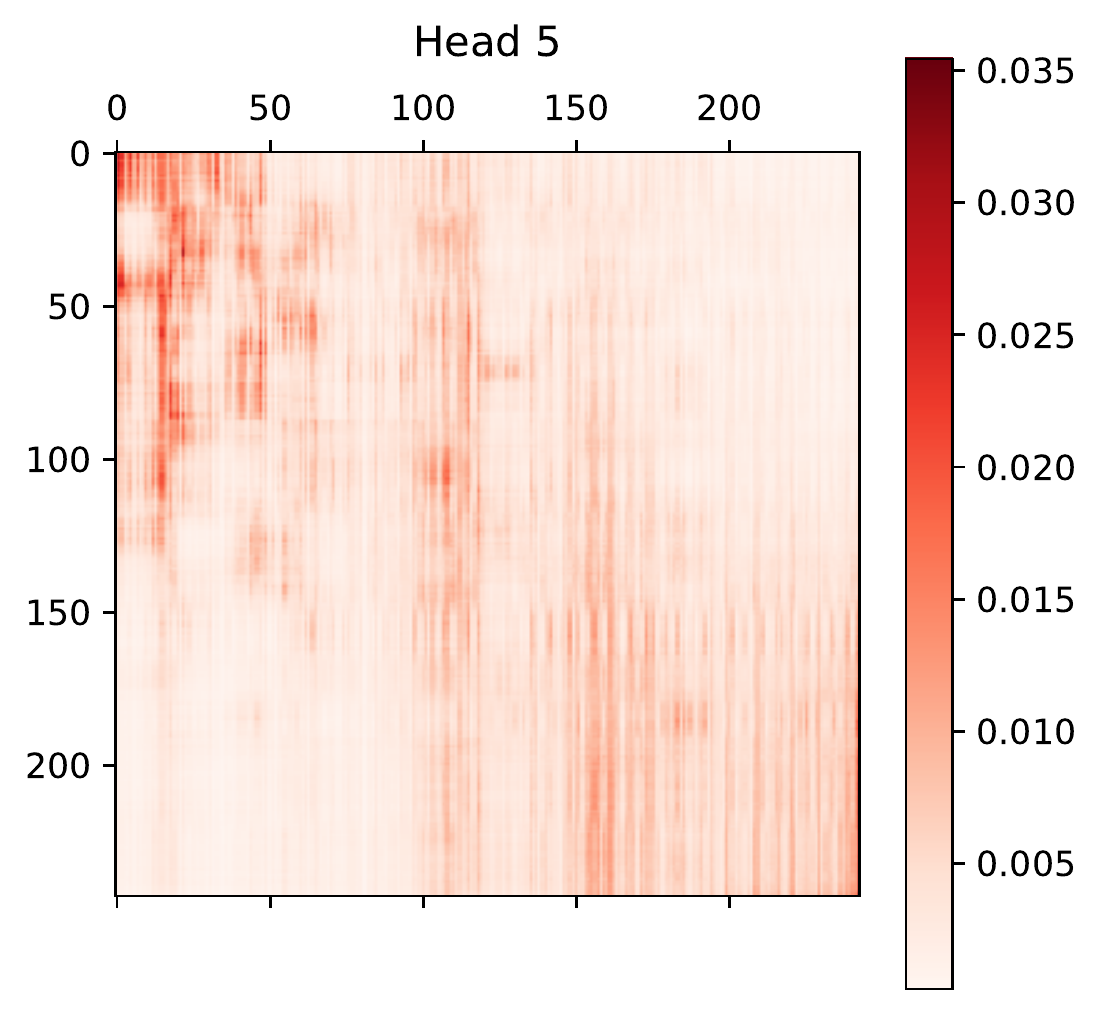}
\includegraphics[width=0.24\linewidth]{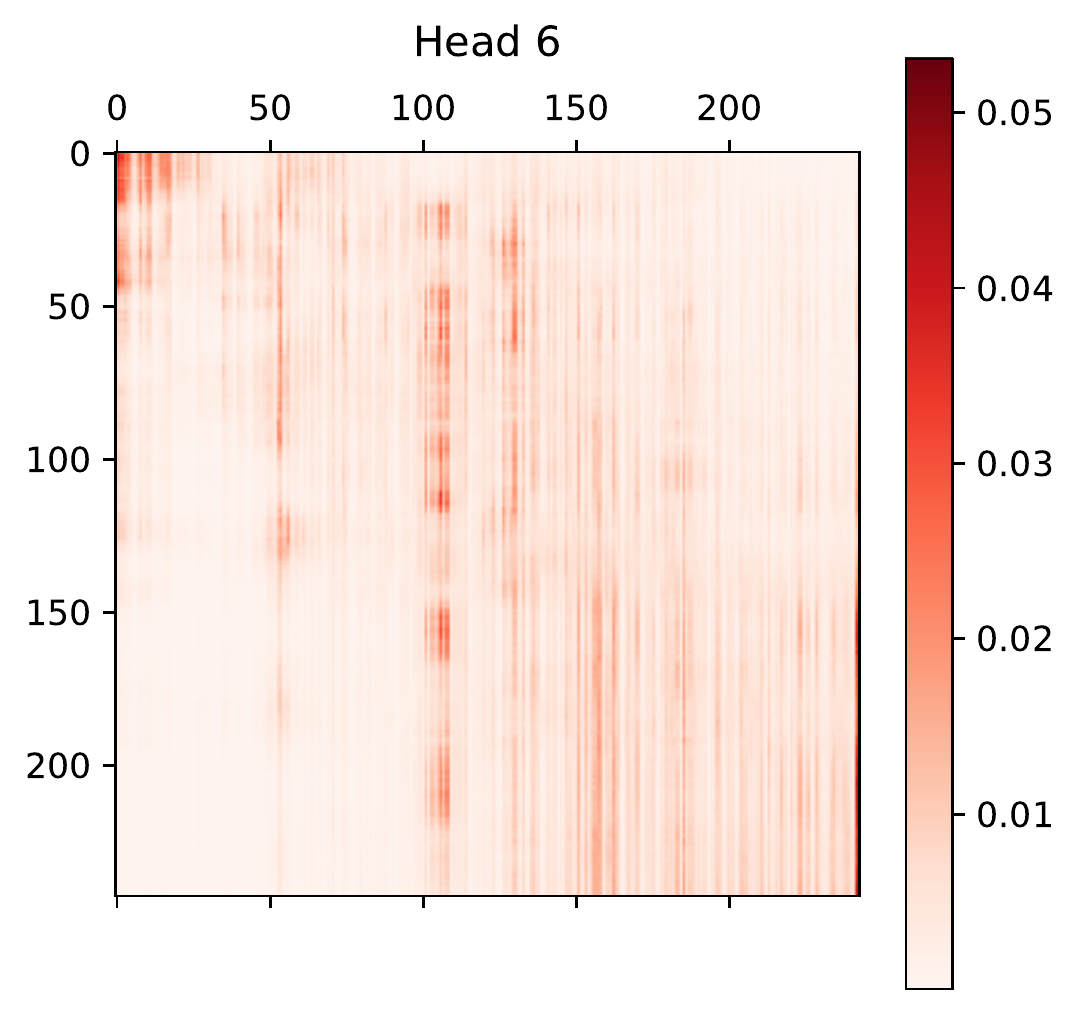}
\includegraphics[width=0.24\linewidth]{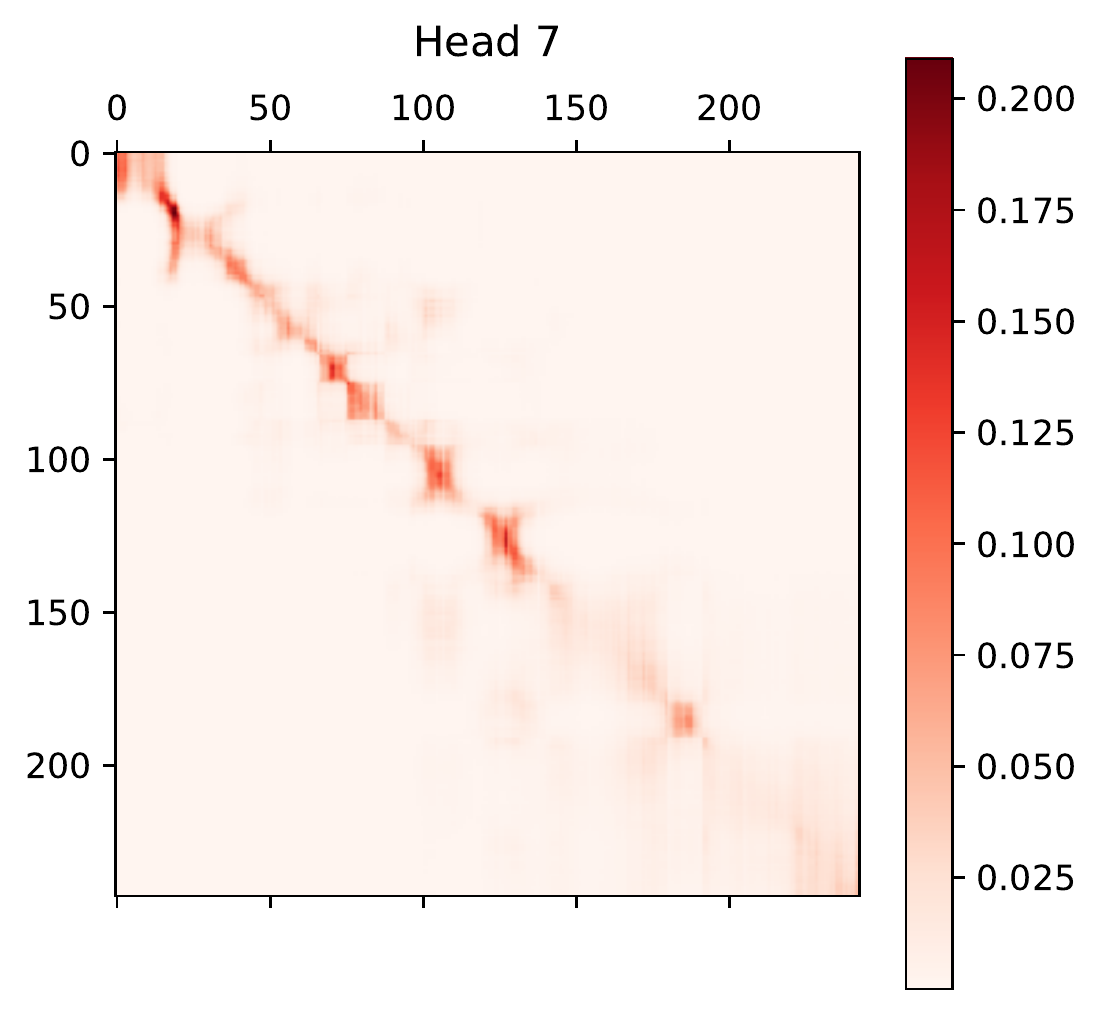}
\end{minipage}}
\vspace{-0.2cm}
\caption{Visualization of multi-head attention maps of TEM in Stage \RNum{2}.}
\vspace{-0.2cm}
\label{fig:stage2_enc}
\end{figure}

After the pre-trained model is loaded into STMO, TEM is tuned to acquire long- and short-term information that contributes to the overall 3D pose sequence prediction. As illustrated in Fig.~\ref{fig:stage2_enc}, Transformer learns a variety of patterns. The local patterns (e.g., head 0) focus on a small region around the query frame, while the global patterns (e.g., head 5) exploit relevant features over a longer time span.

\begin{figure}[t]
\centering
\subfigure[Layer 0]{
\begin{minipage}[t]{0.48\linewidth}
\includegraphics[width=0.48\linewidth]{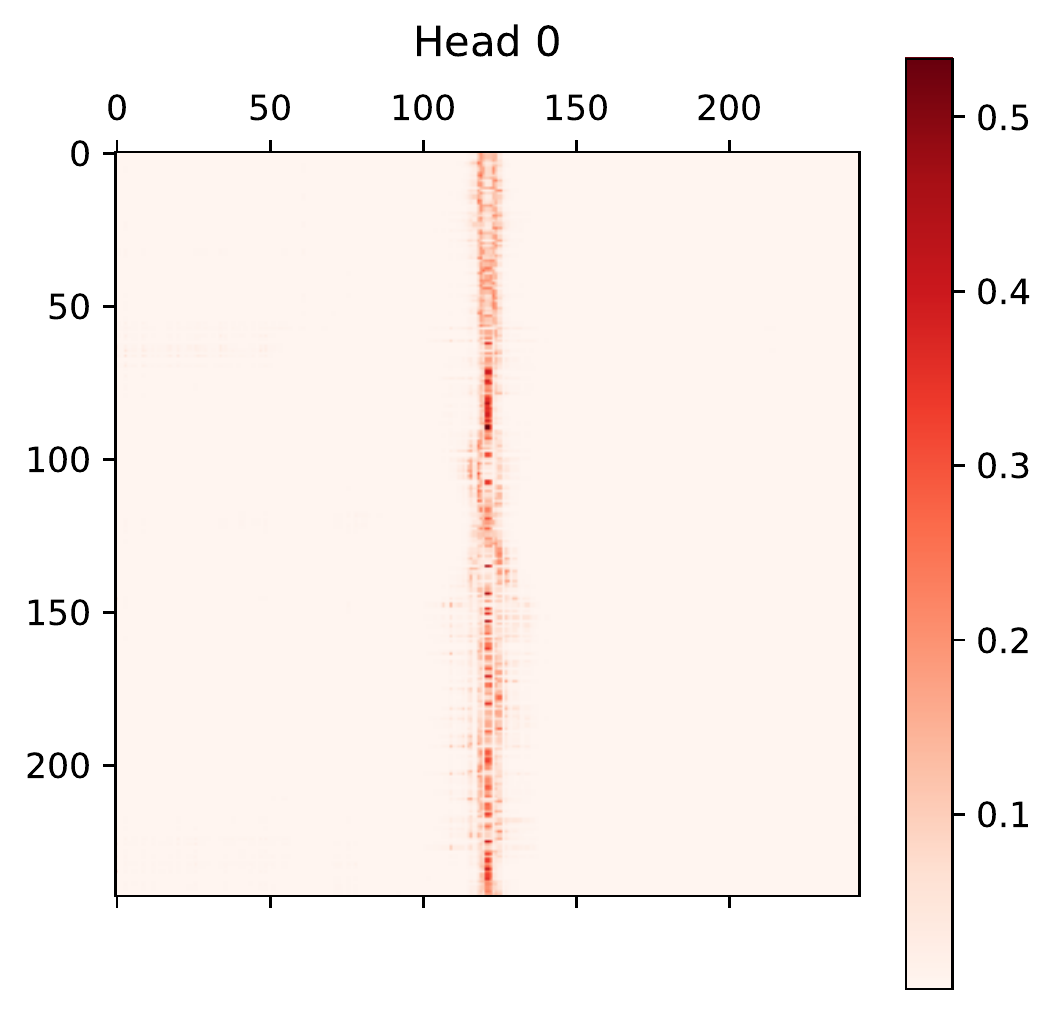}
\includegraphics[width=0.48\linewidth]{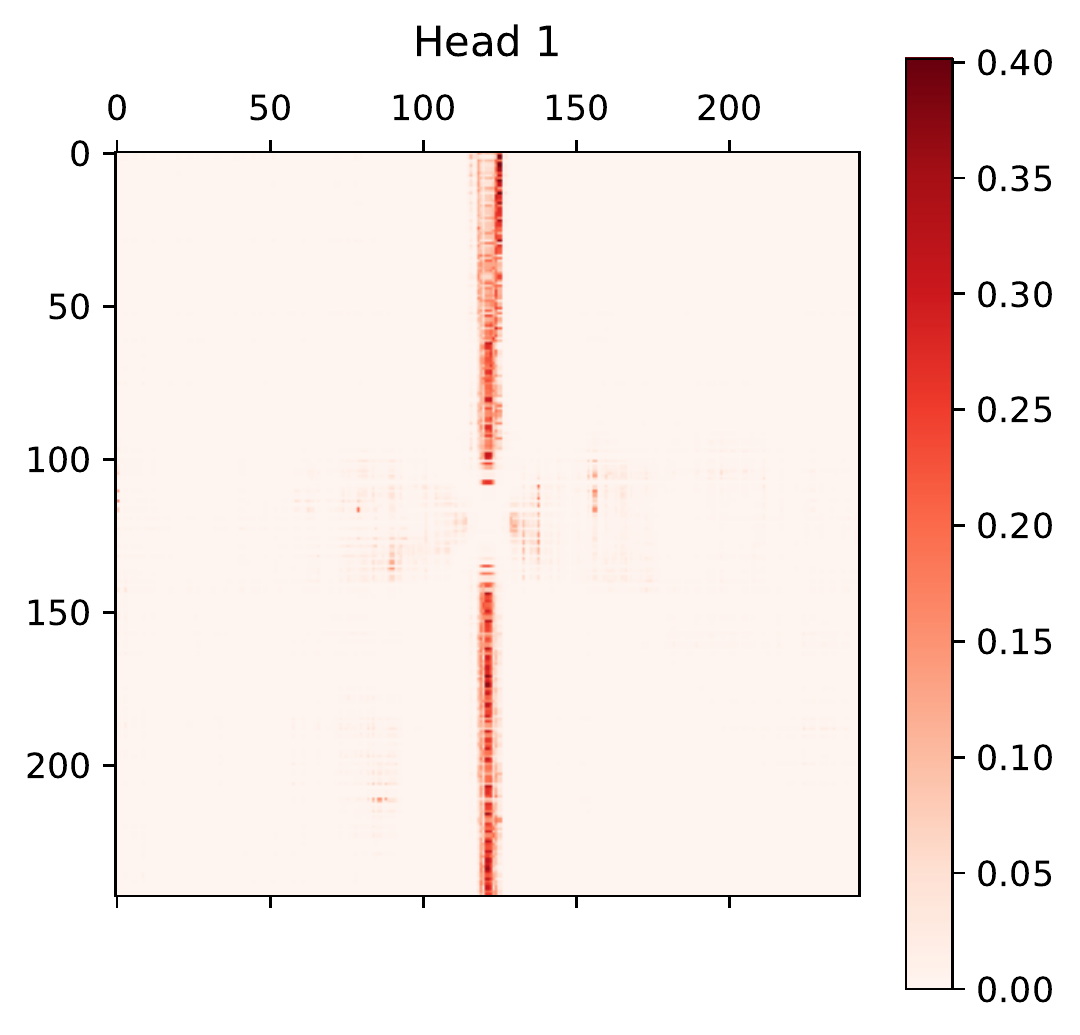}
\end{minipage}}
\subfigure[Layer 1]{
\begin{minipage}[t]{0.47\linewidth}
\includegraphics[width=0.48\linewidth]{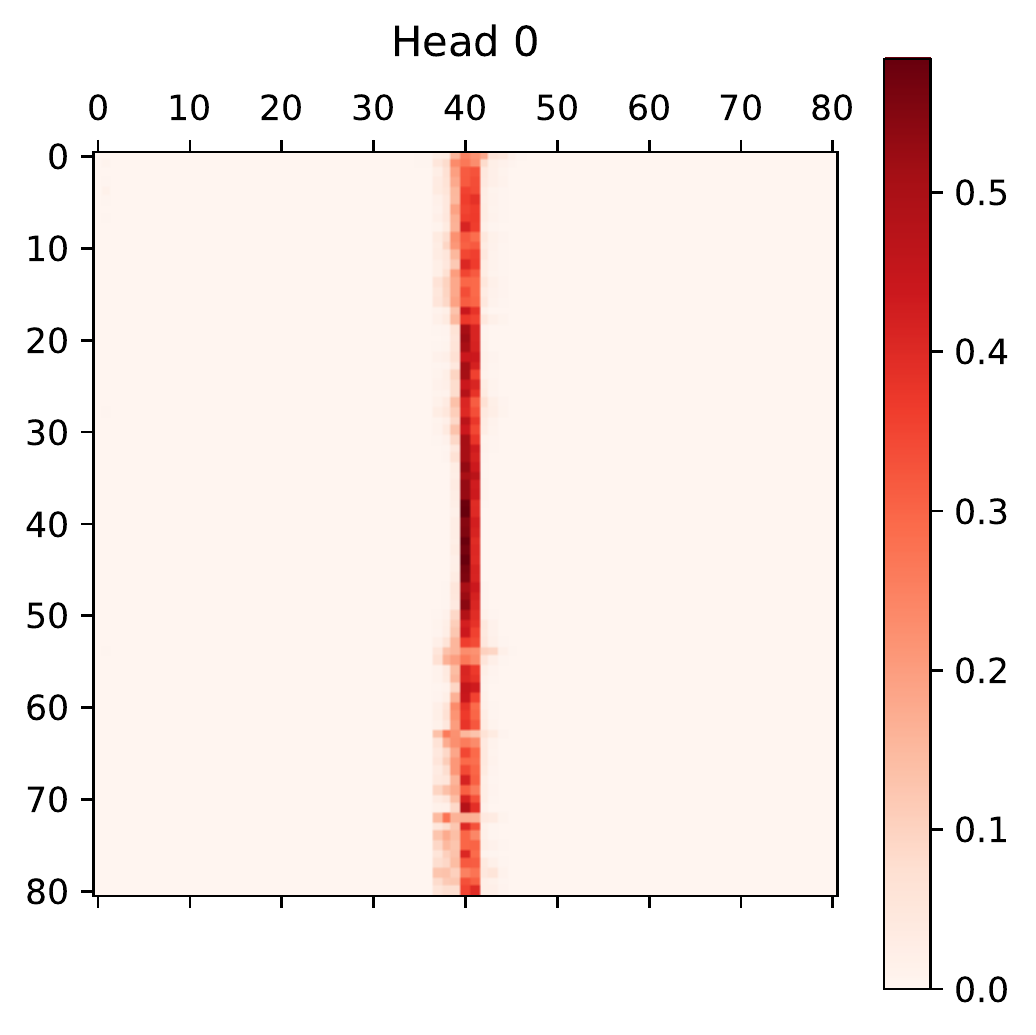}
\includegraphics[width=0.48\linewidth]{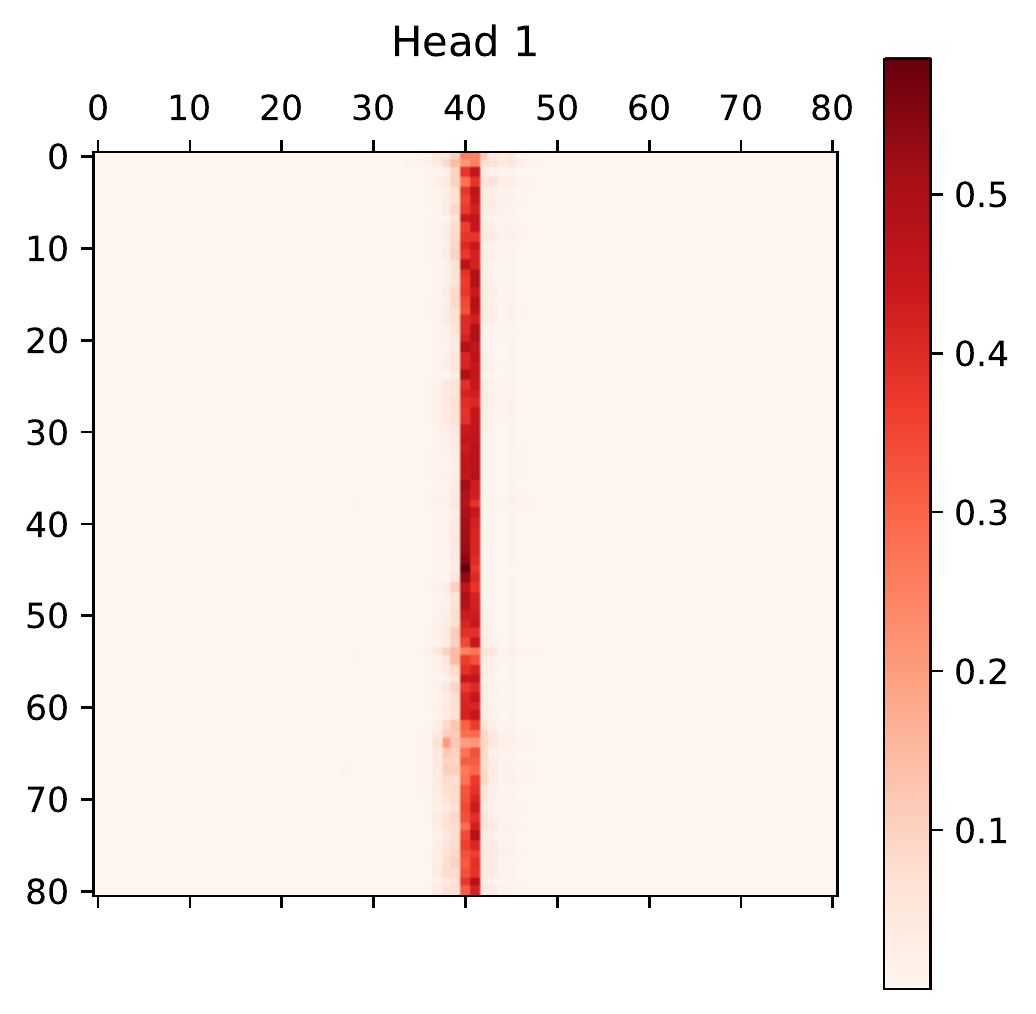}
\end{minipage}}
\subfigure[Layer 2]{
\begin{minipage}[t]{0.48\linewidth}
\includegraphics[width=0.48\linewidth]{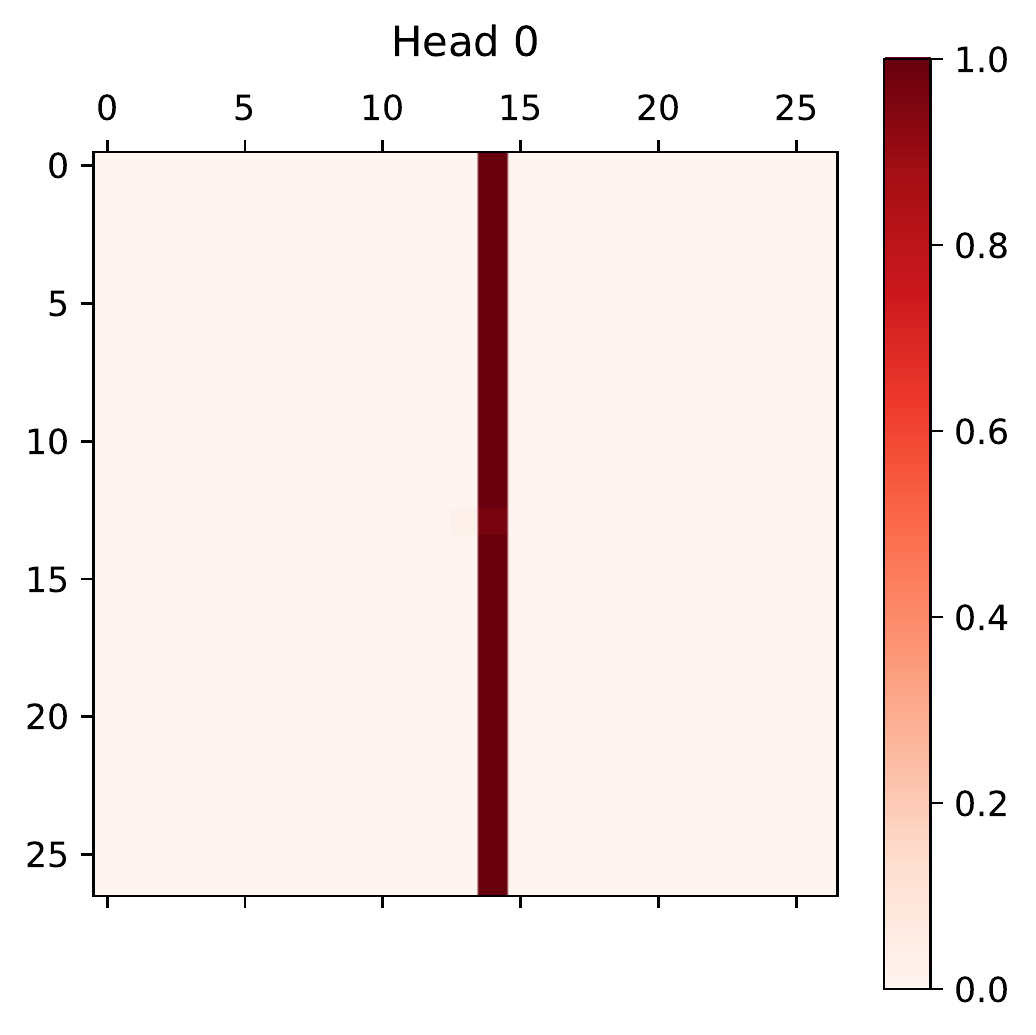}
\includegraphics[width=0.48\linewidth]{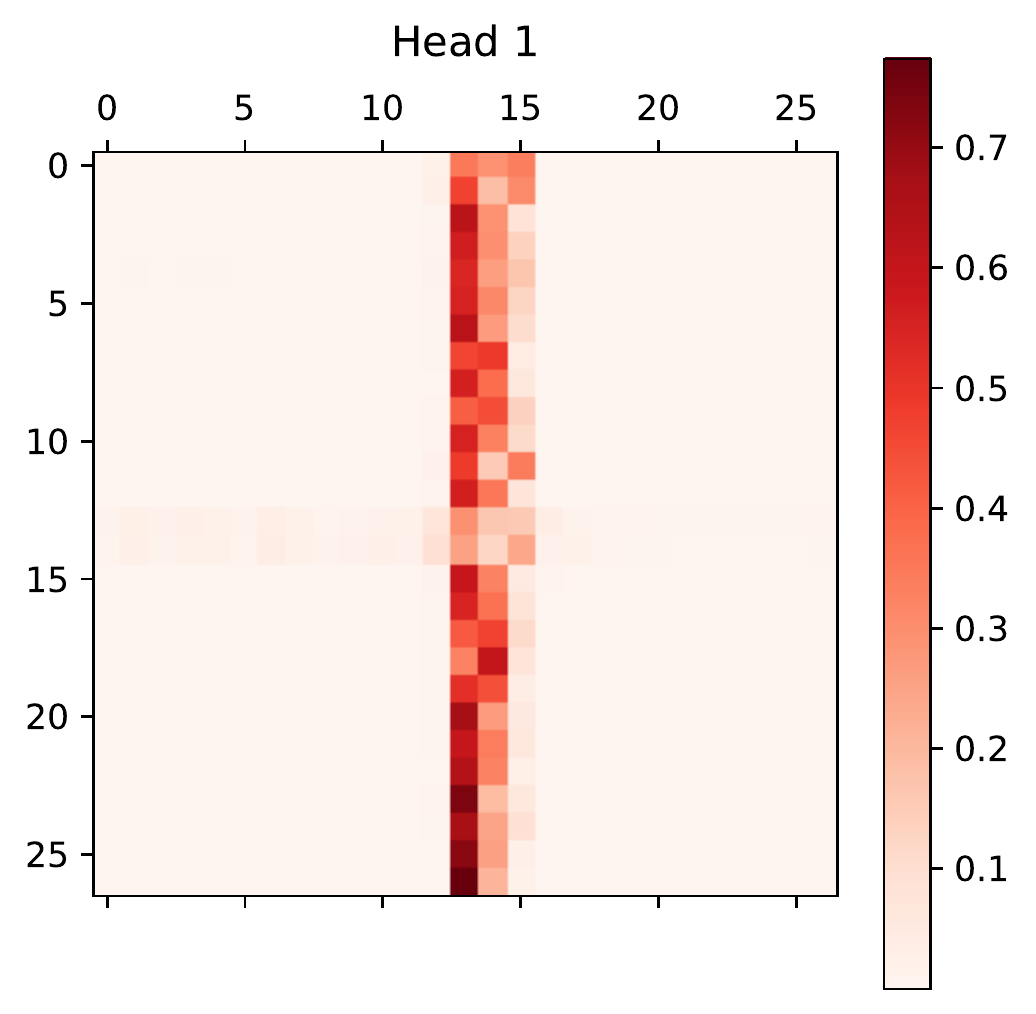}
\end{minipage}}
\subfigure[Layer 3]{
\begin{minipage}[t]{0.47\linewidth}
\includegraphics[width=0.48\linewidth]{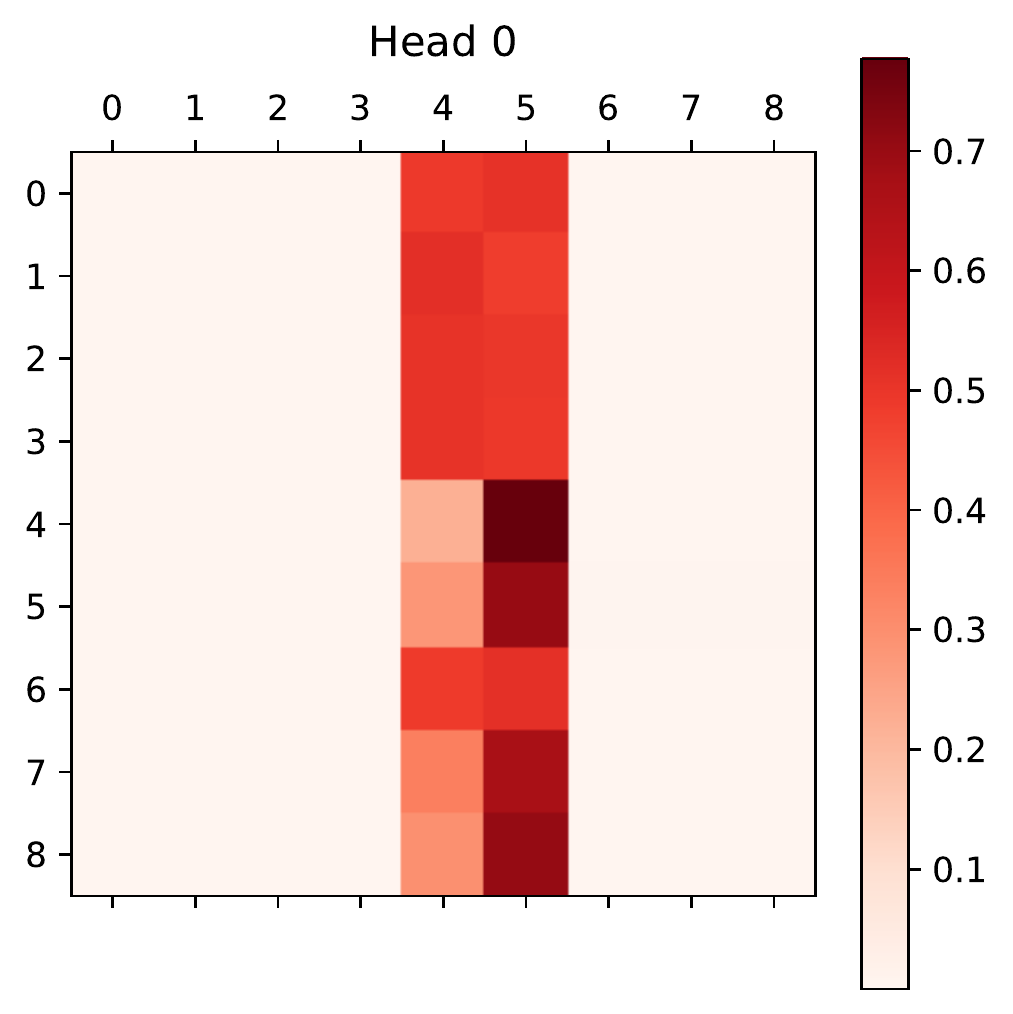}
\includegraphics[width=0.48\linewidth]{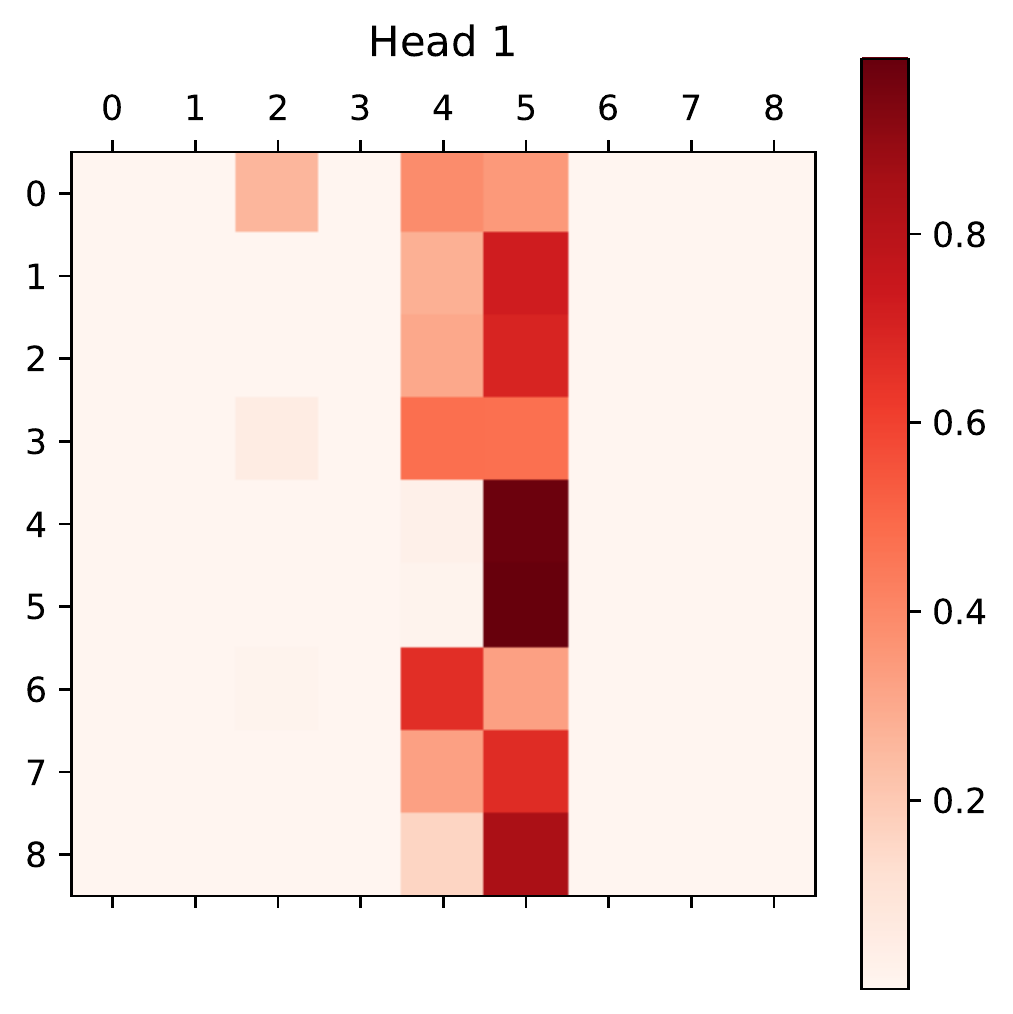}
\end{minipage}}
\vspace{-0.2cm}
\caption{Visualization of multi-head attention maps of MOFA in Stage \RNum{2}. We only show the first two heads in each Transformer layer.}
\vspace{-0.2cm}
\label{fig:stage2_STE}
\end{figure}

As we use STE~\cite{li2022exploiting}, a Transformer-based method, as the backbone network of MOFA, its attention maps can also be visualized in Fig.~\ref{fig:stage2_STE}. For $N=243$, the depth of Transformer is 5. Only the first 4 layers are shown. STE uses 1D convolution to reduce the temporal dimension layer by layer. As we set the stride to 3, the number of frames in each layer is reduced to 1/3 of that in the previous layer. A vertical line can be seen in all heads because MOFA is designed to predict the 3D pose in the current frame. Transformer forces all frames to attend to the current frame and its neighbours. Besides, the multi-head attention maps of MOFA and TEM show different patterns. The divergence of the functionality of these two modules is attributed to the presence of multi-frame loss.

\subsection{Qualitative Results on in-the-wild Videos}
\begin{figure}[t]
\centering
\subfigure{
\begin{minipage}[t]{\linewidth}
\centering
\includegraphics[width=0.32\linewidth]{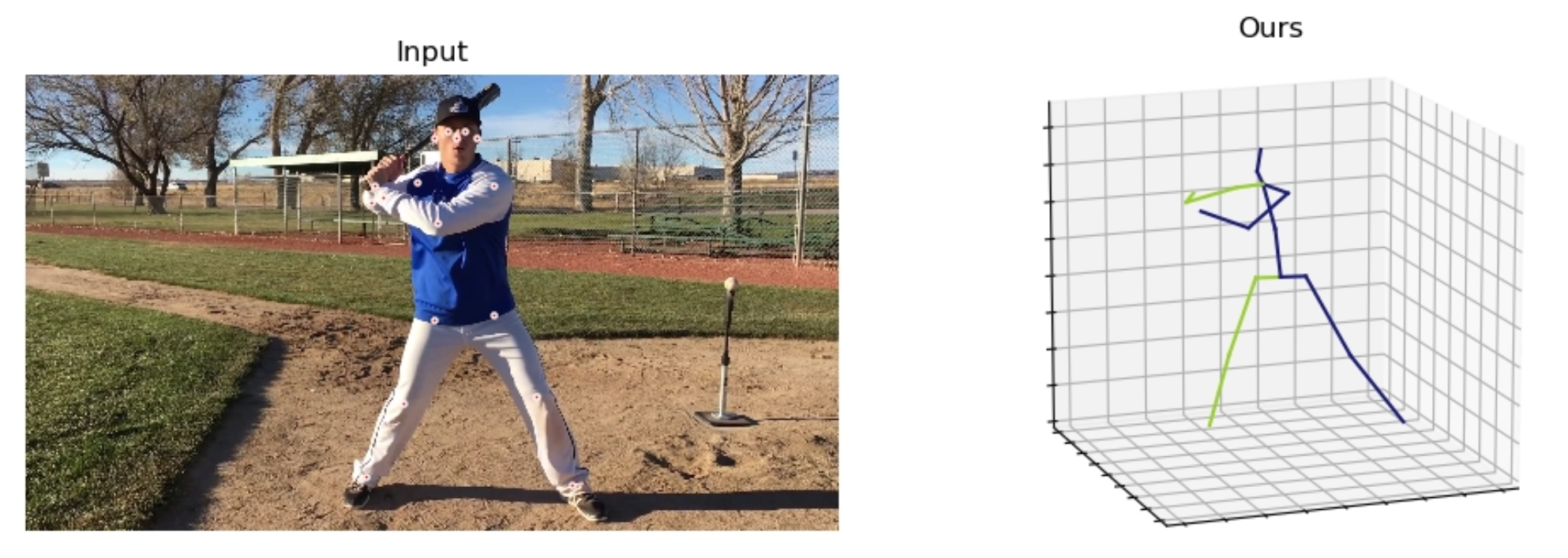}
\includegraphics[width=0.32\linewidth]{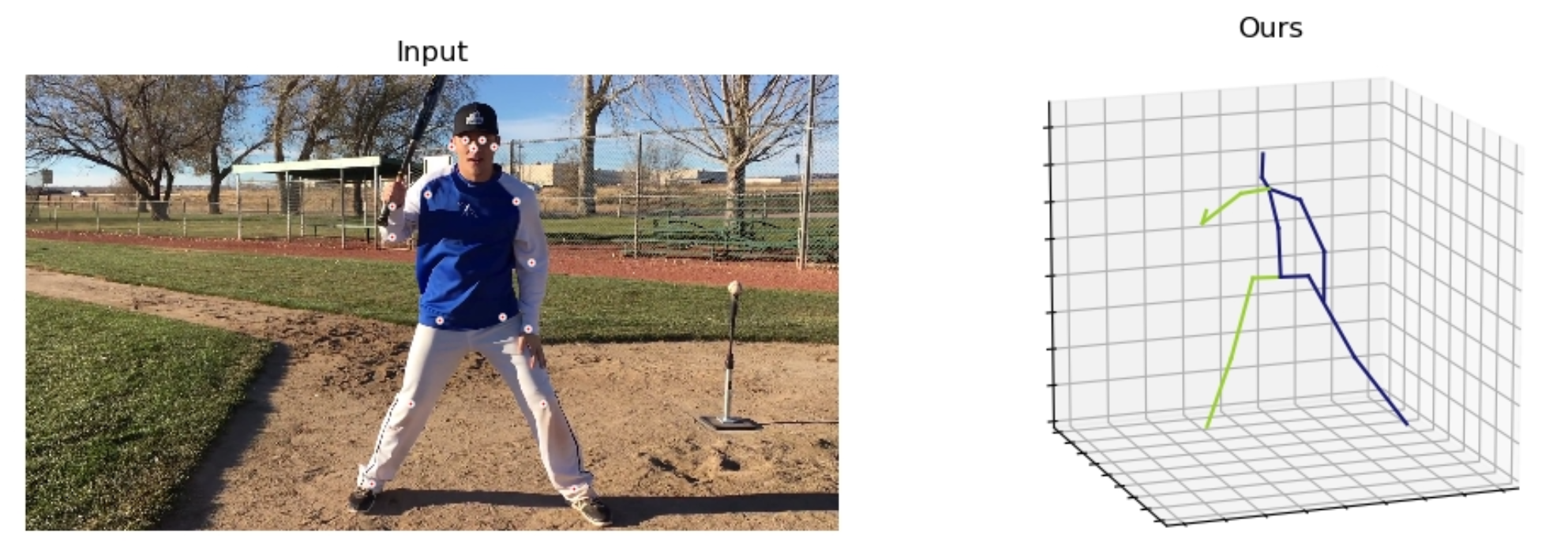}
\includegraphics[width=0.32\linewidth]{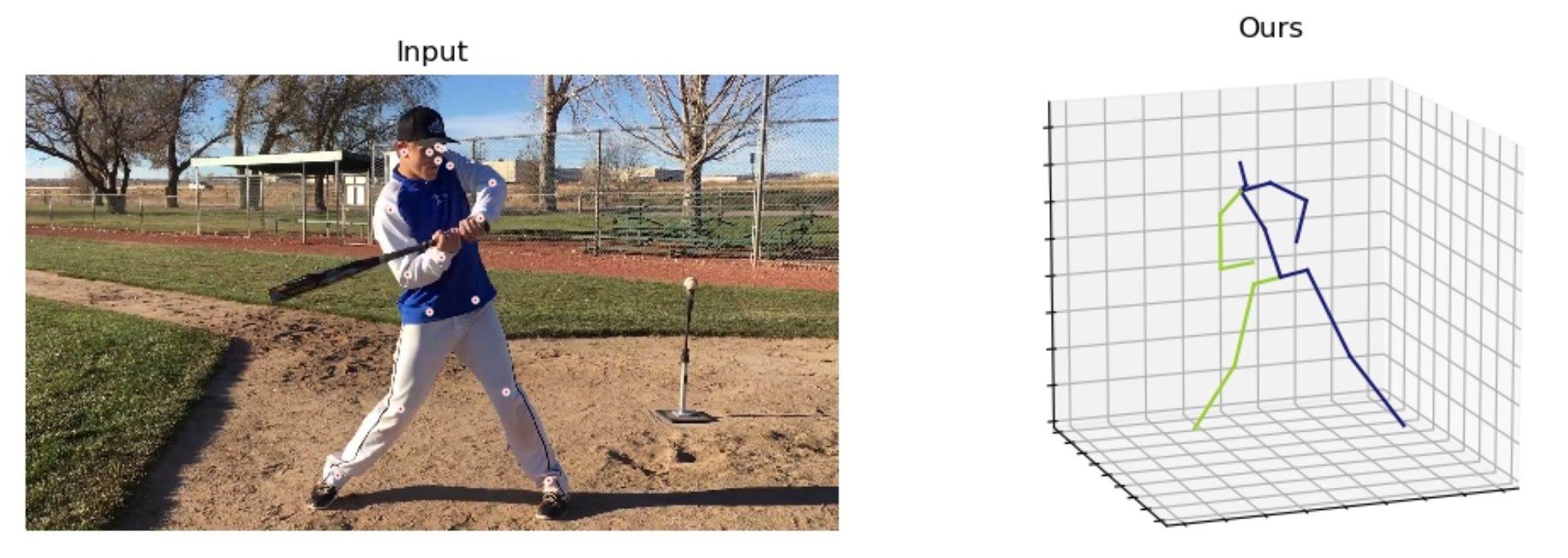}
\end{minipage}}
\subfigure{
\begin{minipage}[t]{\linewidth}
\centering
\includegraphics[width=0.32\linewidth]{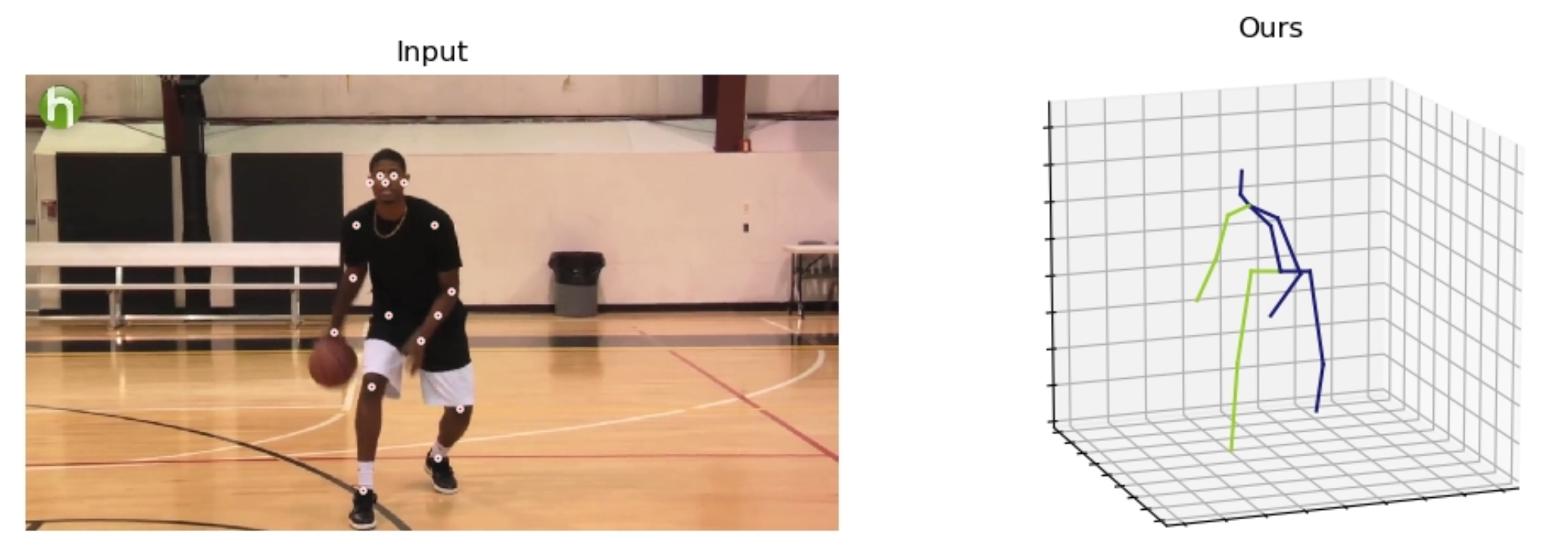}
\includegraphics[width=0.32\linewidth]{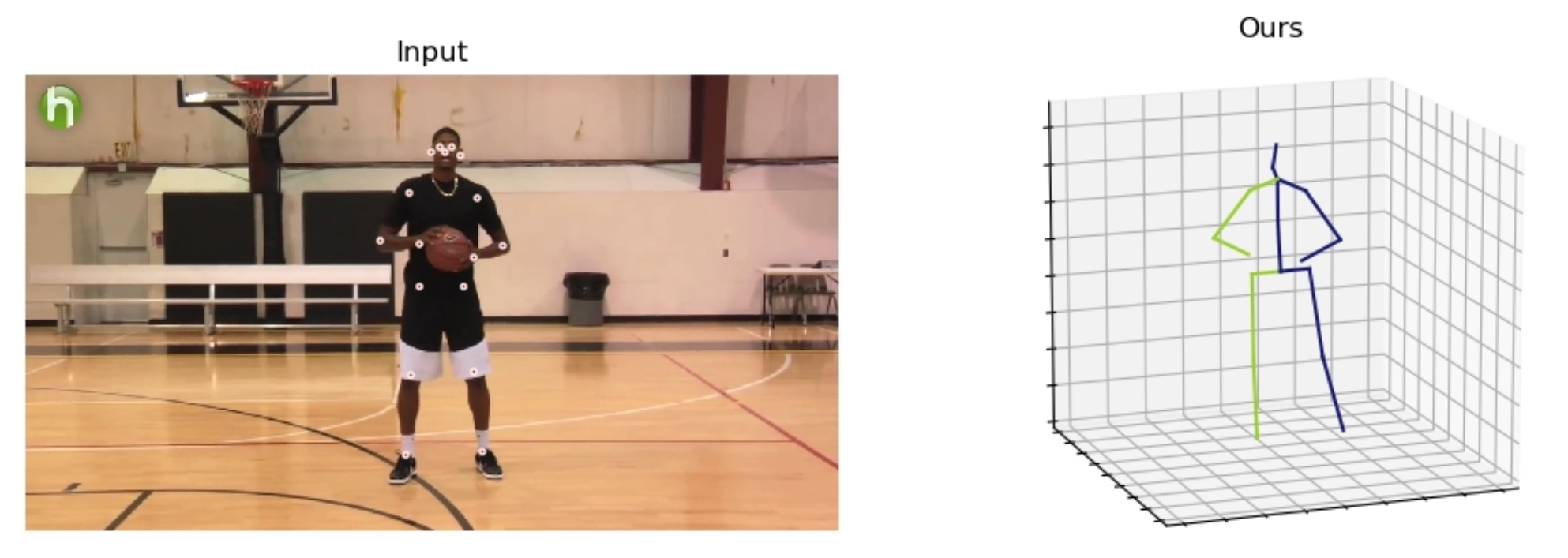}
\includegraphics[width=0.32\linewidth]{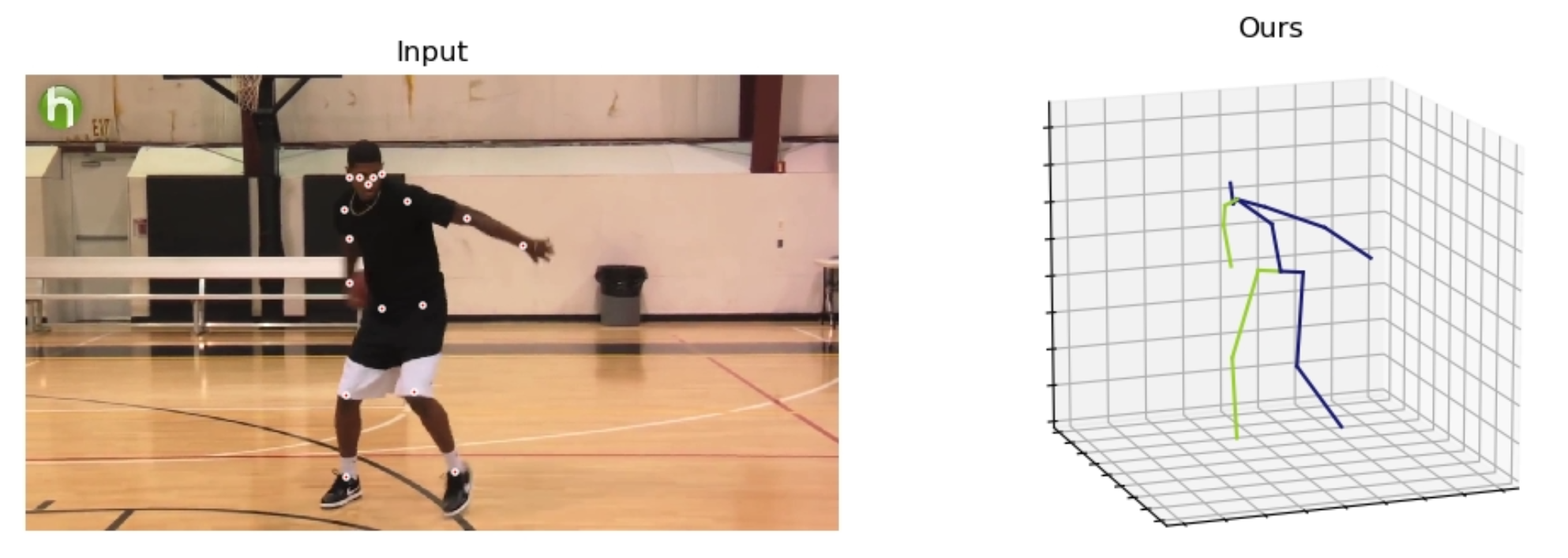}
\end{minipage}}
\subfigure{
\begin{minipage}[t]{\linewidth}
\centering
\includegraphics[width=0.32\linewidth]{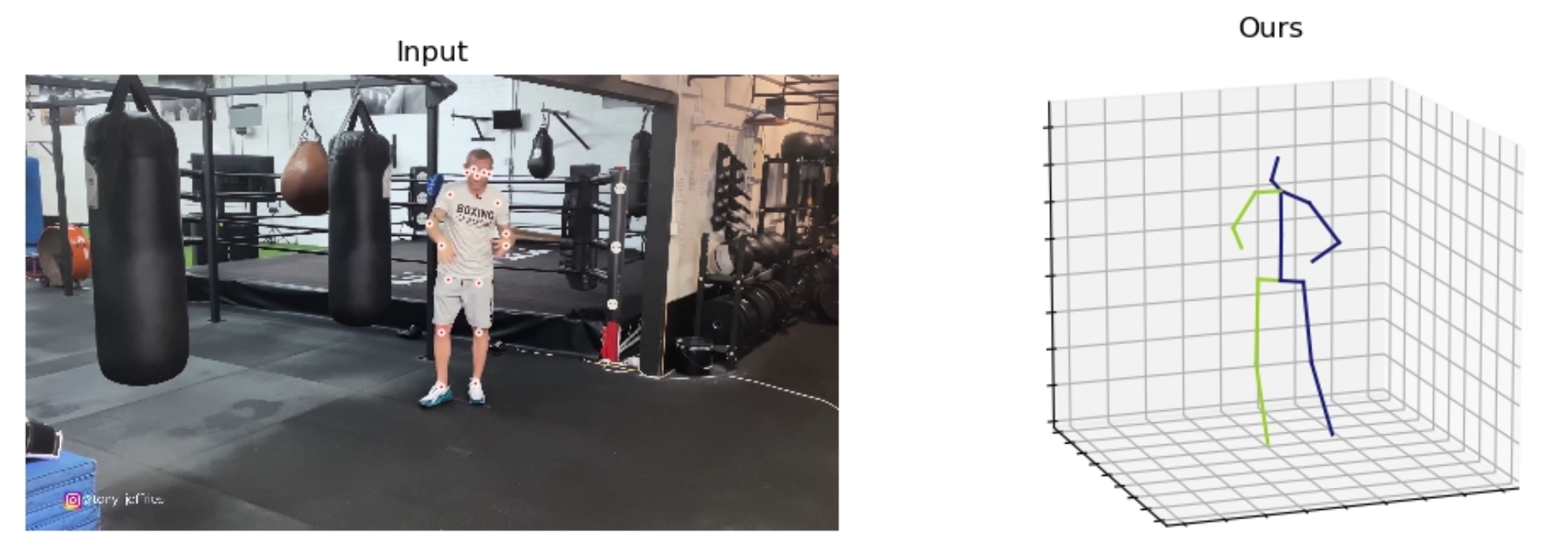}
\includegraphics[width=0.32\linewidth]{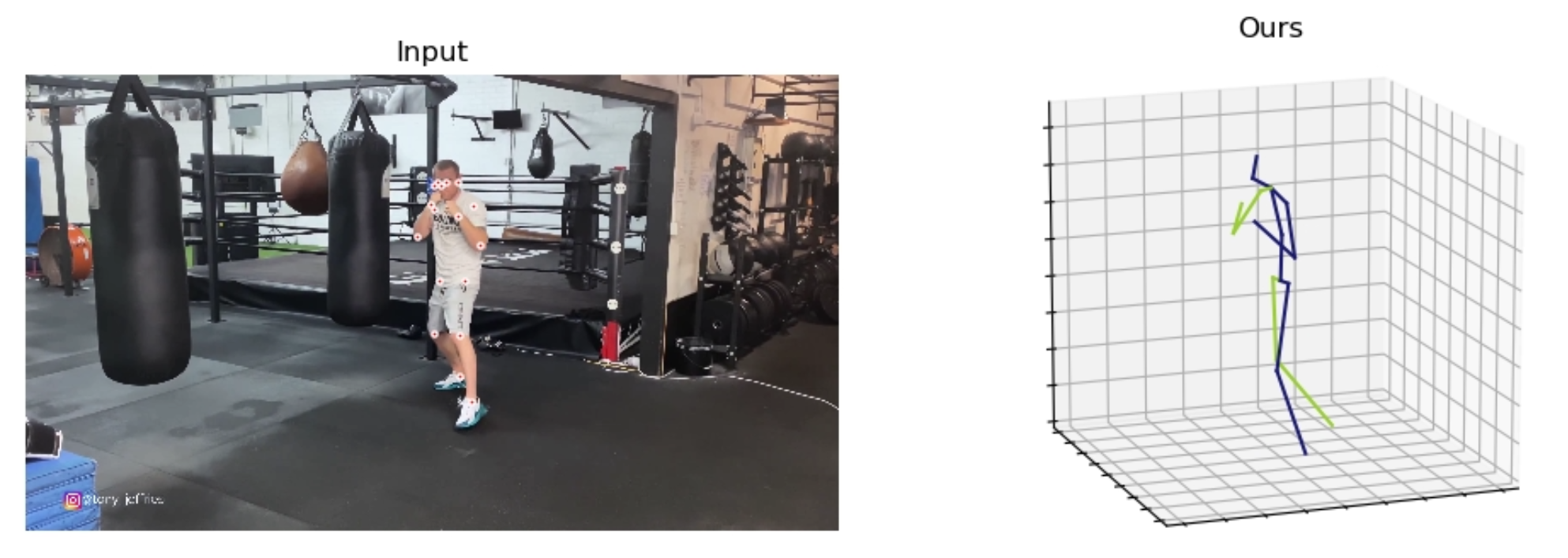}
\includegraphics[width=0.32\linewidth]{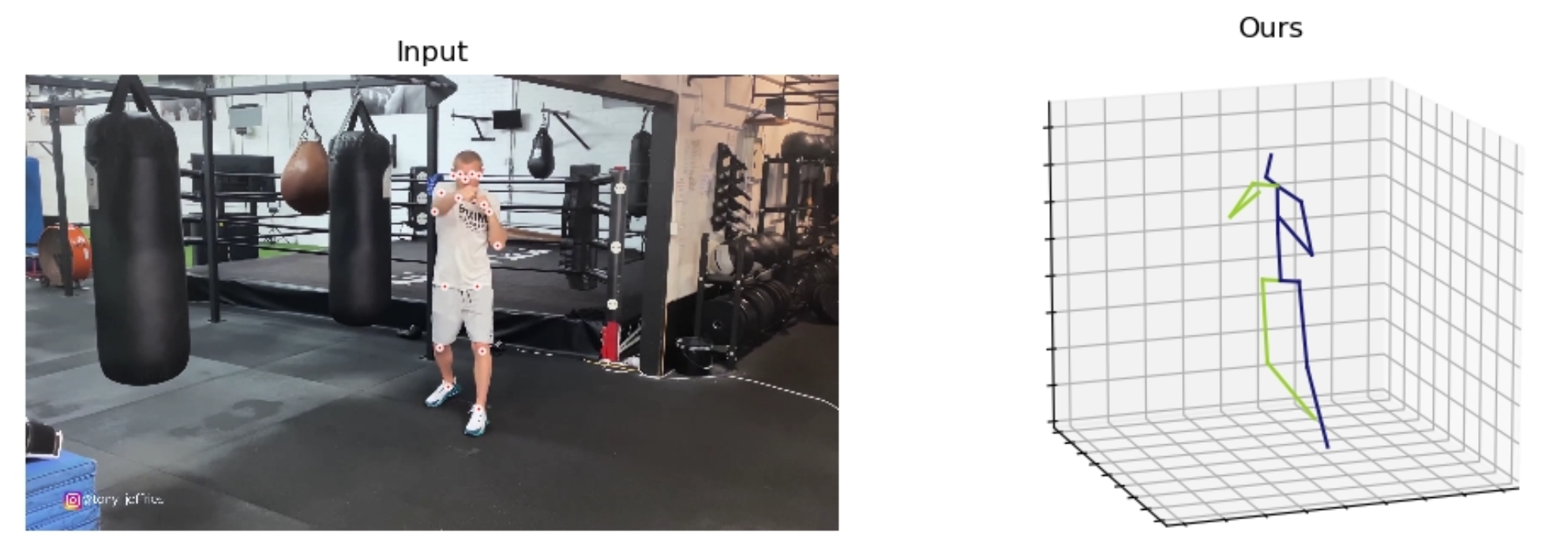}
\end{minipage}}
\subfigure{
\begin{minipage}[t]{\linewidth}
\centering
\includegraphics[width=0.32\linewidth]{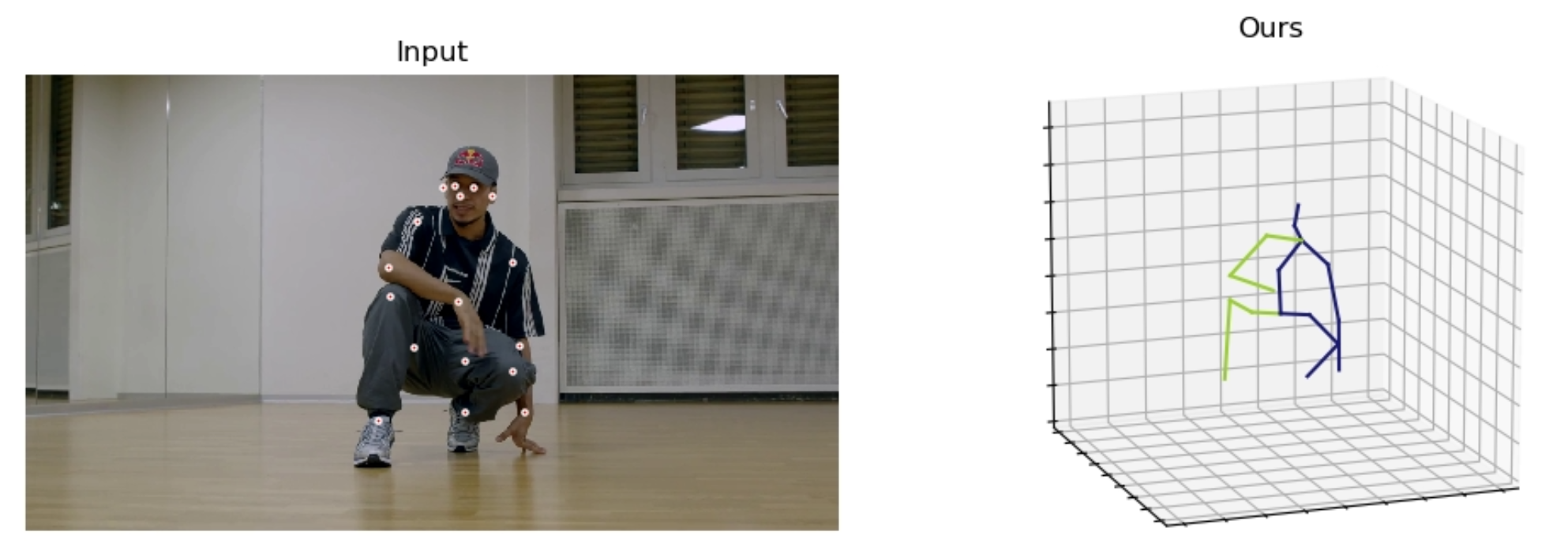}
\includegraphics[width=0.32\linewidth]{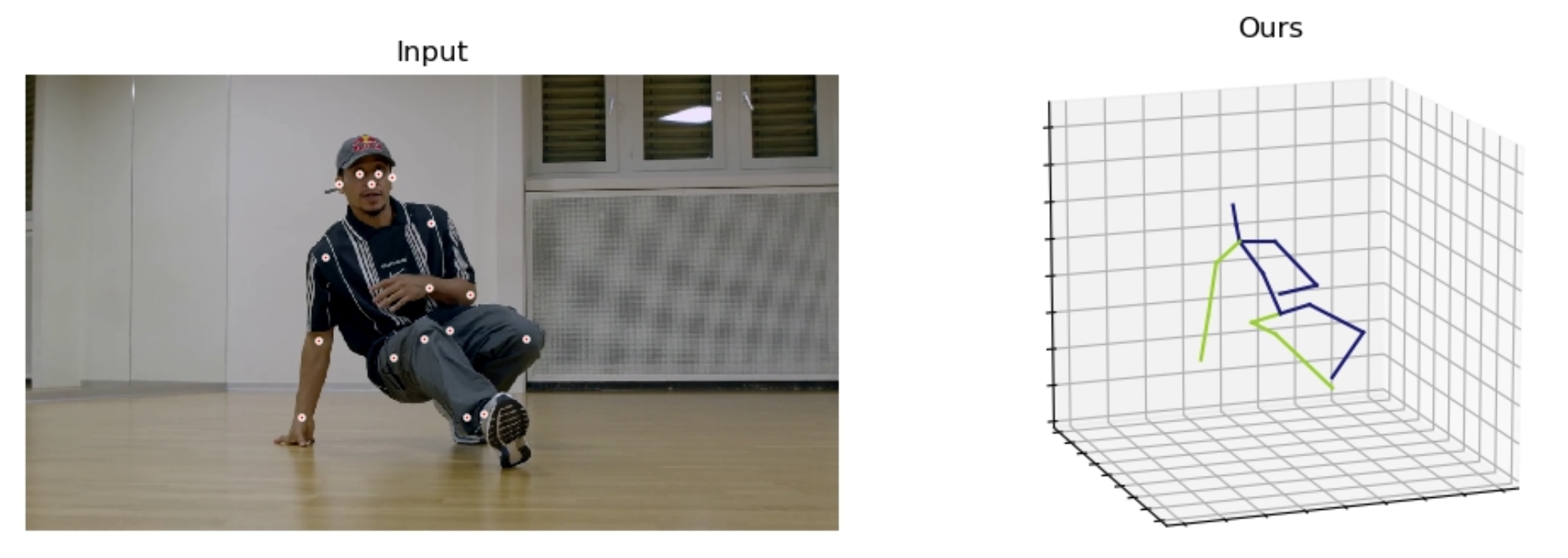}
\includegraphics[width=0.32\linewidth]{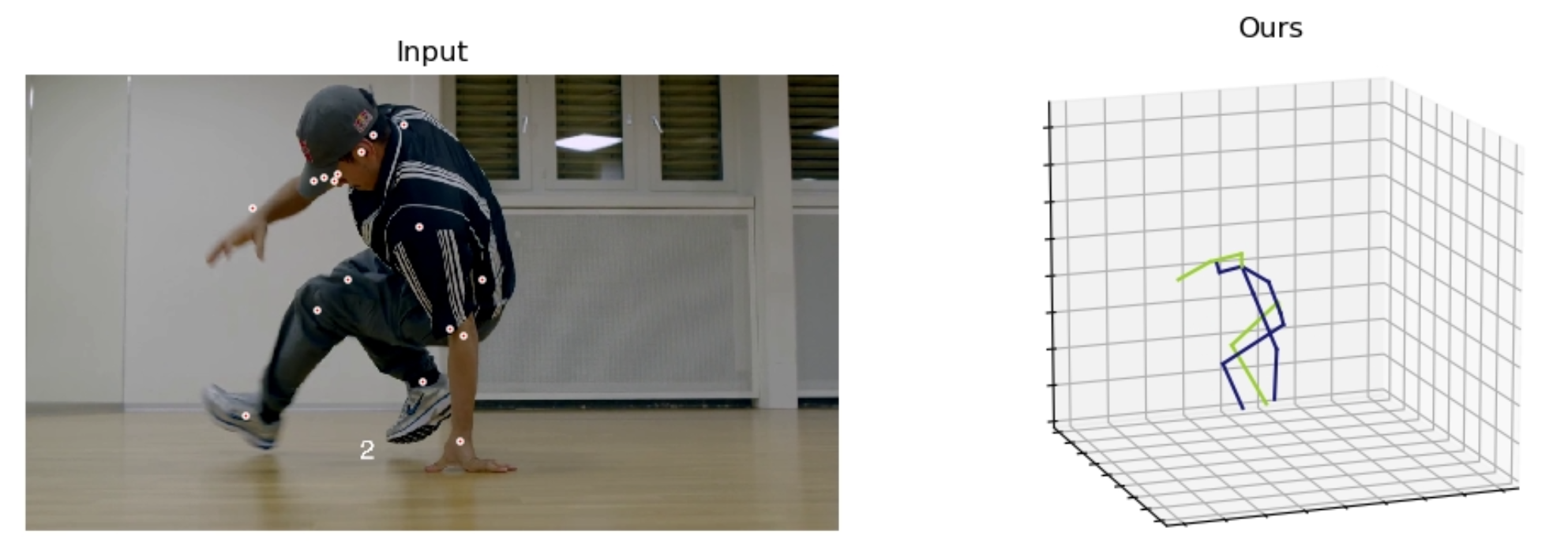}
\end{minipage}}
\subfigure{
\begin{minipage}[t]{\linewidth}
\centering
\includegraphics[width=0.32\linewidth]{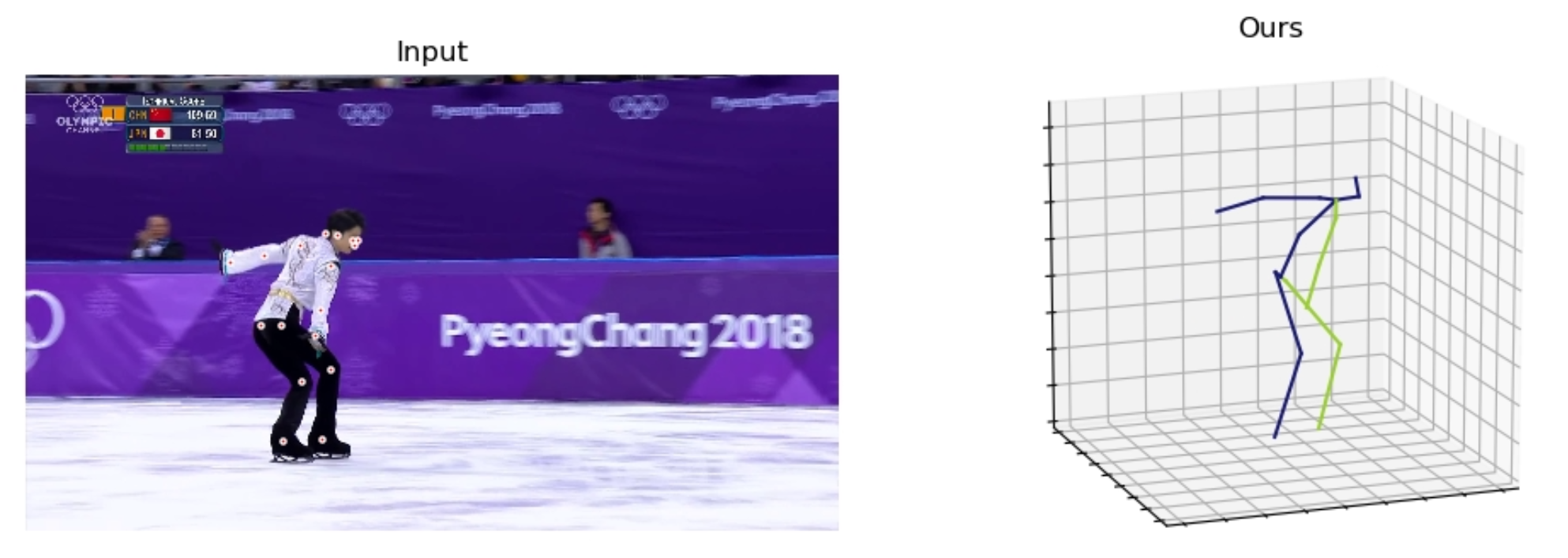}
\includegraphics[width=0.32\linewidth]{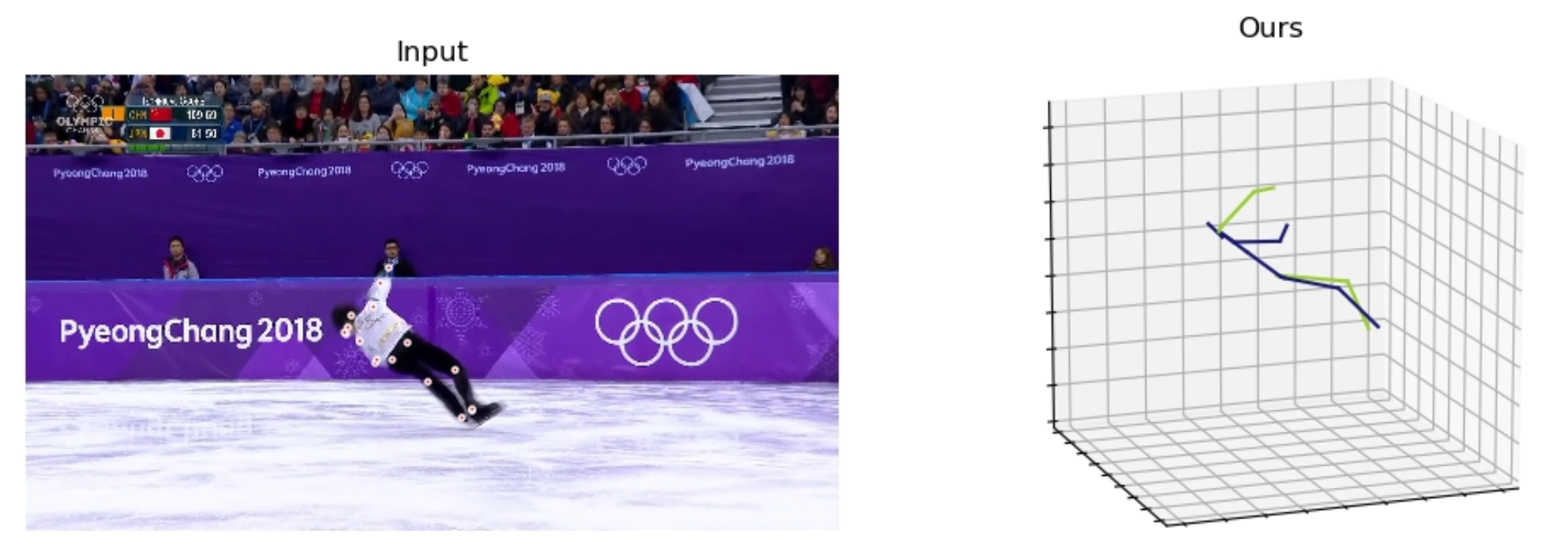}
\includegraphics[width=0.32\linewidth]{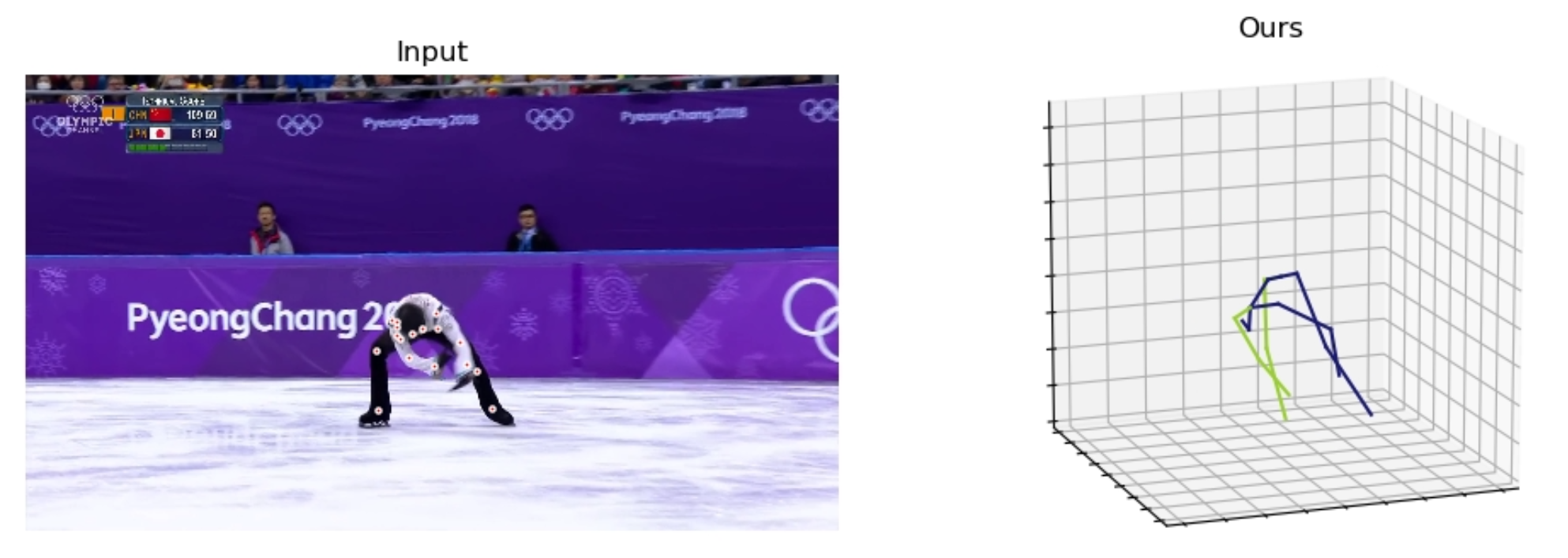}
\end{minipage}}
\subfigure{
\begin{minipage}[t]{\linewidth}
\centering
\includegraphics[width=0.32\linewidth]{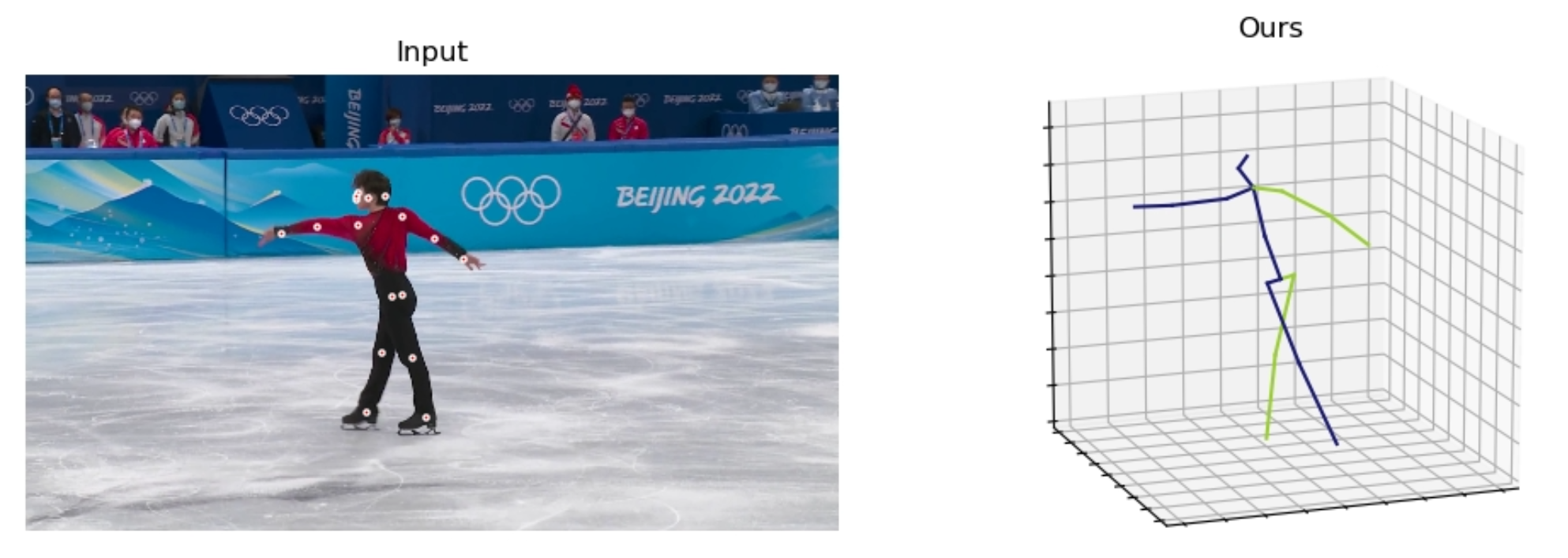}
\includegraphics[width=0.32\linewidth]{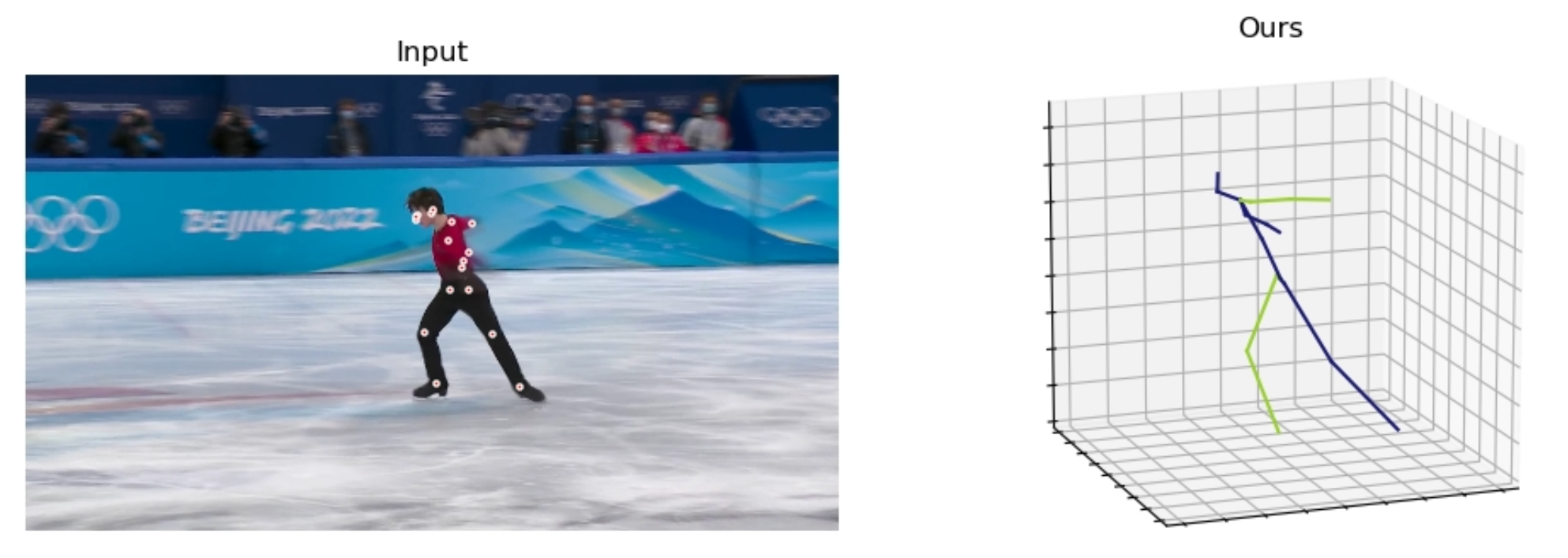}
\includegraphics[width=0.32\linewidth]{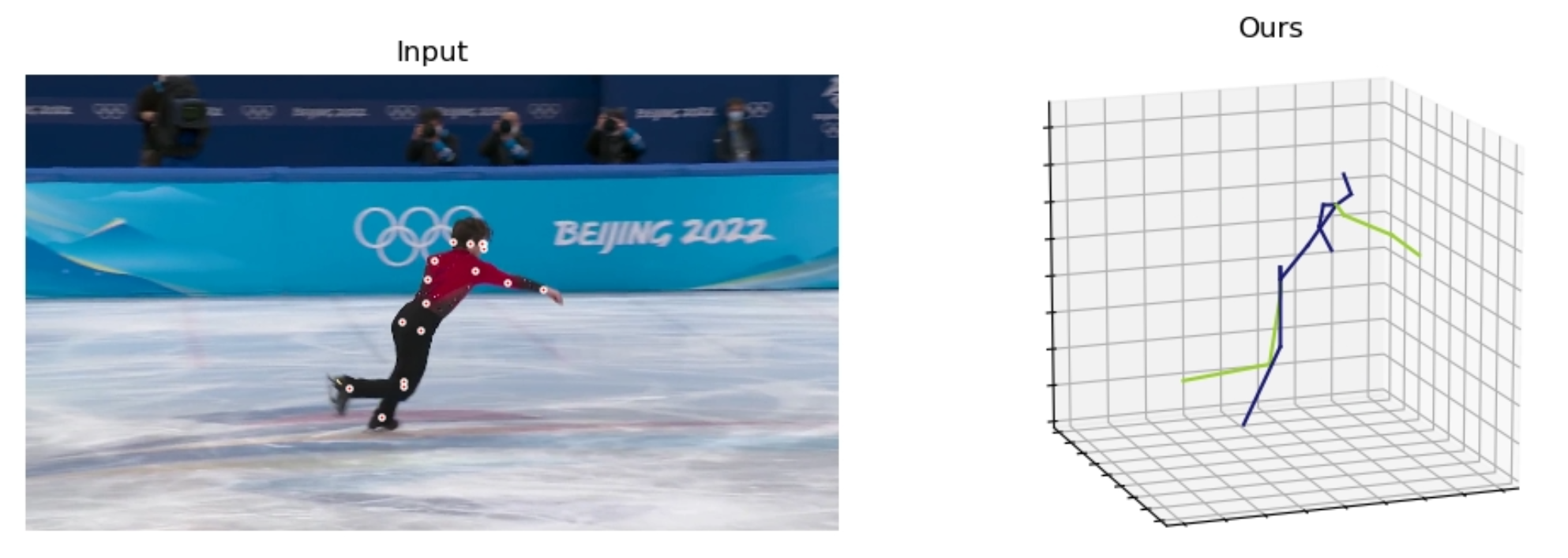}
\end{minipage}}
\subfigure{
\begin{minipage}[t]{\linewidth}
\centering
\includegraphics[width=0.32\linewidth]{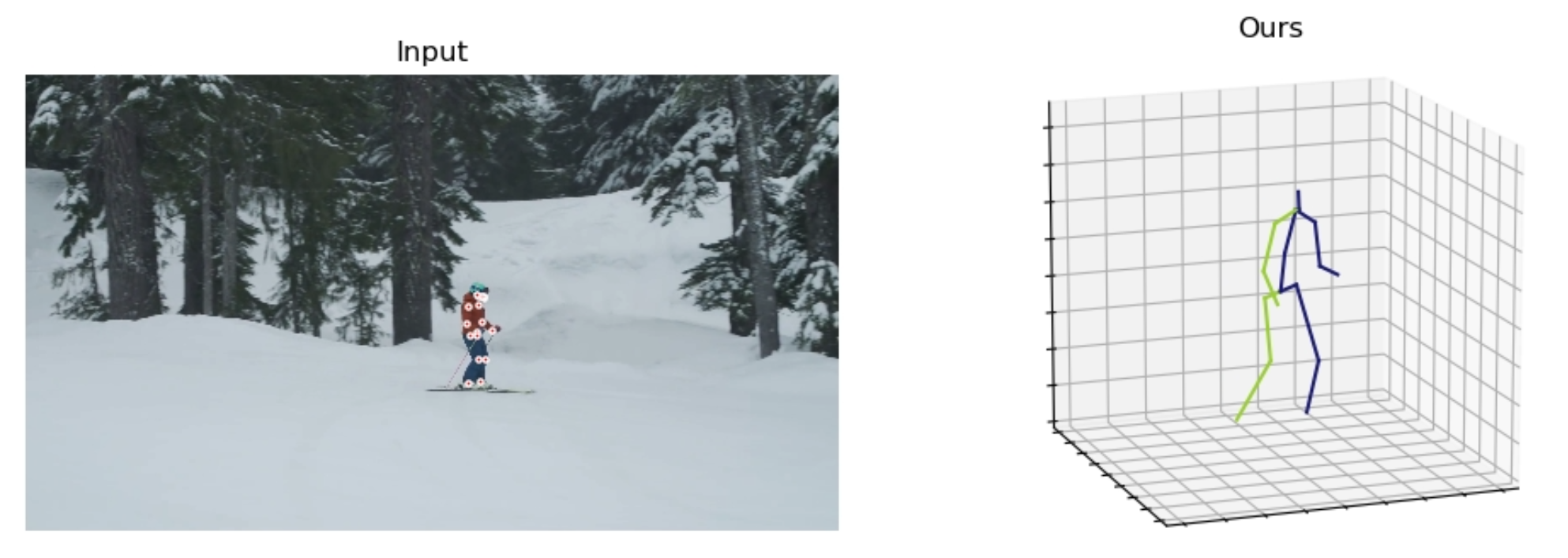}
\includegraphics[width=0.32\linewidth]{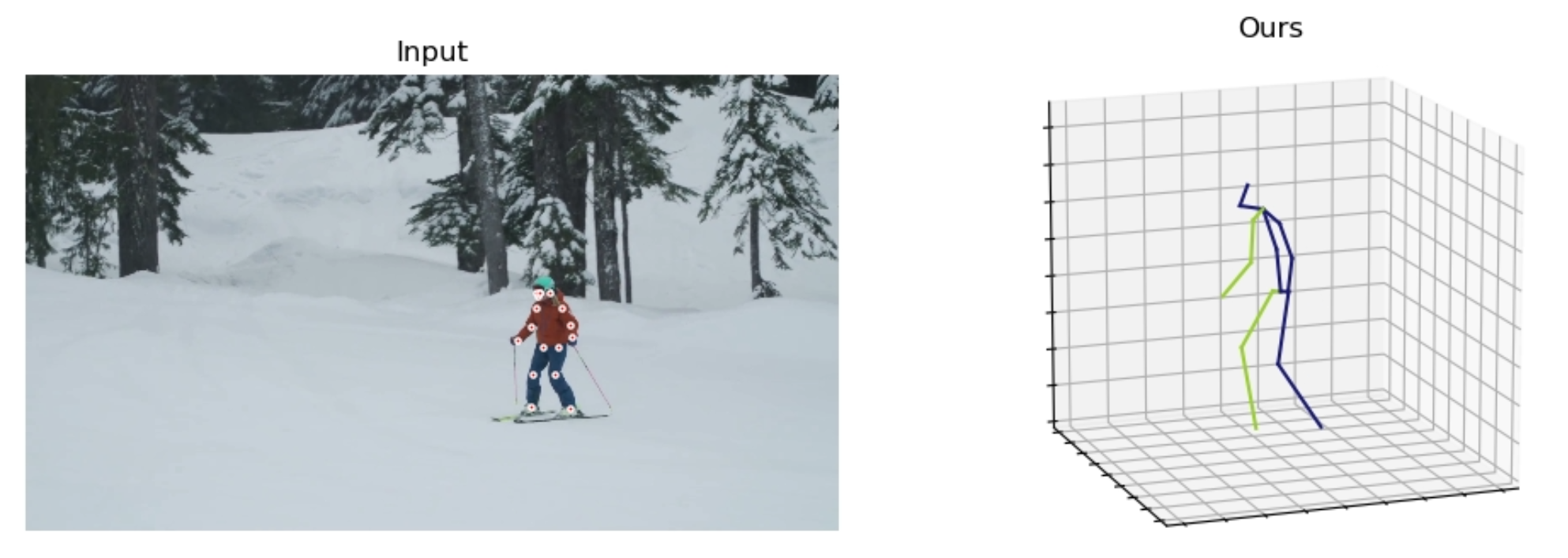}
\includegraphics[width=0.32\linewidth]{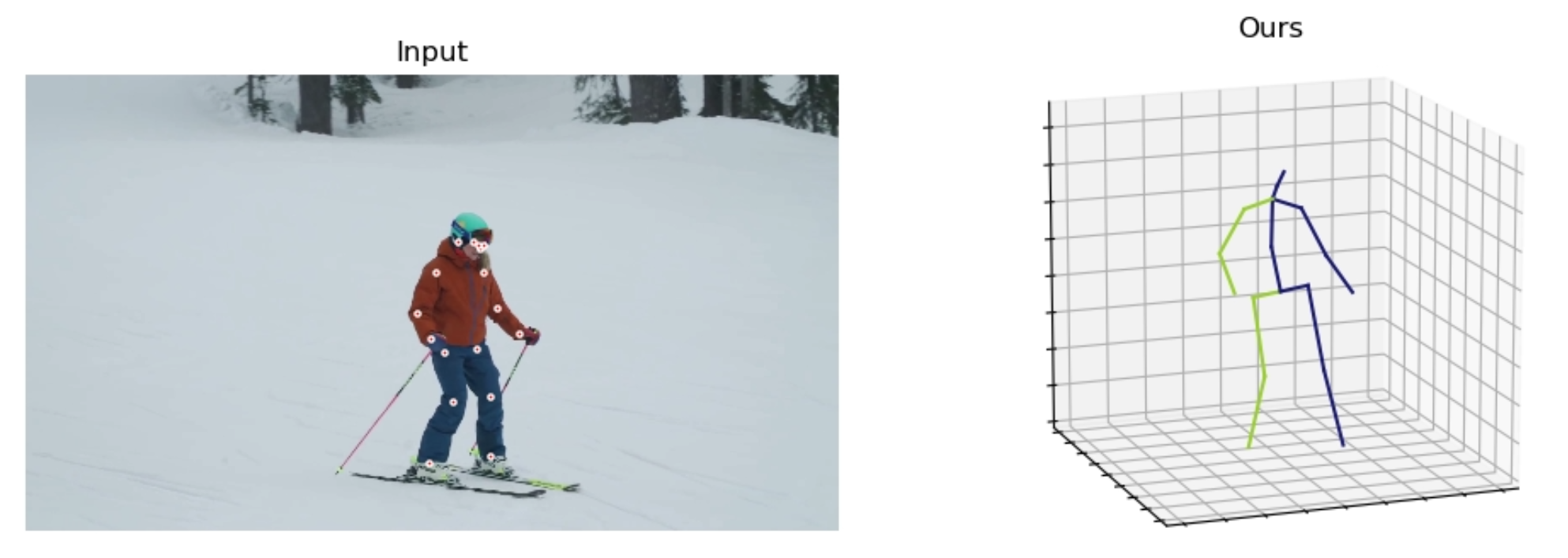}
\end{minipage}}
\subfigure{
\begin{minipage}[t]{\linewidth}
\centering
\includegraphics[width=0.32\linewidth]{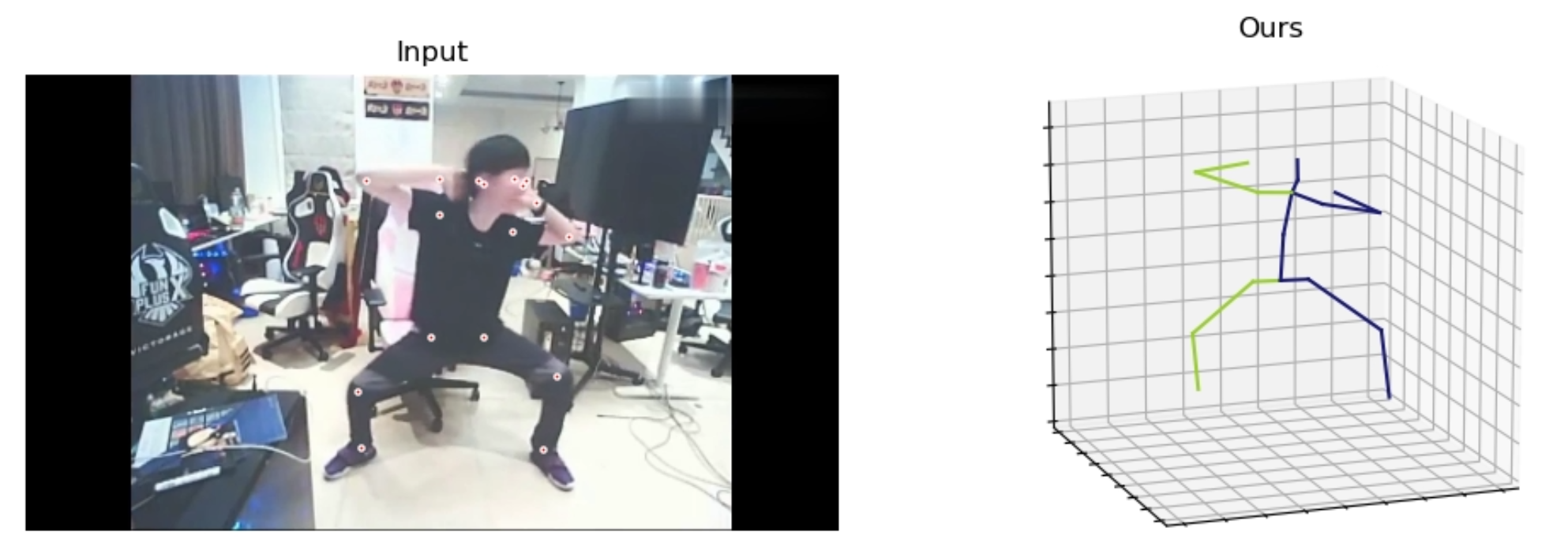}
\includegraphics[width=0.32\linewidth]{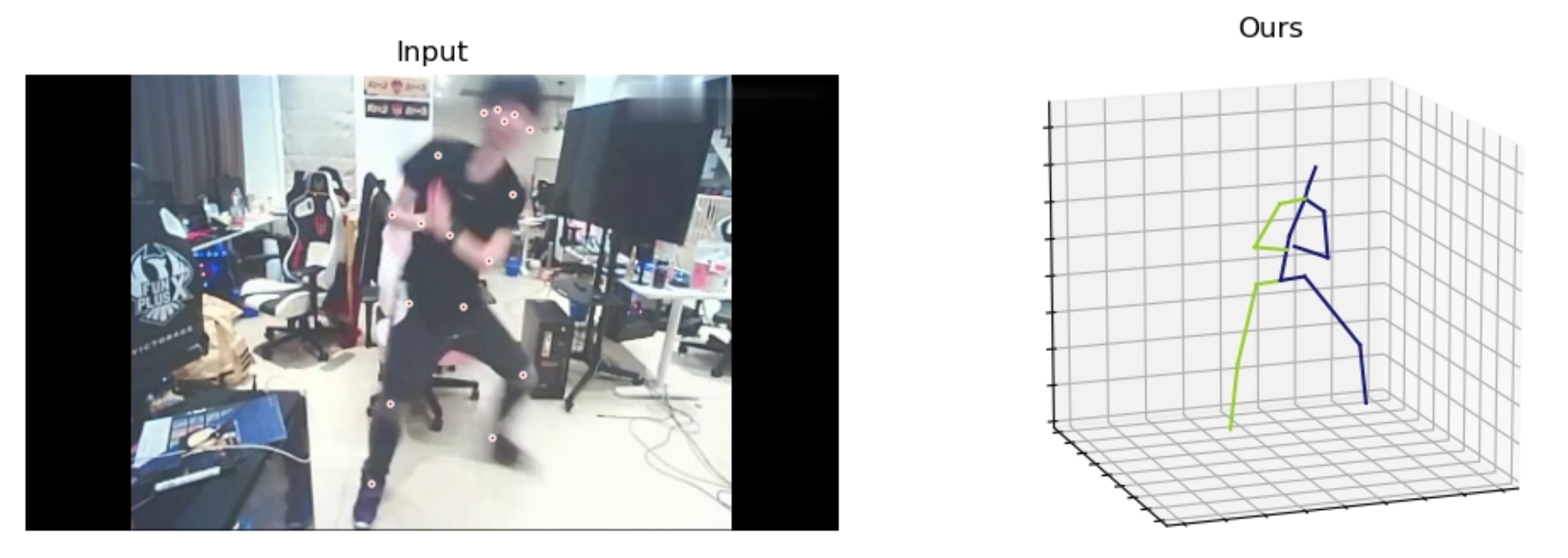}
\includegraphics[width=0.32\linewidth]{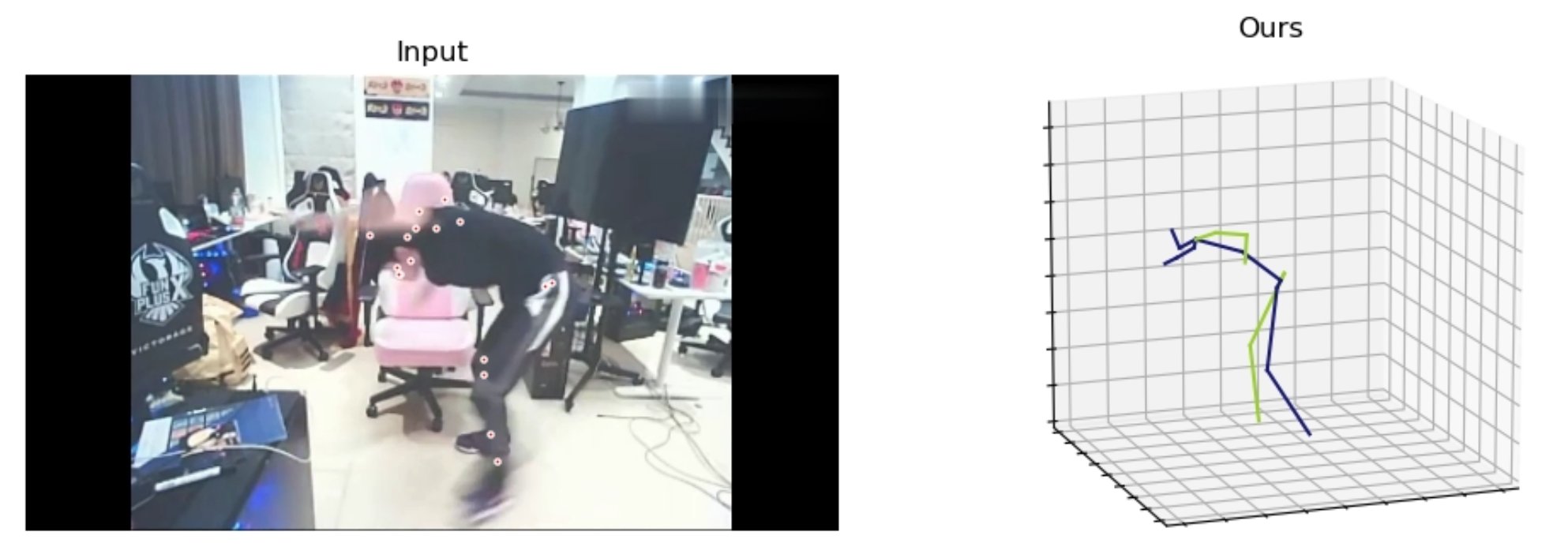}
\end{minipage}}
\vspace{-0.2cm}
\caption{Qualitative results of the proposed method on in-the-wild videos.}
\vspace{-0.2cm}
\label{fig:wild}
\end{figure}

We train our method on Human3.6M dataset and evaluate on in-the-wild videos. We use AlphaPose~\cite{fang2017rmpe} as the 2D keypoint detector to generate 2D poses. As shown in Fig.~\ref{fig:wild}, our method generalizes well to in-the-wild videos that often contain rare or unseen poses in the training set. 

\begin{table}[t]
\begin{center}
\caption{Analysis on the robustness in terms of MPJPE$\downarrow$. \ddag: w/o pre-training. RS: random shuffle.}
\vspace{-0cm}
\label{tab:noise}
\resizebox{\columnwidth}{!}{\begin{tabular}{c|cccccc}
\hline
Method&GT&GT+$\mathcal N$(0,0.01)&GT+$\mathcal N$(0,0.03)&GT+$\mathcal N$(0,0.05)&GT+RS&CPN\\
\hline
STMO-S\ddag&30.3&34.8&48.2&70.9&59.4&43.9\\
P-STMO-S&\textbf{29.3}($\downarrow$3.3\%)&\textbf{33.7}($\downarrow$3.2\%)&\textbf{46.9}($\downarrow$2.7\%)&\textbf{69.5}($\downarrow$2.0\%)&\textbf{57.7}($\downarrow$2.9\%)&\textbf{43.0}($\downarrow$2.1\%)\\
\hline
\end{tabular}}
\end{center}
\vspace{-0.4cm}
\end{table}

\subsection{Generalization Ability and Robustness}
To validate the generalization ability of the proposed pre-training method, we conduct experiments on Human3.6M, where three camera views (cam 0,1,2) are used in Stage \RNum{1} and the other camera view (cam 3) is used in Stage \RNum{2}. The results show that P-STMO-S (w/ pre-training) and STMO-S (w/o pre-training) achieve 44.4mm and 45.2mm MPJPE respectively, which demonstrates that pre-training on several cameras can help the network generalize to a different camera view.

To validate the robustness of the proposed pre-training method, we utilize the ground truth (GT) of 2D keypoints as noiseless inputs, to which Gaussian noise with different $\sigma$ is added. Note that CPN can be regarded as adding complex noise introduced by the 2D detector to the ground truth. Besides, the random shuffle (RS) strategy is tested. This strategy randomly disrupts the order of the input frames while leaving the order of the output frames unchanged. Table~\ref{tab:noise} shows pre-training still delivers performance gains in the case of input noise and RS, which verifies its robustness. 

\clearpage

\clearpage

\bibliographystyle{splncs04}
\bibliography{ref}
\end{document}